\pdfoutput=1
\documentclass[11pt]{article}

\usepackage[]{ACL2023}

\usepackage{times}
\usepackage{latexsym}
\usepackage{graphicx}
\usepackage{multirow}
\usepackage{algorithm}
\usepackage{algorithmic}
\usepackage{amsfonts}
\usepackage[T1]{fontenc}
\usepackage[utf8]{inputenc}

\usepackage{microtype}

\usepackage{inconsolata}

\title{\textsc{IndicSentEval}: How Effectively do Multilingual Transformer Models encode Linguistic Properties for Indic Languages?}

\author{Akhilesh Aravapalli$^1$, Mounika Marreddy$^{2}$, \\ \textbf{Radhika Mamidi}$^{1}$, \textbf{Manish Gupta}$^{1,3}$, \textbf{Subba Reddy Oota}$^{4}$\\
$^1$IIIT Hyderabad, India,
  $^2$University of Bonn, Germany,
   $^3$Microsoft, India, $^4$Inria, France\\
  \small \texttt{aforakhilesh@gmail.com, mmarredd@uni-bonn.de, radhika.mamidi@iiit.ac.in}, \\ \small \texttt{gmanish@microsoft.com, subbareddyoota@gmail.com}}

\begin{document}
\maketitle
\begin{abstract}
Transformer-based models have revolutionized the field of natural language processing. To understand why they perform so well and to assess their reliability, several studies have focused on questions such as: Which linguistic properties are encoded by these models, and to what extent? How robust are these models in encoding linguistic properties when faced with perturbations in the input text? However, these studies have mainly focused on BERT and the English language. In this paper, we investigate similar questions regarding encoding capability and robustness for 8 linguistic properties across 13 different perturbations in 6 Indic languages, using 9 multilingual Transformer models (7 universal and 2 Indic-specific). To conduct this study, we introduce a novel multilingual benchmark dataset, \textsc{IndicSentEval}, containing approximately $\sim$47K sentences. Our probing analysis of surface, syntactic, and semantic properties reveals that, while almost all multilingual models demonstrate consistent encoding performance for English, surprisingly, they show mixed results for Indic languages. As expected, Indic-specific multilingual models capture linguistic properties in Indic languages better than universal models. Intriguingly, universal models broadly exhibit better robustness compared to Indic-specific models, particularly under perturbations such as dropping both nouns and verbs, dropping only verbs, or keeping only nouns. Overall, this study provides valuable insights into probing and perturbation-specific strengths and weaknesses of popular multilingual Transformer-based models for different Indic languages. We make our code and dataset publicly available\footnote{\url{https://github.com/aforakhilesh/IndicBertology}\label{datafootnote}}.

\end{abstract}

% \vspace{-0.2cm}
\begin{figure}[!t]
    \centering
    \includegraphics[width=\columnwidth]{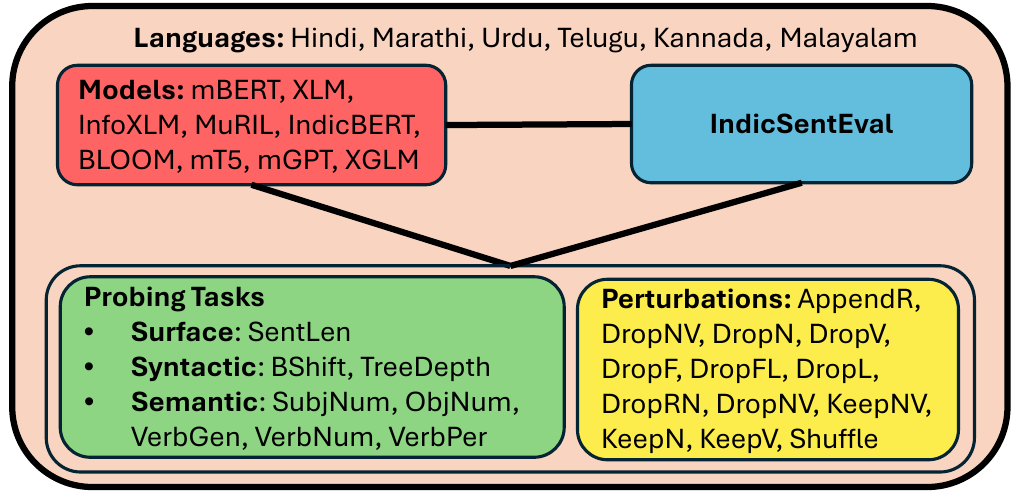}
     % \vspace{-0.3cm}
    \caption{We evaluate 9 multilingual Transformer models on 8 probing tasks in 6 Indic languages using our \textsc{IndicSentEval} dataset. We analyze the effects of 13 perturbations on the performance of these models.}    
    \label{fig:conceptualDiagram}
\end{figure}

\section{Introduction}
%\vspace{-0.1cm}
 Transformer-based language models~\citep{vaswani2017attention}, pretrained for both mono-lingual and multilingual contexts using millions of text documents, have demonstrated substantial enhancements in the performance of various natural language processing (NLP) tasks~\citep{devlin2018bert,pires2019multilingual,conneau2019unsupervised,radford2019language,brown2020language,wang2019superglue,wang2018glue}. To understand what types of linguistic properties (surface, syntactic and semantic) are encoded across layers of Transformer-based models, initial studies have investigated the layer-wise representations via a range of probing tasks~\citep{adi2016fine,hupkes2018visualisation,conneau2018you,rogers2020primer,jawahar2019does,mohebbi2021exploring}. However, these studies focus solely on English. 
Although, several studies have examined the presence of shared representation aspects across widely spoken languages in multilingual models~\citep{chi2020finding,acs2024morphosyntactic}, 
unfortunately, there is no work that investigates the extent to which multilingual Transformer-based models encode linguistic properties for different Indic languages.
Further, recent neuro-AI studies have revealed that the brain uses different parsing strategies for different linguistic properties, which further differ across languages~\citep{zhang-etal-2022-brain,oota2023syntactic}. This inspired us to study the nuances of how multilingual Transformer-based models capture linguistic properties across layers and languages.

%Understanding the degree to which a particular Transformer-based language model encodes a basic linguistic property for a particular language can lead to more insights into the NLP systems and drive task-specific model choice decisions. 
% Several recent studies focus on widely spoken languages such as French, Spanish, German and Chinese to examine the presence of shared representation aspects across languages in multilingual models~\cite{chi2020finding,acs2023morphosyntactic}. 

Indic languages offer a rich tapestry of linguistic features that contribute to the global linguistic diversity. Hence, in this paper, we focus on Indic languages. We study the degree to which linguistic properties of Indic languages are captured by two kinds of multilingual models: universal and Indic-specific models. Universal multilingual models have been pretrained using a variety of pretraining objectives and using data which contains a small and varying fraction of Indic languages across languages and models (Tables~\ref{tab:modelDetails},~\ref{tab:Tokens_multilingual}, and~\ref{tab:data_proprtion_models} in Appendix~\ref{app:modelDetails}). Indic-specific models~\citep{kakwani2020indicnlpsuite,dabre2021indicbart,kumar2022indicnlg} have been specifically trained on Indic language data only. %Unfortunately, the linguistic properties encoded in different layers for different languages remains unclear. In this paper, 
Specifically, we focus on 6 Indic languages: three Indo-European languages (Hindi, Marathi, Urdu) and three Dravidian languages (Telugu, Kannada, Malayalam).

Even if multilingual models encode linguistic properties of Indic languages effectively, such encodings may not be robust to input text perturbations for particular models. Lack of such robustness may make some models less reliable than others for real-world applications. Although there exist many such studies~\citep{wang2021adversarial,jin2020bert,li2020bert, garg2020bae,sanyal-etal-2022-robustlr,neerudu2023robustness} on robustness analysis of Transformer-based models, they focus on English, and on downstream tasks. Robustness analysis for prediction of linguistic properties in a multilingual setting for Indic languages is under explored. Hence, we systematically study how various multilingual Transformer-based models may differ in their robustness for different linguistic properties with respect to different kinds of input text perturbations across languages. We provide detailed related work in Appendix~\ref{app:relatedwork}, focusing on three aspects: (i) probing in non-English and multilingual contexts, (ii) differences from English-centric findings, and (iii) critiques of probing methodology and recent advances.

We analyze 7 universal multilingual language models, each pretrained on data spanning $\sim$100+ languages, with only a small amount of Indic language data. These include mBERT~\citep{pires2019multilingual}, XLM-R~\citep{conneau2019unsupervised}, InfoXLM~\citep{chi2020InfoXLM}, mGPT~\citep{shliazhko2024mgpt}, XGLM~\citep{lin-etal-2022-shot}, BLOOM~\citep{workshop2022bloom}, and mT5~\citep{xue2020mt5}. Additionally, we examine 2 Indic-specific models, IndicBERT~\citep{kakwani2020indicnlpsuite} and MuRIL~\citep{khanuja2021muril}, which are trained using corpora for Indic languages along with English. While it is expected that these Indic-specific models would be better at encoding and robustness for Indic language input, are there some linguistic properties which are better encoded by universal models? Are the universal models more robust to particular perturbation types? How effectively and robustly are English language properties encoded by these universal and Indic-specific models?

To perform such detailed analyses, we curate a novel multilingual dataset, \textsc{IndicSentEval}, from resources generated by the ``Indian Language Machine Translation'' (ILMT) initiative.
%\footnote{\url{https://tinyurl.com/mr7rxy6x}\label{ilmtfn}}.
%https://tdil-dc.in/index.php?option=com_vertical&parentid=74&lang=en}}. 
%https://tinyurl.com/6hhdcce9
This dataset contains information about three types of linguistic properties per Indic language: surface, syntactic and semantic. To investigate encoding and robustness ability of multilingual models, we design probing tasks for prediction of each property. 
%Thus, there are three kinds of probing tasks: surface, syntactic and semantic. 
The surface task probes whether the model learns a representation which is predictive of sentence length (SentLen). Syntactic tasks test for sensitivity to word order, i.e., bigram shift (BShift) and the depth of the syntactic tree (TreeDepth). Semantic tasks check for the subject and the direct object number in the main clause (SubjNum and ObjNum, respectively). Tasks mentioned so far were discussed in English focused studies too. However, morphology for Indic languages is significantly different from English, primarily in regard to the main verb used in a sentence. 
% In English, verbs undergo conjugation based on the subject and tense. For instance, ``I walk,'' ``he walks,'' and ``she walked.'' In contrast, Indic languages often feature more intricate conjugation systems, wherein verbs change according to factors like number, gender, and person. 
This prompts us to expand our analysis to encompass three additional semantic probing tasks related to the main verb within the sentence: verb gender (VerbGen), verb number (VerbNum), and verb person (VerbPer). %Moreover, we also investigate the robustness of the multilingual models using 13 different perturbations for these 8 probing tasks. 
Fig.~\ref{fig:conceptualDiagram} shows the conceptual diagram of our study.

Overall, the main contributions of this paper are as follows.
(1) We perform an extensive study of the degree to which 9 multilingual Transformer-based models capture 8 linguistic properties across 6 Indic languages. (2) We contribute a novel dataset, \textsc{IndicSentEval}, with $\sim$47K sentences across the 6 languages. (3) We find that Indic-specific models like MuRIL and IndicBERT best capture linguistic properties for Indic languages, while universal models like mBERT, InfoXLM, mGPT and BLOOM show mixed results across properties. 
(4) Surprisingly, our robustness analysis with respect to 13 text perturbations shows that universal multilingual models (InfoXLM, BLOOM, mGPT, XGLM and mT5) demonstrate higher resilience to perturbations compared to BERT-like models (mBERT, IndicBERT and MuRIL). 
%We make our code and dataset publicly available\footref{datafootnote}.
%\item We also find that 
%We find that mBERT and MuRIL best capture language hierarchy for Indic languages (lower layers encode surface properties; middle layers encode syntactic properties while higher layers encode semantic properties). 
%(surface-lower, syntactic-middle, and semantic-higher) 
%while XLM-R, InfoXLM, and IndicBERT models have different linguistic structures for different languages. 
%Verbs and the order of words are the most important signals that help the models to encode linguistic structures. Tree depth is the most sensitive to any perturbation while subject and object number are the most resilient.
%\end{itemize}

%The remainder of the paper is organized as follows. 
%First, we discuss the set of probing tasks we created and analyze. Next, we discuss details of our \textsc{IndicSentEval} dataset. Then, we list the set of text perturbations for robustness analysis. Then, we present details of our methodology including discussion on multilingual models, dataset splits and evaluation metrics. Further, we present both probing and perturbation results. Lastly, we conclude with a brief summary. 

%\vspace{-0.1cm}
\section{\textsc{IndicSentEval} Dataset}
\label{sec:dataset}
%\vspace{-0.15cm}
We curate the \textsc{IndicSentEval} dataset from resources generated by the ILMT initiative, %\footref{ilmtfn}, 
which serves as an Indic language counterpart to SentEval~\citep{conneau2018you} and offers labeled data for the eight probing tasks. We utilize the morph and chunk level Indic languages data~\citep{tandon2017unity,bhatt2009multi,xia2008towards} available in Shakti Standard Format (SSF)~\citep{bharati2007ssf,bharati1995natural}. SSF is a highly readable representation for storing Indic language data with linguistic annotations. Fig.~\ref{fig:SSFExample2} in Appendix~\ref{app:ssfFormat} shows an example of a Hindi sentence in SSF format. 
%The dataset has been derived from SSF data made available by the LTRC lab\footnote{Hindi and Telugu SSF data is taken from \url{https://ltrc.iiit.ac.in/showfile.php?filename=downloads/lingResources/newreleases.html}; data for remaining languages is taken from \url{https://ltrc.iiit.ac.in/showfile.php?filename=downloads/kolhi/}} at IIIT Hyderabad, India.

\noindent\textbf{Probing Tasks.}
Probing tasks~\citep{adi2016fine,hupkes2018visualisation,jawahar2019does,mohebbi2021exploring} help unpack the linguistic features possibly encoded in neural language models. These probing tasks are formulated as prediction tasks and focus on several aspects of sentence structure. We experiment with eight probing tasks to evaluate how effectively multilingual models encode linguistic properties across six Indic languages: Hindi (\texttt{hi}), Telugu (\texttt{te}), Marathi (\texttt{mr}), Kannada (\texttt{kn}), Urdu (\texttt{ur}), and Malayalam (\texttt{ml}). These eight probing tasks are grouped into three categories: surface (SentLen), syntactic (BShift and TreeDepth), and semantic (SubjNum, ObjNum, VerbGen, VerbPerson and VerbNumber). We selected these tasks because they cover different aspects of language and require different levels of abstraction and generalization. These tasks involve 3 binary and 5 multi-class classification problems. The specifics of the initial five probing tasks are thoroughly outlined in~\citep{conneau2018you} as well. We also provide brief descriptions for each probing task in Appendix~\ref{probing_tasks}, with a summary of class labels for each task in Table~\ref{tab:taskDetails}. 

\begin{table}[t]
\scriptsize
\centering
\resizebox{\columnwidth}{!}{\begin{tabular}{|l|c|p{0.75\columnwidth}|} \hline 
\textbf{Task} & \textbf{$|$C$|$} & \textbf{Labels} \\ \hline 
SentLen& 8 &  (0-5),(6-8),(9-12),(13-16),(17-20),(21-25),(26-28),(29-32)\\ \hline
% WordContent&$L_i$&One of the $L_i$ words in language $i$\\\hline
TreeDepth  & 5 & (0-2),(3-5),(6-8),(9-11),(12-20) \\ \hline   
BShift& 2 & 0,1 \\ \hline   
SubjNum & 2 & singular, plural \\ \hline   
ObjNum& 2 &  singular, plural \\ \hline
VerbGen& 4 & masculine, feminine, neutral, any \\ \hline    
VerbNum & 3 & singular, plural, any\\ \hline   
VerbPer & 7 & $1^{st}$ person, $2^{nd}$ person, $3^{rd}$ person, $1^{st}$ person honorific, $2^{nd}$ person honorific, $3^{rd}$ person honorific, any \\ \hline   
\end{tabular}}
%\vspace{-0.25cm}
\caption{Probing task details: number of classes ($|C|$) and class labels.}
\label{tab:taskDetails}
\end{table}

\noindent\textbf{\textsc{IndicSentEval} curation details.} For each property, we gather data per language as follows. 
 \noindent(1) \textit{SentLen:} We iterate through all the nodes in the SSF format representation of the sentence and count number of words in each chunk. 
% \noindent(2) \textit{Word Content:} For this task, we follow a two-step approach. First, we sample vocabulary words from the mid-frequency range (f = 25-100). In the second step, we select sentences from the dataset which contain these sampled words ensuring that each sentence has only one of the sampled words occurring exactly once. The resulting dataset, is then employed for an $L_i$-way classifier problem. The value of $L_i$ depends on the number of words falling within the mid-frequency range for each specific language. $L_i$ values for each language are as follows: Hindi: 781, Telugu: 34, Marathi: 626, Malayalam: 188, Kannada: 416, and Urdu: 153. 
(2) \textit{TreeDepth:} We utilize the data from the dependency tree to perform a traversal, specifically employing breadth first search. This traversal enables us to calculate the tree depth of the sentence. 
(3) \textit{BShift:} For this task, we randomly (probability=0.2) select the sentences from the dataset, and then a randomly selected bigram (equal probability for all bigrams) is inverted. Sentences with inverted bigrams are marked as 1, and the rest as 0. 
(4) \textit{SubjNum/ObjNum:} We identify the subject/object (a noun that can be singular or plural) of a sentence using the assigned semantic roles in the SSF format, and use the NN/NNS annotations.
(5) \textit{Verb Gender/Person/Number:} We first locate the chunk containing the main verb using the annotated chunk label from the SSF format. Then, we extract the gender/person/number information from the annotated morph output. Detailed statistics of number of samples across 6 Indic languages for 8 probing tasks are provided in Tables~\ref{tab:ProbingTrainTestSplit} and~\ref{tab:vocab} in Appendix~\ref{app:dataStats}. Further, Table~\ref{tab:ProbingtaskExamples} in  Appendix~\ref{probing_tasks} shows examples for each probing task per language.

% \textsc{IndicSentEval} is a collection of eight probing tasks in six Indic languages. The detailed statistics of the probing task and its corresponding labels are presented in Table~\ref{Task and Label Details}, 

%https://docs.google.com/drawings/d/1ntn8r-8_m45vEhTjbZjT4pdY9g3x0ryWNDjUhG-inbE/edit?usp=sharing

% \noindent\textbf{\textsc{IndicSentEval} probing statistics}: Detailed statistics of samples across six Indic languages for eight probing tasks are provided in the Table~\ref{tab:ProbingTrainTestSplit}. 

\section{Text Perturbation Analysis}
\label{text-perturbations}
%\vspace{-0.1cm}
While probing reveals what linguistic features are present in representations, perturbation-based robustness tests address a different question: ``how robust is a model to various types of noise and whether it can still understand and process the core meaning of the text despite the introduced variation?''
Since our current IndicSentEval dataset lacks noise examples, hence, in this study, we conducted our perturbation analysis on the input dataset to evaluate its robustness.
% Probing tasks can help us understand the extent to which various multilingual models encode different linguistic properties. But another interesting question is – which particular words (such as nouns, verbs and others) in the input sentence help models encode these structures? 
To answer this question, we experiment with three different categories of perturbations: AppendR,
 DropText and Positional. We chose these perturbations because they simulate types of noise found in real datasets by introducing different degrees of noise and variation in the input text. Particularly, we experiment with the following text perturbations.

\noindent\textbf{AppendR.} We append a random (R) phrase to original sentence. This mimics real scenarios where additional, irrelevant data is included in text input.

\noindent\textbf{DropText.} DropText perturbations reflect situations where critical information is missing or only certain types of words are retained, which is common in incomplete or corrupted datasets. This includes \textit{DropNV} (dropping words based on their part-of-speech tag, specifically both nouns (N) and verbs (V)), \textit{DropN} (dropping all nouns), \textit{DropV} (dropping all verbs), \textit{DropRN} (dropping  one random noun), \textit{DropRV} (dropping one random verb), \textit{KeepNV} (dropping all words except nouns and verbs), \textit{KeepN} (dropping all words except nouns), and \textit{KeepV} (dropping all words except verbs). DropText perturbations are designed to provide deeper insights into word-level attention mechanisms within models, specifically aiming to determine whether models focus more on objects (nouns), actions (verbs), or the contextual elements surrounding these key parts-of-speech.

\noindent\textbf{Positional.} This includes \textit{DropF/DropL/DropFL} (replacing first/last/both words by ``[UNK]'' to maintain the original phrase length)
%, \textit{DropFL} (replacing first and last word  by ``[UNK]'' to maintain the original phrase length), \textit{DropL} (replacing last word  by ``[UNK]'' to maintain the original phrase length), 
and \textit{Shuffle} (randomly shuffling the words in a sentence). These position-based text perturbations help us understand the extent to which words at specific positions (first/last) or relative positions impact the language structure encoding capabilities of various models.

Overall, these text perturbations help in understanding the contribution of specific word types and sentence structures to the encoding capabilities of multilingual models.
%Next, we design text perturbation tasks where we retain/keep one or more groups of these words in the input sentence. \textcolor{red}{these are the insights related to the three tables, need to improve these sentences}
%We hope that the results will help us understand which word groups are critical for different (model, language) pairs to encode a particular language structure. Yet another set of perturbations are based on position of words in the input sentence. 
%These perturbations help us understand the extent to which words at specific positions (first/last) or relative positions impact the language structure encoding capabilities of various models.
%\end{enumerate}
Tables~\ref{fig:Sample of Telugu Perturbations}-\ref{fig:Sample of Urdu Perturbations} in Appendix~\ref{app:perturbationExamples} display examples of perturbations for each language.

\begin{figure*}[!t]
    \centering
      \includegraphics[width=0.8\linewidth]{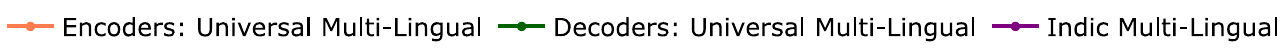} \\
      \begin{minipage}{\textwidth}
      \centering
      \includegraphics[width=\linewidth]{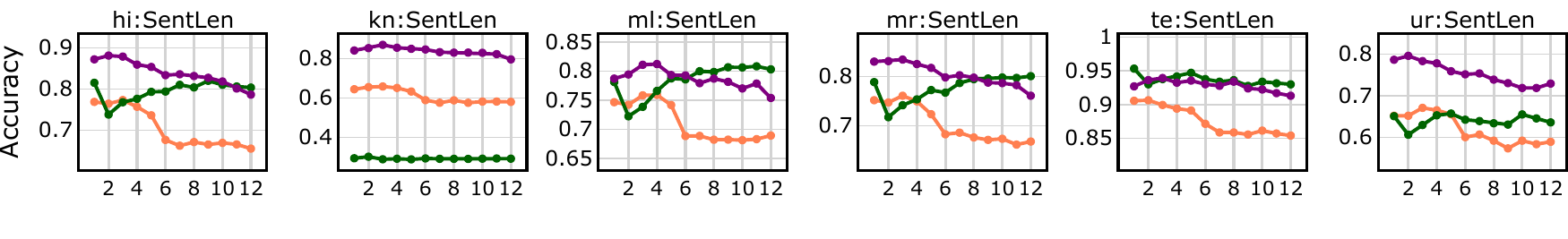}
      %\vspace{-0.4cm}
      \end{minipage}
\begin{minipage}{\textwidth}
\centering
      \includegraphics[width=\linewidth]{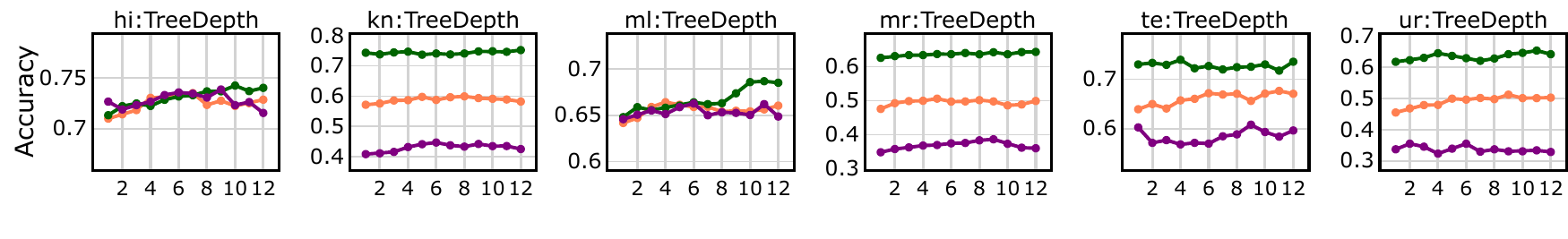} 
      %\vspace{-0.4cm}
      \end{minipage}
\begin{minipage}{\textwidth}
\centering
      \includegraphics[width=\linewidth]{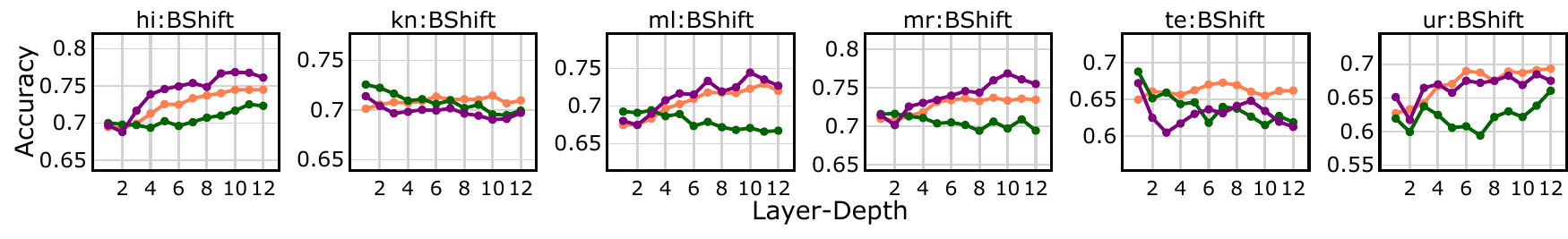} 
      \end{minipage} 
%\vspace{-0.45cm}
\caption{Probing task results: Layerwise accuracy comparisons between various multilingual representations on surface (top row) and syntactic (bottom two rows) probing tasks. We report the layerwise probing accuracies for individual multilingual models in Figs.~\ref{fig:hi_probing_tasks} to~\ref{fig:ur_probing_tasks} in Appendix~\ref{probing_results}.}
\label{fig:surface_syntactic_tasks_average} 
\end{figure*}

%\vspace{-0.1cm}

\section{Methodology}
\label{sec:methodology}
%\vspace{-0.1cm}
\noindent\textbf{Multilingual language models.}
%We extract sentence representations from seven multilingual language models.
We experiment with nine multilingual Transformer-based models (listed in Table~\ref{tab:modelDetails} in Appendix~\ref{app:modelDetails}). First seven have been trained across 100+ languages; IndicBERT and MuRIL support 12 and 17 Indic languages, respectively. 
Representations are extracted from the encoder layers of mBERT-base, IndicBERT-base, mT5-base, XLM-R, InfoXLM and MuRIL; and from the decoder layers of BLOOM, mGPT and XGLM. 
We use pretrained model checkpoints from Hugging Face~\citep{wolf2020transformers}. 

%Details of multilingual Transformer-based models used in this study. MLM=Masked Language Modeling, CLM=Causal LM, TLM=Translation LM, XLCO=Cross-Lingual Contrastive Learning. MSL=Maximum Sequence length.

\noindent\textbf{Probing tasks classifier.}
To evaluate each probing task using a multilingual  model representation, we use logistic regression~\citep{wright1995logistic} classifier with sentence representations as input and the probing task label as target. The base model is frozen. We use mean pooling across tokens to get the sentence representation. Details of the hyperparameters are reported in Appendix~\ref{app:modelDetails}.
%which is fed to the output softmax layer.

\noindent\textbf{Dataset splits.}
% To ensure a comprehensive evaluation of our probing tasks classifiers, we employed a rigorous methodology. 
% As shown in Table~\ref{ProbingTrainTestSplit}, we have separate datasets for training and testing for each language. For every probing task, we train the probing classifier on training data and report the accuracy for test dataset.
% split the data into training and testing sets, with 80\% of the data used for training and 20\% used for testing. 
We use a stratified five-fold cross-validation approach which involves splitting the dataset into five equal parts, where four parts are used for training and the remaining part is used for testing. This process is repeated five times, with each part used for testing once.
To report our results, we calculate the average performance of the model across all five folds. %This approach allows us to assess the model's performance consistently and reduces the impact of any random variability in the data.

% \noindent{\textbf{Evaluation Metrics}}
% To assess the effectiveness of machine learning models on probing and perturbation data, a commonly utilized classification metric is accuracy. This metric determines the percentage of correctly classified instances in relation to the total number of instances.

\noindent\textbf{Evaluation metrics.}
Similar to earlier studies~\citep{conneau2018you,jawahar2019does,mohebbi2021exploring}, for all the probing tasks, we use \emph{classification accuracy} as the evaluation metric. Let $A_{c}$ and $A_{p}$ be accuracy of a model on the clean and perturbed test sets, respectively. To evaluate the perturbation results for probing tasks, we use \emph{robustness score (RS)} defined as  $RS = 1 - \frac{A_{c} - A_{p}}{A_{c}}$. Typically, \emph{RS} of a model ranges between 0 and 1 where 0 indicates that the model is not robust, and 1 indicates that the model is completely robust. Score > 1 suggests that the model's performance improves when the perturbation is applied.

%\vspace{-0.1cm}
\section{Experimental Results}
\label{sec:expts}
%\vspace{-0.1cm}
%Here, we examine the encoding of linguistic information by multilingual Transformer models across six Indic languages. We present the findings using \textsc{IndicSentEval} dataset to shed light on the linguistic properties that contribute to, or do not contribute to, a hierarchy of language structure.  
We measure probing accuracy independently for each multilingual model, within each layer separately. Along with six Indic languages, we measure the probing accuracy for English language across all multilingual models and compare the findings for Indic languages against English. Further, we investigate the robustness of these multilingual models by perturbing the input sentences. %and identifying the underlying properties that result in a linguistic hierarchy among the six Indic languages. 
% To evaluate the sentence-level representations of mBERT ~\cite{pires2019multilingual}, IndicBERT ~\cite{kakwani2020indicnlpsuite}, XLM-R ~\cite{conneau2019unsupervised}, InfoXLM~\cite{chi2020InfoXLM}, and MuRIL ~\cite{khanuja2021muril}, we use the datasets created for the evaluation tasks and perform layer-wise probing. In all experiments, we use a single linear-layer probing classifier, initialized with identical weight s, for all the tasks and models. for evaluating the robustness experiments, \textcolor{red}{we used metrics to compute robustness.}. We access the pre-trained models through Hugging Face’s transformers library, and the probing classifier is trained using the PyTorch library. Since all evaluation tasks are structured as classification problems, we track the accuracy scores to assess the performance of the models (refer to Fig1).

%\vspace{-0.1cm}
\subsection{Probing Results}
%\vspace{-0.1cm}
\noindent\textbf{How effectively do multilingual models encode hierarchy of linguistic structure for Indic languages?}
We assess the linguistic structure by contrasting universal models trained on 100+ languages with those specifically trained on Indic languages only.
Unless otherwise specified, the results presented in the main paper reflect an average accuracy across encoder-based universal models, decoder-based universal models and Indic-specific models.
%Specifically, we examine encoder-based unmBERT, XLM-R, InfoXLM, mT5 and BLOOM as universal multilingual models. For Indic multilingual models, we focus on IndicBERT and MuRIL.
%\noindent\textbf{Universal multilingual vs. Indic multilingual models}

\noindent\textbf{Surface-level tasks.}
We show the accuracy scores obtained for surface level task (i.e. SentLen) in the first row in Fig.~\ref{fig:surface_syntactic_tasks_average}. Analyzing the performance across languages, we observe the following patterns: (i) For encoder-based universal models as well as for Indic models, there is a trend of higher accuracy in the early (or lower) layers, which decreases in the later (or higher) layers. This pattern is expected, as surface-level tasks generally require minimal processing. (ii) Notably, the decoder-based universal multilingual models deviates from this trend, showing lower accuracy in the early layers and higher accuracy in the later layers. This seemingly unusual pattern in decoder-based universal models is actually intuitive because masked self attention in their autoregressive architectural design implies that only deeper layers in such models can effectively grasp the input length. (iii) Overall, among all the models, Indic models report best accuracy, while encoder-based universal models display poorer performance. This is likely because Indic models are specifically trained on Indic languages, making them more attuned to the nuances and idiosyncrasies of these languages. 
On the other hand, universal models, which are designed for universal applicability, might struggle with specific linguistic features unique to Indic languages. These features include script differences, morphological complexity, and visual factors such as orthography and word length. Tokenizers of universal models tokenize Indic language inputs to many tokens with little correlation with actual input length in words.

We present the individual model-specific results across languages in  Figs.~\ref{fig:hi_probing_tasks} to~\ref{fig:ur_probing_tasks} in Appendix~\ref{probing_results}. We observe that among all the models, IndicBERT reports the best accuracy, while universal multilingual model XLM-R displays poor performance. More detailed analysis is reported in Appendix~\ref{probing_results}.
%The surface level patterns are
%language-specific since they are influenced by visual factors such as orthography and word length, we observe poor performances in cross-lingual language model XLM-R or opposite trend in autoregressive BLOOM model.
%A more in-depth analysis of processing of the specific languages in question would be necessary.
%Also, the best scores are for Hindi and Telugu which could be because of their high representation in IndicBERT's pretraining corpus.

%For the word content probing task, except for Hindi and Telugu, the probing task accuracy is negligible across models for the remaining four Indic languages.

\begin{figure*}[!t]
    \centering
    \includegraphics[width=0.8\linewidth]{images/colorbar_avg.pdf} \\
    %\vspace{-0.1cm}
%\includegraphics[width=0.16\linewidth]{images/HI_subnum.pdf}\vspace{2pt}\hspace*{-0.22em}
\begin{minipage}{1.0\textwidth}
\centering
    \includegraphics[width=0.98\linewidth]{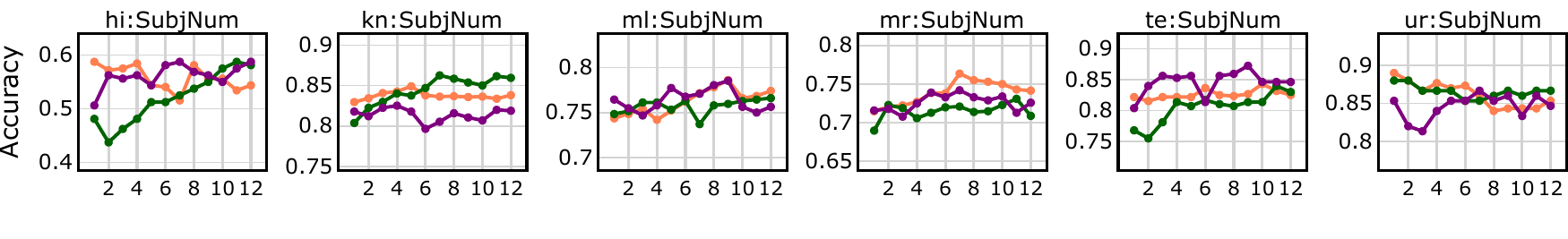} 
    %\vspace{-0.42cm}
\end{minipage}
\begin{minipage}{1.0\textwidth}
\centering
    \includegraphics[width=0.98\linewidth]{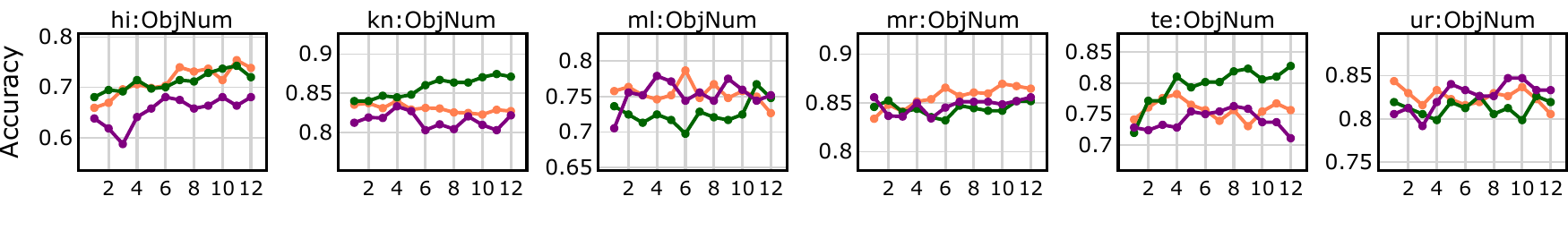} 
    %\vspace{-0.42cm}
\end{minipage}
\begin{minipage}{0.9\linewidth}
\centering
      \includegraphics[width=0.98\linewidth]{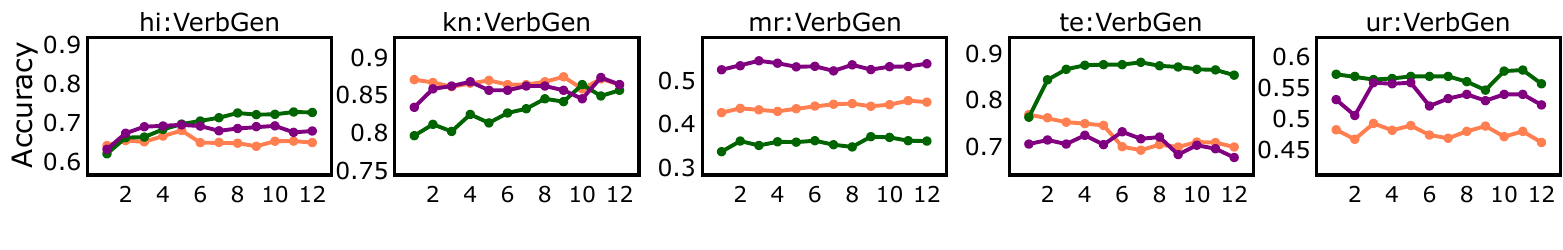}
      %\vspace{-0.4cm}
\end{minipage}
\begin{minipage}{0.9\textwidth}
\centering
\includegraphics[width=0.98\textwidth]{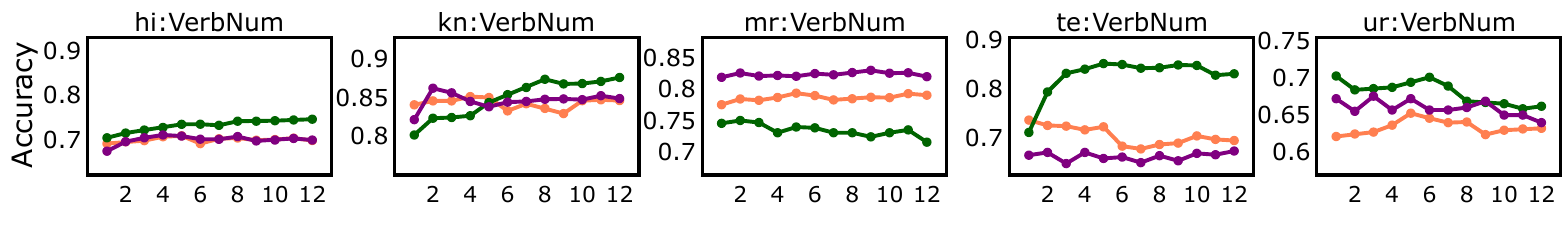} 
%\vspace{-0.4cm}
\end{minipage}
% %\vspace{-0.2cm}
\begin{minipage}{0.9\textwidth}
\centering
      \includegraphics[width=0.98\linewidth]{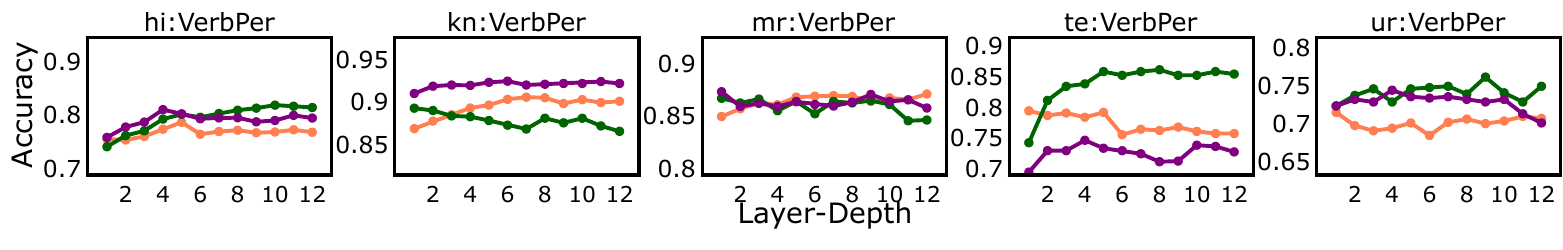}
\end{minipage}							   
%\vspace{-0.5cm}
\caption{Probing task results: Layerwise accuracy comparisons between various multilingual representations on semantic probing tasks. For Malayalam, there is an absence of SSF data
for the VerbGen, VerbPer, and VerbNum tasks. We report the layerwise probing accuracies for individual multilingual models in Figs.~\ref{fig:hi_probing_tasks} to~\ref{fig:ur_probing_tasks} in Appendix~\ref{probing_results}.}
    \label{fig:semantic_tasks_average}
\end{figure*}

% \begin{figure*}[t]
%     \centering
%     \includegraphics[width=0.5\linewidth]{images/legend.PNG} \\
%       \includegraphics[width=0.16\linewidth]{images/HI_treedepth.pdf}\hspace*{-0.22em}
%       \includegraphics[width=0.16\linewidth]{images/KA_treedepth.pdf}\hspace*{-0.22em}
%       \includegraphics[width=0.16\linewidth]{images/ML_treedepth.pdf}\hspace*{-0.22em}
%       \includegraphics[width=0.16\linewidth]{images/MR_treedepth.pdf}\hspace*{-0.22em}
%       \includegraphics[width=0.16\linewidth]{images/TE_treedepth.pdf}\hspace*{-0.22em}
%       \includegraphics[width=0.16\linewidth]{images/UR_treedepth.pdf}

%       \includegraphics[width=0.16\linewidth]{images/HI_bshift.pdf}\hspace*{-0.22em}
%       \includegraphics[width=0.16\linewidth]{images/KA_bshift.pdf}\hspace*{-0.22em}
%       \includegraphics[width=0.16\linewidth]{images/ML_bshift.pdf}\hspace*{-0.22em}
%       \includegraphics[width=0.16\linewidth]{images/MR_bshift.pdf}\hspace*{-0.22em}
%       \includegraphics[width=0.16\linewidth]{images/TE_bshift.pdf}\hspace*{-0.22em}
%       \includegraphics[width=0.16\linewidth]{images/UR_bshift.pdf}     
% \caption{Syntactic Tasks: The figure illustrates the results of layerwise probing task comparisons between multilingual representations like mBERT, IndicBERT, XLM, InfoXLM and MuRIL on syntactic-level probing tasks.}
%     \label{fig:syntactic_tasks}
% \end{figure*}

\noindent\textbf{Syntactic tasks.}
% In the bottom two rows of Fig.~\ref{fig:surface_syntactic_tasks_average}, we display the accuracy scores for syntactic tasks. For TreeDepth, we observe that probing accuracy tends to be higher in the middle layers for various (model, language) combinations. This trend is particularly notable in both encoder-based universal models and Indic models. Moreover, this pattern is consistent for three languages: \texttt{hi}, \texttt{kn} and \texttt{ml}. However, decoder-based multilingual models do not exhibit any clear layer-wise trend, and the same applies to the other three Indic languages. This suggests a model-specific and language-specific affinity in handling syntactic complexity, where certain models are more adept at processing syntactic information in specific layers, and this proficiency varies across different languages.
In the bottom two rows of Fig.~\ref{fig:surface_syntactic_tasks_average}, we display the accuracy scores for syntactic tasks. For TreeDepth, we observe that probing accuracy tends to be higher in the middle layers for various (model, language) combinations. This trend is particularly notable in both encoder-based universal models and Indic models. Moreover, this pattern is consistent for three languages: \texttt{hi}, \texttt{kn} and \texttt{ml}. However, decoder-based multilingual models do not exhibit any clear layer-wise trend, and the same applies to the other three Indic languages. Encoders, with their bidirectional attention, might be inherently better at capturing hierarchical structures, while decoders, with their unidirectional attention, might excel in tasks requiring sequential processing. The lack of a clear trend in decoder-based multilingual models might indicate that these models distribute syntactic processing more evenly across layers. The varying performance across languages underscores the importance of considering linguistic diversity in model training.

% In contrast to TreeDepth, for BShift task, we generally observe higher probing accuracy in the later layers for both encoder-based universal and Indic models across various languages. Notably, decoder-based universal models exhibit a decreasing trend in accuracy for \texttt{kn}, \texttt{ml} and \texttt{te}, while showing an increasing trend for \texttt{hi} and \texttt{ur}. This suggests that different models' layers may specialize in different types of syntactic processing, with some models better handling tasks like BShift in their later layers.
% Overall, when comparing performance across models and tasks, decoder-based universal models stand out for their superior accuracy in capturing tree depth information across different languages. Conversely, Indic models show notable proficiency in BShift. These distinctions highlight how different models may be better suited for different types of syntactic analysis. These observations contribute to a deeper understanding of how various multilingual models process syntactic information, demonstrating both model-specific and language-specific trends and capabilities in linguistic tasks.
In contrast to TreeDepth, for BShift task, we generally observe higher probing accuracy in the later layers for both encoder-based universal and Indic models across various languages. Notably, decoder-based universal models exhibit a decreasing trend in accuracy for \texttt{kn}, \texttt{ml} and \texttt{te}, while showing an increasing trend for \texttt{hi} and \texttt{ur}. This suggests that languages with different syntactic structures may require different layers to process syntactic tasks effectively. 

Overall, when comparing performance across models and tasks, decoder-based universal models stand out for their superior accuracy in capturing tree depth information across different languages. Conversely, Indic models show notable proficiency in BShift. These findings highlight the need for comprehensive evaluation across multiple tasks and languages to understand model capabilities fully. Evaluating models on diverse syntactic tasks can reveal strengths and weaknesses that might not be apparent from a single task. 
These observations contribute to a deeper understanding of how various multilingual models process syntactic information, demonstrating both model-specific and language-specific trends and capabilities in linguistic tasks.

From the individual model specific results in Figs.~\ref{fig:hi_probing_tasks} to~\ref{fig:ur_probing_tasks} in Appendix~\ref{probing_results}, we observe that among 
all the models, InfoXLM and mT5 stand out for their superior accuracy in capturing tree depth information across different languages. Conversely, Indic model MuRIL shows notable proficiency in BShift. More detailed analysis is reported in Appendix~\ref{probing_results}.
%for some languages, XLM-R works better while for other languages, MuRIL gives best results.
%(iii) 
%Overall, we conclude that multilingual Transformer models process syntactic information from middle to later layers. Also, there is no one best model that works across all six languages. 

\noindent\textbf{Semantic tasks.}
We plot the accuracy scores obtained for semantic tasks in Fig.~\ref{fig:semantic_tasks_average}: SubjNum and ObjNum in the first 2 rows, and VerbGen, VerbNum and VerbPer in the last 3 rows. The last three rows do not have results for Malayalam (\texttt{ml}) since we do not have labeled data for \texttt{ml} for those tasks.

From SubjNum and ObjNum results, we make the following observations. For decoder-based universal models, we observe an increasing trend from lower to higher layers for \texttt{hi}, \texttt{kn} and \texttt{te}. Dravidian languages often have more complex morphological systems for marking plurality, with a variety of suffixes and sometimes even changes in the noun stem itself. Hence, this increasing trend makes sense for \texttt{kn} and \texttt{te}.
%This suggests that for these languages, the models become more proficient in handling semantic tasks related to SubjNum and ObjNum as we move to the higher layers. %Within the Indic models, specifically MuRIL, this increasing trend is also evident for the same three languages (Hindi, Marathi, and Telugu). This consistency across both universal and Indic models for these languages indicates a shared pattern in how these models process semantic information. 
When considering other languages and models, we note that both encoder-based and Indic models exhibit an increasing trend for the ObjNum task, specifically for \texttt{hi} and \texttt{mr}. This highlights the variability in model performance based on the language. Interestingly, the middle layers of the encoder-based universal models and Indic models show higher probing accuracy for languages such as \texttt{hi}, \texttt{ml}, \texttt{mr} and \texttt{te} in the SubjNum task. %This indicates that, for these languages, the mid-level layers are more effective in semantic processing related to SubjNums. 
%Similar to its performance in surface and syntactic tasks, decoder-based universal models exhibit lower accuracy and lacks a discernible trend when it comes to capturing semantics.
% We make the following observations for the SubjNum and ObjNum tasks: (1) An increasing trend from lower to higher layers observed in universal multilingual (mBERT, InfoXLM, and BLOOM) for languages Hindi, Marathi, and Telugu. In Indic models, we observe this pattern in MuRIL for same three languages. Considering other languages and models, Indic BERT has increasing trend for ObjNum for Urdu. We also observe that middle layers display higher probing accuracy for languages such as Hind and Malayalam for SubjNum.
%MuRIL has the highest probing accuracy although it performs the worst for kn ObjNum task.\\
%(2) Finding SubjNum in Hindi and ObjNum in Malayalam is difficult. (3) In general, later layers seem to show better results compared to earlier or middle layers.

For the verb-related tasks, decoder-based universal models perform the best for most languages and tasks. Also, in most cases, the last layer is the most predictive, except in \texttt{mr} and \texttt{ur} for gender, number and person detection, where initial layers provide better results. This indicates that gender, number and person detection in \texttt{mr} and \texttt{ur} is straightforward and does not need deep processing. We observe a decreasing trend for encoder-based universal models for language \texttt{te}. For the Indic models, we observe an increasing trend for two languages \texttt{hi} and \texttt{kn} across tasks. Overall, both encoder-based universal and Indic models show mixed trend in performance across layers.

From the individual model specific results in  Figs.~\ref{fig:hi_probing_tasks} to~\ref{fig:ur_probing_tasks} in Appendix~\ref{probing_results}, we observe that XLM-R exhibits lower accuracy and lacks a discernible trend when it comes to capturing semantics. MuRIL has the highest probing accuracy although it performs the worst for \texttt{kn} ObjNum task. More detailed analysis is reported in Appendix~\ref{probing_results}.

\noindent\textbf{Comparison of encoding performance of multilingual models for linguistic properties for English vs Indic languages.}
We conduct probing for English  across nine multilingual models for five tasks: SentLen, TreeDepth, BShift, SubjNum, and ObjNum using SentEval dataset~\citep{conneau2018you}. Fig.~\ref{fig:appendixenglishresults} in Appendix~\ref{probing_results} reports the probing accuracy. Surprisingly, across all multilingual models, we observe that surface features show a decreasing trend from lower to higher layers. For syntactic TreeDepth feature probing accuracy is higher in the middle layers, while BShift has an increasing trend from lower to higher layers. Finally, the semantic tasks SujbNum and ObjNum are best encoded in the later layers. This implies that English is encoded in the same manner in both universal and Indic multilingual models, whereas Indic languages show mixed results across models.

\noindent\textbf{Overall insights from probing experiments.} While Indic-specific models like MuRIL and IndicBERT are likely the best at capturing language properties within the realm of Indic languages due to their targeted training, both encoder and decoder-based universal models like mBERT, InfoXLM, BLOOM and mGPT show mixed results. Their broader training might enable them to capture more general properties across many languages, but they may lack a deep understanding specific to each language, particularly those less represented in their training corpus. The effectiveness of these models thus depends on specific linguistic features and tasks, as well as the range of languages being considered.
% \noindent\textbf{Overall insights for Indic languages} 
mBERT and MuRIL capture linguistic features for \texttt{hi} very well. For \texttt{mr} and \texttt{te}, mT5 and MuRIL show better accuracy for surface, syntactic and semantic tasks. This is in line with the fact that \texttt{hi}, \texttt{mr}, and \texttt{te}  are better represented in pretraining datasets for these models. %, would typically be higher in this hierarchy, reflecting the models' greater proficiency in processing their structures.

\begin{table}[t]
\begin{minipage}{0.48\textwidth}
    \scriptsize
\centering
\begin{tabular}{|l|c|c|c|c|c|c|}
\hline
& \textbf{\texttt{hi}} & \textbf{\texttt{kn}} & \textbf{\texttt{ml}} & \textbf{\texttt{mr}} & \textbf{\texttt{te}} & \textbf{\texttt{ur}} \\ \hline
\textbf{mBERT} & 0.794 & 0.736 & 0.583 & 0.626 & \textbf{0.980}  & 0.715 \\ \hline
\textbf{IndicBERT} & 0.807 & 0.777 & 0.576 & 0.662 & 0.921 & \textbf{0.781} \\ \hline
\textbf{XLM-R}  & 0.883 & 0.883 & 0.675 & 0.692 & \textbf{0.981} & \textbf{0.865} \\ \hline
\textbf{InfoXLM} & \textbf{0.921} & \textbf{1.093} & 0.698 & 0.702 & 0.900 & 0.662 \\ \hline
\textbf{MuRIL} & 0.778 & 0.728 & 0.575 & 0.604 & \textbf{0.990}  & 0.704 \\ \hline
\textbf{BLOOM} & \textbf{0.957} & \textbf{0.903} & \textbf{0.731} & \textbf{0.742} & 0.966 & 0.707 \\ \hline
\textbf{mT5} & \textbf{0.961} & 0.790 & \textbf{0.773} & \textbf{0.767} & 0.952  & \textbf{0.757} \\ \hline
\textbf{mGPT} & 0.950 & \textbf{1.040} & \textbf{0.728} & \textbf{0.736} & 0.946  & 0.723 \\ \hline
\textbf{XGLM} &0.954  &0.972  &0.724  &0.730  & 0.956  & 0.715 \\ \hline
\end{tabular}
%\vspace{-0.2cm}
\caption{Comparison of robustness scores on probing tasks: multilingual models across languages, averaged across layers, highlighting top-3 scores.}
\label{tab:robustness_models_languages}
\end{minipage}
\hfill
\begin{minipage}{0.48\textwidth}
    \centering
\scriptsize
\begin{tabular}{|l|l|l|l|l|l|l|l|}
\hline
\multicolumn{1}{|c|}{} & \multicolumn{1}{c|}{\textbf{Sent}} & \multicolumn{1}{c|}{\textbf{Tree}}  & \multicolumn{1}{c|}{\textbf{Subj}} & \multicolumn{1}{c|}{\textbf{Obj}} & \multicolumn{1}{c|}{\textbf{Verb}} & \multicolumn{1}{c|}{\textbf{Verb}} & \multicolumn{1}{c|}{\textbf{Verb}} \\
% \multicolumn{1}{|c|}{} & \multicolumn{1}{c|}{\textbf{Sent Len}} & \multicolumn{1}{c|}{\textbf{Tree Depth}}  & \multicolumn{1}{c|}{\textbf{Subj Num}} & \multicolumn{1}{c|}{\textbf{Obj Num}} & \multicolumn{1}{c|}{\textbf{Verb Gen}} & \multicolumn{1}{c|}{\textbf{Verb Num}} & \multicolumn{1}{c|}{\textbf{Verb Per}} \\
\multicolumn{1}{|c|}{\multirow{-2}{*}{\textbf{}}} & \multicolumn{1}{c|}{\textbf{Len}}  & \multicolumn{1}{c|}{\textbf{Depth}} & \multicolumn{1}{c|}{\textbf{Num}}  & \multicolumn{1}{c|}{\textbf{Num}} & \multicolumn{1}{c|}{\textbf{Gen}}  & \multicolumn{1}{c|}{\textbf{Num}}  & \multicolumn{1}{c|}{\textbf{Per}}  \\ 
\hline
    \textbf{mBERT} & 0.398 & 0.487 & 0.916 & 0.924 & 0.839 & 0.903 & \textbf{0.939} \\ \hline
\textbf{IndicBERT} & 0.364 & 0.504 & 0.928 & 0.931 & 0.874 & \textbf{0.947} & \textbf{0.969} \\ \hline
    \textbf{XLM-R} & 0.492 & 0.564 & \textbf{1.030} & \textbf{0.996} & \textbf{0.968} & \textbf{1.001} & \textbf{1.006} \\ \hline
\textbf{InfoXLM} & \textbf{0.855} & 0.836 & 0.895 & 0.916 & 0.738 & 0.808 & 0.926 \\ \hline
\textbf{MuRIL} & 0.385 & 0.471 & 0.900 & \textbf{0.935} & 0.833 & \textbf{0.906} & 0.918 \\ \hline
\textbf{BLOOM} & 0.604 & \textbf{0.870} & \textbf{0.933} & 0.930 & \textbf{0.892} & 0.849 & 0.899 \\ \hline
\textbf{mT5} & 0.582 & \textbf{0.905} & 0.915 & 0.931 & \textbf{0.898} & 0.853 & 0.914 \\ \hline
\textbf{mGPT} & \textbf{0.918} & \textbf{0.872} &\textbf{0.932}  &\textbf{0.936}  & 0.820  & 0.879 & 0.931\\ \hline
\textbf{XGLM} &  \textbf{0.761} & 0.865 & 0.930 & 0.925 &  0.834 & 0.864 &0.904 \\ \hline
\end{tabular}
%\vspace{-0.2cm}
\caption{Comparison of robustness scores on probing tasks: multilingual models vs. probing tasks, averaged across layers, highlighting top-3 scores.}
\label{tab:robustness_models_tasks}
\end{minipage}
\end{table}

%\vspace{-0.2cm}
\subsection{Perturbation Results}
%\vspace{-0.1cm}
%While the prior probing analyses identified which multilingual language models more effectively capture the hierarchical structure of language for six Indic languages, focusing on eight linguistic properties at individual layers, we would also like to understand the underlying properties that result in a linguistic hierarchy among the six Indic languages. For this purpose, 
We perform 13 different text perturbations to understand the contribution of specific word types and sentence structures to the encoding capabilities of multilingual language models\footnote{Perturbations do not make sense for BShift probing task.}. 
We analyze the impact of such text perturbations for every pair of (model, language), (model, probing task), (perturbation, model) and (language, probing task) in Tables~\ref{tab:robustness_models_languages},~\ref{tab:robustness_models_tasks},~\ref{tab:robustness_models_perturbations} and~\ref{tab:robustness_languages_tasks} respectively.

We report results for the layerwise perturbation analysis and other pairs like (perturbation, language) and (perturbation, probing task) in Tables~\ref{tab:Robustness-layerwiseanalysis},~\ref{tab:robustness_languages_perturbations} and \ref{tab:robustness_perturbations_tasks}. All these tables show weighted averages across marginalized dimensions.

\noindent\textbf{Which multilingual models are more robust to perturbations in Indic languages?}
%Table~\ref{tab:robustness_models_languages} presents the robustness performance of each model across six Indic languages. Notably, 
Table~\ref{tab:robustness_models_languages} shows that universal models like InfoXLM, BLOOM, mT5 and mGPT show greater resilience to perturbations in at least four languages. In contrast, the universal model (mBERT) and the Indic-specific models (IndicBERT and MuRIL) display a more significant accuracy drop across all the Indic languages. Accuracy drop for BERT-specific models is perhaps because cross-lingual transfer might be less effective, resulting in decreased accuracy compared to other multilingual models.

\noindent\textbf{Which multilingual models are more robust across probing tasks?}
Table~\ref{tab:robustness_models_tasks}  %presents the robustness performance of each multilingual model across seven probing tasks. Consistent with previous findings, 
shows that universal models have greater robustness compared to Indic models and mBERT. 
%Specifically, decoder-based models are more robust compared to encoder-based and Indic models. These results align with previous studies, such as those by \citet{neerudu2023robustness}, which observe that GPT-based models exhibit robustness in downstream fine-tuning tasks.
Additionally, surface and syntactic probing tasks are significantly impacted by perturbations compared to semantic properties.

\setlength{\tabcolsep}{0.2pt}
\begin{table}[!t]
\centering
\scriptsize
% \resizebox{\columnwidth}{!}{
\begin{tabular}{|l|c|c|c|c|c|c|c|c|c|}
\hline
\textbf{} & \multirow{2}{*}{\rotatebox{30}{\textbf{mBERT}}} & \textbf{Indic} & \multirow{2}{*}{\rotatebox{30}{\textbf{XLM-R}}}                & \textbf{Info} & \multirow{2}{*}{\rotatebox{30}{\textbf{MuRIL}}} & \multirow{2}{*}{\rotatebox{30}{\textbf{BLOOM}}} & \multirow{2}{*}{\rotatebox{30}{\textbf{mT5}}} & \multirow{2}{*}{\rotatebox{30}{\textbf{mGPT}}} &\multirow{2}{*}{\rotatebox{30}{\textbf{XGLM}}}\\ 
\textbf{} & & \textbf{BERT} &                & \textbf{XLM} & &  &  &  &\\ \hline
\textbf{AppendR} & 0.810 & \textbf{0.764} & 0.886 & 0.877 & 0.790 & 0.839 &0.837 & 0.938 & 0.888\\ \hline
\textbf{DropNV}  & \textbf{0.735} & \textbf{0.780} & \textbf{0.870} & \textbf{0.793} & \textbf{0.735}&\textbf{0.810} &\textbf{0.802} & \textbf{0.865} & \textbf{0.838}\\ \hline
\textbf{DropN}   & 0.819 & 0.847 & \textbf{0.876} & 0.865 & 0.808&0.826 &0.825 & 0.886 & 0.856\\ \hline
\textbf{DropV}   & \textbf{0.747} & \textbf{0.779} & 0.882 & \textbf{0.822} & \textbf{0.743} &0.831 &0.838 & 0.914 & 0.873\\ \hline
\textbf{DropF}   & 0.828 & 0.847 & 0.884 & 0.892 & 0.814 &0.847 &0.849 & 0.950 & 0.898 \\ \hline
\textbf{DropFL}  & 0.760 & 0.794 & 0.886 & 0.844 & 0.754 &0.849 & 0.859 & 0.952 & 0.901\\ \hline
\textbf{DropL}   & 0.768 & 0.796 & 0.887 & 0.848 & 0.759 &0.849 & 0.852 & 0.953 & 0.901\\ \hline
\textbf{DropRN}  & 0.820 & 0.845 & 0.883 & 0.890 & 0.811 &0.845 & 0.837 & 0.938 & 0.891\\ \hline
\textbf{DropRV}  & 0.768 & 0.793 & 0.886 & 0.849 & 0.760 & 0.847 & 0.851 & 0.949 & 0.898\\ \hline
\textbf{KeepNV}  & 0.812 & 0.842 & 0.890 & 0.853 & 0.803 & \textbf{0.817} & \textbf{0.813}& 0.881 & 0.849\\ \hline
\textbf{KeepN}   & \textbf{0.734} & \textbf{0.779} & 0.889 & \textbf{0.780} & \textbf{0.731} &0.823 & \textbf{0.817} & \textbf{0.872} & \textbf{0.847}\\ \hline
\textbf{KeepV}   & 0.824 & 0.866 & \textbf{0.849} & 0.851 & 0.825 &\textbf{0.820} & \textbf{0.809} & \textbf{0.858} & \textbf{0.839}\\ \hline
\textbf{Shuffle} & 0.812 & \textbf{0.746} & 0.889 & 0.884 & 0.798 &0.845 & 0.844 & 0.935 & 0.890\\ \hline
\end{tabular}
% }
%\vspace{-0.2cm}
\caption{Comparison of robustness scores on probing tasks: multilingual models vs. perturbation types, averaged across layers, highlighting top-3 scores.}
\label{tab:robustness_models_perturbations}
\end{table}
\setlength{\tabcolsep}{3pt}

\noindent\textbf{Which text perturbations have the greatest impact on multilingual models?}
From Table~\ref{tab:robustness_models_perturbations}, we observe: (i) Dropping both nouns and verbs has an adverse effect on all models. 
(ii) Similarly, dropping only verbs or retaining only nouns affects the performance drop across models.
Thus, eliminating nouns and verbs can 
lead to losing vital information necessary for accurate predictions. Discarding nouns or verbs can disrupt the syntactic coherence of text, making it more challenging for the model to comprehend and process.
%\textcolor{green}{perhaps the larger multilingual models are more robust as they rely less on language specific wordorder, etc, and other short-cuts instead?}

%We apply thirteen different perturbations divided into three categories and measure results on all seven models across six languages and seven probing tasks\footnote{Perturbations do not make sense for the other two probing tasks i.e., WC and BShift}. 

\begin{table}[t]
\centering
\scriptsize
\begin{tabular}{|l|l|l|l|l|l|l|l|}
\hline
\multicolumn{1}{|c|}{} & \multicolumn{1}{c|}{\textbf{Sent}} & \multicolumn{1}{c|}{\textbf{Tree}}  & \multicolumn{1}{c|}{\textbf{Subj}} & \multicolumn{1}{c|}{\textbf{Obj}} & \multicolumn{1}{c|}{\textbf{Verb}} & \multicolumn{1}{c|}{\textbf{Verb}} & \multicolumn{1}{c|}{\textbf{Verb}} \\
\multicolumn{1}{|c|}{\multirow{-2}{*}{\textbf{}}} & \multicolumn{1}{c|}{\textbf{Len}}  & \multicolumn{1}{c|}{\textbf{Depth}} & \multicolumn{1}{c|}{\textbf{Num}}  & \multicolumn{1}{c|}{\textbf{Num}} & \multicolumn{1}{c|}{\textbf{Gen}}  & \multicolumn{1}{c|}{\textbf{Num}}  & \multicolumn{1}{c|}{\textbf{Per}}  \\ \hline
\textbf{\texttt{hi}} & 0.436 & 0.338 & \textbf{1.104} & \textbf{1.009} & \textbf{0.931} & \textbf{0.951} & \textbf{0.972} \\ \hline
\textbf{\texttt{kn}} & \textbf{0.793} & \textbf{0.632} & 0.902 & 0.896 & 0.818 & 0.860 & 0.967 \\ \hline
\textbf{\texttt{ml}} & 0.362 & 0.367 & 0.859 & 0.897 & \textbf{0.968} & \textbf{1.001} & \textbf{1.006} \\ \hline
\textbf{\texttt{mr}} & 0.277 & 0.578 & 0.836 & \textbf{0.938} & 0.738 & 0.808 & 0.926 \\ \hline
\textbf{\texttt{te}} & \textbf{0.895} & \textbf{0.939} & \textbf{0.952} & \textbf{0.968} & \textbf{0.915} & \textbf{1.003} & \textbf{0.994} \\ \hline
\textbf{\texttt{ur}} & 0.232 & 0.580 & \textbf{0.949} & 0.935 & 0.738 & 0.837 & 0.875 \\ \hline
\end{tabular}
%\vspace{-0.2cm}
\caption{Comparison of robustness scores on probing tasks: Indic languages vs. probing tasks, averaged across layers, highlighting top-3 scores.}
\label{tab:robustness_languages_tasks}
\end{table}

\noindent\textbf{Which Indic languages are more robust across probing tasks?}
Table~\ref{tab:robustness_languages_tasks}  %displays the robustness performance of each language across seven probing tasks. We observe that 
shows that models exhibit greater robustness across probing tasks for \texttt{hi} and \texttt{te}, especially for semantic properties. Conversely, surface and syntactic properties are more affected by perturbations, except for similar language structure observed for \texttt{te} and \texttt{kn}. Additionally, languages such as \texttt{ur} and \texttt{mr} are more impacted by perturbations due to relatively lower training token counts compared to other languages.

\noindent\textbf{Which layers are more affected due to text perturbations for Indic languages?}
Table~\ref{tab:Robustness-layerwiseanalysis}, we observe: (i) across all Indic languages, early layers are impacted for surface properties. (ii) for TreeDepth syntactic property, middle layers are affected for all languages except \texttt{te}. (iii) Surprisingly, later layers are impacted more than early and middle layers. Regarding semantic properties, for \texttt{ur}, early to middle layers are impacted more than later layers, except for ObjNum. Similarly, for \texttt{kn}, SubjNum and ObjNum are impacted on middle layers; for \texttt{mr}, VerbNum  and VerbPer are affected. 
%Regarding semantic properties, results vary across Indic languages except for Hindi. Specifically, 

% Table~\ref{tab:Robustness-layerwiseanalysis} in the Appendix reports the layerwise perturbation analysis for each probing task for each Indic language. We make the following observations: (i) For surface property, early layers are impacted across all Indic languages. (ii) For syntactic property, TreeDepth, middle layers effected for all languages except Telugu. Surprisingly, later layers impacted more than ealry and middle layers.
% (iii) For semantic properties, we observe mixed results across Indic languages except Hindi. Speficifically, Urdu language has early to middle layers impacted than later layers except for ObjNum property. Similarly, for Kannada language, verb num and ObjNum impacted on middle layers, while verbnum and verb per impacted for Marathi language.

\noindent\textbf{Which text perturbations have the greatest impact on probing tasks across six Indic languages?}
Tables~\ref{tab:robustness_languages_perturbations} and~\ref{tab:robustness_perturbations_tasks}  show that dropping nouns and verbs significantly affects accuracy across all six languages. Specifically, dropping verbs impacts accuracy of three verb-based semantic tasks. Similarly, position perturbations, such as keeping nouns in specific positions, affect verb tasks, while keeping verbs or nouns affects surface and syntactic tasks.

%Comparison of robustness scores on probing tasks for six Indic languages vs. seven probing tasks. Top-3 highest robustness scores for each language across seven models are highlighted in bold.

\begin{table}[!t]
\scriptsize
\centering
\begin{tabular}{|l|l|l|l|l|l|l|}
\hline
 & \textbf{\texttt{hi}} & \textbf{\texttt{kn}} & \textbf{\texttt{ml}}& \textbf{\texttt{mr}} &\textbf{\texttt{te}} &\textbf{\texttt{ur}}\\ \hline
\textbf{SentLen} &1, 2, 3  & 4, 3, 2 & 3, 4, 1 & 1, 3, 2 & 1, 2 &  3, 1, 2\\ \hline
\textbf{TreeDepth} &7, 5, 6  & 5, 4, 9 & 4, 5, 3 & 4, 8, 9 & 10, 11, 12 & 3, 2, 5\\ \hline
\textbf{VerbGen} & 10, 8, 5 & 11, 4, 12 & - & 11, 10, 12 & 6, 8, 9 & 3, 8, 9\\ \hline
\textbf{VerbNum} & Equal & 3, 4, 5 & - & 2, 4, 5 & 10, 12 & 5, 4, 3 \\ \hline
\textbf{VerbPer} & Equal & Equal & - & 3, 4, 5 & 5, 11, 12 & 4, 5\\ \hline
\textbf{SubjNum} & N/A & Equal & 9, 8, 7 & 7, 10, 9 & 10, 11 & 1, 2\\ \hline
\textbf{ObjNum} & 11, 12 & 4, 3, 9 & Equal & 11, 10, 12 & 11, 12 & 10, 9, 11\\ \hline
\end{tabular}
%\vspace{-0.2cm}
\caption{Summary of the layerwise robustness analysis averaged across multilingual models, considering 13 perturbations. Each cell reports the most affected layers after text perturbations for each probing task and language. ``Equal'' denotes that all layers are affected, while `-' indicates the absence of a probing dataset for that particular language.}
\label{tab:Robustness-layerwiseanalysis}
\end{table}

\begin{table}[!t]
\centering
\scriptsize
\begin{tabular}{|l|l|l|l|l|l|l|}
\hline
\textbf{} & \textbf{\texttt{hi}} & \textbf{\texttt{kn}} & \textbf{\texttt{ml}} & \textbf{\texttt{mr}} & \textbf{\texttt{te}} & \textbf{\texttt{ur}} \\ \hline
\textbf{AppendR} & \textbf{0.788} & 0.854 & 0.615 & 0.666 & 1.000 & 0.743 \\ \hline
\textbf{DropNV}  & \textbf{0.812} & \textbf{0.826} & \textbf{0.605} & \textbf{0.664} & \textbf{0.833} & \textbf{0.740} \\ \hline
\textbf{DropN}   & 0.866 & \textbf{0.828} & \textbf{0.609} & 0.666 & 1.023 & 0.743 \\ \hline
\textbf{DropV} & \textbf{0.813} & 0.847 & 0.619 & 0.666 & \textbf{0.878} & \textbf{0.742} \\ \hline
\textbf{DropF} & 0.865 & 0.853 & 0.618 & 0.666 & 1.067 & \textbf{0.742} \\ \hline
\textbf{DropFL} & 0.847 & 0.851 & 0.618 & 0.666 & 0.906 & 0.746 \\ \hline
\textbf{DropL} & 0.848 & 0.855 & 0.616 & \textbf{0.665} & 0.908 & 0.751 \\ \hline
\textbf{DropRN} & 0.862 & 0.855 & \textbf{0.612} & 0.667 & 1.045 & 0.745 \\ \hline
\textbf{DropRV} & 0.842 & 0.856 & 0.618 & 0.666 & 0.913 & 0.748 \\ \hline
\textbf{KeepNV} & 0.848 & 0.833 & 0.619 & 0.673 & 1.022 & \textbf{0.741} \\ \hline
\textbf{KeepN} & \textbf{0.771} & \textbf{0.827} & 0.634 & 0.672 & \textbf{0.857} & 0.755 \\ \hline
\textbf{KeepV} & 0.870 & \textbf{0.828} & 0.678 & \textbf{0.540} & 1.021 & 0.746 \\ \hline
\textbf{Shuffle} & 0.845 & 0.854 & 0.615 & 0.666 & 0.935 & 0.747 \\ \hline
\end{tabular}
\caption{Comparison of robustness scores on probing tasks: Indic languages vs. perturbation types, averaged across layers, Lowest robustness values per column are highlighted in bold.}
\label{tab:robustness_languages_perturbations}
\end{table}

\begin{table}[!t]
\centering
\scriptsize
\resizebox{\columnwidth}{!}{\begin{tabular}{|l|l|l|l|l|l|l|l|}
\hline
\multicolumn{1}{|c|}{}                            & \multicolumn{1}{c|}{\textbf{Sent}} & \multicolumn{1}{c|}{\textbf{Tree}}  & \multicolumn{1}{c|}{\textbf{Subj}} & \multicolumn{1}{c|}{\textbf{Obj}} & \multicolumn{1}{c|}{\textbf{Verb}} & \multicolumn{1}{c|}{\textbf{Verb}} & \multicolumn{1}{c|}{\textbf{Verb}} \\ 
\multicolumn{1}{|c|}{\multirow{-2}{*}{\textbf{}}} & \multicolumn{1}{c|}{\textbf{Len}}  & \multicolumn{1}{c|}{\textbf{Depth}} & \multicolumn{1}{c|}{\textbf{Num}}  & \multicolumn{1}{c|}{\textbf{Num}} & \multicolumn{1}{c|}{\textbf{Gen}}  & \multicolumn{1}{c|}{\textbf{Num}}  & \multicolumn{1}{c|}{\textbf{Per}}  \\ \hline
\textbf{AppendR} & 0.525 & 0.573 & \textbf{0.899} & \textbf{0.898} & 0.885 & 0.932 & 0.964 \\ \hline
\textbf{DropNV} & \textbf{0.446} & 0.564 & \textbf{0.922} & 0.931 & \textbf{0.771} & \textbf{0.861}  & \textbf{0.916} \\ \hline
\textbf{DropN} & 0.472 & 0.573 & 0.934 & 0.924 & 0.921 & 0.967 & 0.993 \\ \hline
\textbf{DropV} & 0.508 & 0.572 & 0.941 & 0.955 & \textbf{0.775} & \textbf{0.861} & \textbf{0.901} \\ \hline
\textbf{DropF} & 0.544 & 0.585 & 0.940 & 0.960 & 0.908 & 0.955 & 0.984 \\ \hline
\textbf{DropFL} & 0.546 & 0.587 & 0.946 & 0.964 & 0.784 & 0.863 & \textbf{0.916} \\ \hline
\textbf{DropL} & 0.541 & 0.589 & 0.945 & 0.952 & 0.805 & 0.874 & 0.920 \\ \hline
\textbf{DropRN} & 0.537 & 0.586 & 0.935 & 0.933 & 0.907 & 0.958 & 0.990 \\ \hline
\textbf{DropRV} & 0.538 & 0.588 & 0.942 & 0.954 & 0.806 & 0.876  & 0.919 \\ \hline
\textbf{KeepNV} & 0.464 & 0.562 & 0.944 & 0.952 & 0.909 & 0.953 & 0.987 \\ \hline
\textbf{KeepN} & \textbf{0.443} & \textbf{0.559} & 0.963 & 0.972 & \textbf{0.746} & \textbf{0.854} & \textbf{0.893} \\ \hline
\textbf{KeepV} & \textbf{0.417} & \textbf{0.554} & 0.924 & \textbf{0.915} & 0.949 & 0.978 & 1.023 \\ \hline
\textbf{Shuffle} & 0.506 & \textbf{0.550} & \textbf{0.905} & \textbf{0.917} & 0.891 & 0.939 & 0.967      \\ \hline
\end{tabular}}
\caption{Comparison of robustness scores on probing tasks: perturbation types vs. probing tasks, averaged across layers. Lowest robustness values per column are highlighted in bold.}
\label{tab:robustness_perturbations_tasks}
\end{table}

\noindent\textbf{Overall insights from perturbation experiments.} Text perturbation analysis reveals that universal models such as InfoXLM, BLOOM, mGPT, XGLM and mT5 demonstrate higher resilience to perturbations compared to BERT-like models mBERT, IndicBERT, and MuRIL. Perhaps, the larger multilingual models are more robust as they rely less on language specific word order compared to Indic models~\citep{dufter2020identifying,liang2023xlm}.
Specifically, dropping both nouns and verbs proves to be particularly sensitive across all languages, impacting semantic and syntactic properties significantly. Perturbations, such as position alterations, also affect model performance, emphasizing the importance of considering linguistic nuances in robustness testing.

%\vspace{-0.25cm}
\subsection{Correlation Analysis of Probing with Downstream Tasks}
%\vspace{-0.2cm}

We conducted a correlation analysis between our IndicSentEval probing results and downstream task performance from the IndicGLUE benchmark. Specifically, we analyze how the performance of the IndicSentEval probing tasks correlates with the results of the IndicGLUE downstream tasks and other benchmarks. This includes models such as IndicBERT, MuRIL, mBERT, and XLM-R across syntactic and semantic tasks like NER, POS tagging, sentiment analysis, and natural language inference. The Table~\ref{tab:downstream_results} shows the aggregated scores across multiple Indic languages. We make the following observations from Table ~\ref{tab:downstream_results}: (i) Models like MuRIL and IndicBERT, which scored highest on syntactic and semantic probes, also performed best on corresponding tasks like POS tagging, NER (syntactic), and sentiment classification, NLI (semantic). (ii) Universal models like mBERT showed weaker probe performance and correspondingly lower scores on most IndicGLUE tasks, especially for morphologically rich languages. (iii) Tasks requiring deeper understanding (e.g. NLI) correlated more strongly with semantic probes, while syntactic probes aligned better with tagging tasks. This implies alignment between what probes measure and what different task categories demand. These findings suggest that the linguistic properties measured by probing are indeed predictive of real-world task performance, particularly in morphologically rich Indic languages.

\begin{table}[t]
\centering
\scriptsize
\begin{tabular}{|c|c|c|c|c|c|c|}
\hline
\multicolumn{1}{|c|}{\textbf{Model}} & 
\multicolumn{1}{c|}{\textbf{Syntactic}} & 
\multicolumn{1}{c|}{\textbf{Semantic}} & 
\multicolumn{1}{c|}{\textbf{POS}} & 
\multicolumn{1}{c|}{\textbf{NER}} & 
\multicolumn{1}{c|}{\textbf{Sentiment}} &
\multicolumn{1}{c|}{\textbf{NLI}} \\
\multicolumn{1}{|c|}{} & 
\multicolumn{1}{c|}{ \textbf{Probes}} & 
\multicolumn{1}{c|}{ \textbf{Probes}} & 
\multicolumn{1}{c|}{ \textbf{Tagging}} & 
\multicolumn{1}{c|}{} & 
\multicolumn{1}{c|}{\textbf{Analysis}} & 
\multicolumn{1}{c|}{} \\
\hline
IndicBERT & 0.82 & 0.80 & 0.88 & 0.84 & 0.81 & 0.79 \\
MuRIL     & 0.85 & 0.86 & 0.89 & 0.83 & 0.85 & 0.87 \\
mBERT     & 0.72 & 0.70 & 0.76 & 0.88 & 0.70 & 0.69 \\
XLM-R     & 0.80 & 0.78 & 0.85 & 0.85 & 0.83 & 0.82 \\
\hline
\end{tabular}
%\vspace{-0.2cm}
\caption{Performance comparison of multilingual models across various probing and downstream tasks.}
\label{tab:downstream_results}
\end{table}

%\vspace{-0.15cm}
\section{Discussion and Conclusion}
\label{sec:conclusions}
%\vspace{-0.15cm}
%\section{Conclusion}
% We investigated how well multilingual Transformer-based models comprehend linguistic structures in 6 Indic languages by evaluating 9 models using 8 probing tasks. To support this study, we contributed the \textsc{IndicSentEval} dataset. Our results show that Indic-specific models like MuRIL and IndicBERT are the most effective at capturing linguistic properties within Indic languages due to their targeted training. Universal models like mBERT, InfoXLM, BLOOM, mGPT and XGLM exhibit mixed results. Our perturbation analysis reveals that decoder-based models are the most robust, as also observed by \citet{neerudu2023robustness}. Verbs and the order of words are the most important signals that help the models encode linguistic structures. TreeDepth is the most sensitive to all mentioned text perturbations, while SubjNum and ObjNum are the most resilient.
We evaluated 9 multilingual Transformer-based models on 8 probing tasks to understand linguistic structures in 6 Indic languages, using our contributed \textsc{IndicSentEval} dataset. Indic-specific models like MuRIL and IndicBERT excel due to targeted training, while universal models like mBERT, InfoXLM, BLOOM, mGPT, and XGLM show mixed results. Perturbation analysis reveals decoder-based models are the most robust. 
%We demonstrate that not all multilingual models effectively capture the structural properties across the six languages. More crucially, 
Overall, our findings highlight the importance of language model interpretability. Language proficiency is seen in models with comprehensive training datasets. Encoding ability and perturbation impact vary across languages and models, underscoring the need for robust training strategies and tailored architectures to handle linguistic variations and perturbations effectively.

\section{Limitations}
The current work focused on only 6 Indic languages. It would be interesting to expand this to more Indic languages. 

We performed experiments with base versions of various models. It would be interesting to see if larger variants perform better at these tasks and if the trends differ.

% Also, experiments have been done on traditional pre-trained language models with a few hundred million parameters. In the future, we plan to perform such studies for models with billions of parameters like LLaMA and mGPT.

We experimented with a basic set of linguistic properties. It would be nice to explore more complex linguistic properties like morphological tagging, syntactic parsing, and semantic similarity.

\section{Ethics Statement}
All the models used in this work are publicly available on Hugging Face and free for research.  

We utilized publicly accessible resources in SSF format from  \url{https://ltrc.iiit.ac.in/showfile.php?filename=downloads/kolhi/} and  \url{https://ltrc.iiit.ac.in/showfile.php?filename=downloads/lingResources/newreleases.html}. The datasets are licensed under Creative Commons by Non-Commercial 4.0 (CC BY-NC 4.0). No anticipated risks are associated with using the data from these provided links. We adapted the accessible resources to generate diverse probing and perturbation datasets as required.

\bibliography{references}
\bibliographystyle{acl_natbib}
\newpage
\appendix

{\Large \textbf{Overview of Appendix Sections}}

\begin{itemize}
    \item Appendix~\ref{app:relatedwork}: Related Work
    \item Appendix~\ref{app:ssfFormat}: SSF Format.
    \item Appendix~\ref{app:dataStats}: \textsc{IndicSentEval} Dataset Statistics.
    \item Appendix~\ref{probing_tasks}: Probing Tasks.
    \item Appendix~\ref{app:perturbationExamples}: Text Perturbation Examples.
    \item Appendix~\ref{app:modelDetails}: Details of Multilingual Models.
    \item Appendix~\ref{probing_results}: Detailed Probing Results.
    \item Appendix~\ref{app:discussion}: Additional Discussion
\end{itemize}

\section{Related Work}
\label{app:relatedwork}

Our work is most closely related to that of~\citet{jawahar2019does} and~\citet{mohebbi2021exploring}, who focus on understanding the interpretability of language models via probing and observing linguistic structures in English. Our study on multilingual language models for six Indic languages complements their English-focused studies. 

Our work also relates to the growing literature that creates resources for low-resource Indic languages. Several recent studies have developed resources including, Indic language models~\citep{khan2024indicllmsuite}, Indic NLP suite~\citep{kakwani2020indicnlpsuite}, Indic BART~\citep{dabre2021indicbart}, Indic NLG~\citep{kumar2022indicnlg}, IndicMT Eval~\citep{dixit2023indicmt}, and Naamapadam~\citep{mhaske2022naamapadam}.
Our research is the first to interpret both universal and Indic-specific models for Indic languages through probing representations and assessment of robustness, while also providing implications for future research directions. Overall, we complement these works by studying the linguistic structure of a wide range of low-resource Indic languages on multilingual models.

\noindent\textbf{Non-English and Multilingual Probing Studies:}
Research on probing language models has increasingly moved beyond English to multilingual and non-English contexts, particularly focusing on morphologically rich and low-resource languages~\citep{pires2019multilingual,edmiston2020systematic,abdelali2022post,acs2024morphosyntactic}. For instance,~\citet{pires2019multilingual} is one of the earliest probing studies on mBERT, showing that mBERT enables zero-shot cross-lingual transfer using shared representations, but performs best on typologically similar languages, struggling with distant or low-resource ones. Extending from previous work,~\citet{edmiston2020systematic} probe BERT’s hidden layers for discrete morphological features in five languages (e.g. gender, tense, case) and observe that BERT encodes discrete morphological features across multiple languages, but lacked explicit analysis of model robustness across language families. More recently,~\citet{abdelali2022post} probed several Arabic BERT-based models (and mBERT) and observed that Arabic BERT models capture morphology in lower layers and syntax in higher layers, but had limited focus on dialectal variation and cross-model comparison. Very recently,~\citet{acs2024morphosyntactic} introduced a large multilingual probing dataset across 42 languages and found strong morphological encoding in mBERT/XLM-R, but did not analyze model robustness or compare to language-specific models. While these earlier studies primarily focused on morphological encoding in universal multilingual models (like mBERT and XLM-R), IndicSentEval uniquely evaluates both universal multilingual models and Indic-specific models, explicitly comparing their robustness and representational capabilities across six morphologically rich and script-diverse Indic languages.

\noindent\textbf{Critiques of Probing Methodology and Recent Advances}
Probing methods have faced significant criticism in recent years, primarily for overinterpreting what models “know” from linear classifier performance~\cite{hewitt2019designing,voita2020information,ravichander2021probing}. For instance,~\citet{hewitt2019designing} showed that complex probes may "memorize" tasks, not reveal what's encoded in the representations, ensuring probes extract meaningful, not spurious, information. Extended to previous work,~\citet{voita2020information} proposed using Minimum Description Length (MDL) to evaluate how efficiently linguistic features can be extracted. They show that MDL probes reward simpler models and fewer training samples, improving interpretability and robustness. Further,~\citet{ravichander2021probing} demonstrate that models can encode linguistic properties even if these properties are not necessary for the task the model was trained on. Specifically, they highlight the importance of careful controls when designing probing experiments, as high probing accuracy does not necessarily indicate that the probed information is utilized by the model during its primary task. In our IndicSentEval work, we address these critiques by: (i) using lightweight probes with standardized architecture across all models and languages; (ii) focusing on relative comparisons (across layers, languages, and perturbations) rather than absolute performance; (iii) including input-level perturbations to test mechanistic interpretability and model robustness; and (iv) complementing standard probing with morpho-semantic tasks tailored to Indic grammar, allowing deeper insight into model behavior for typologically diverse languages.

\noindent\textbf{Differences from English-Centric Findings:}
Probing studies in non-English and multilingual contexts have revealed notable differences from English-centric findings, due to linguistic diversity and the multilingual training regime~\cite{zheng2022probing,tikhonova2023ad,godunova2024probing,dang2024morphology}. For instance,~\citet{zheng2022probing} showed that multilingual models encode universal syntactic features well but struggle with fine-grained morphology (e.g., verb agreement), unlike their performance on English tasks. Similarly,~\citet{tikhonova2023ad} reported that mBERT lacks the clear layer-wise linguistic hierarchy found in English BERT, suggesting architectural behavior does not generalize across languages. More recently, studies including~\citet{godunova2024probing} found that discourse-level understanding remains uniformly weak across high- and low-resource languages, contrasting with better performance on lower-level tasks in English. Also,~\citet{dang2024morphology} found that GPT models generalize well to inflect unseen words in morphologically simple languages, but performance declines sharply as morphological complexity increases. In contrast to prior studies that either focused primarily on English or reported high-level trends across many languages, IndicSentEval reveals that while multilingual models exhibit consistent representational hierarchies for English, they show mixed and often unstable behavior across Indic languages, especially in encoding syntactic and morphological features-highlighting language-specific gaps that broader multilingual benchmarks often obscure.

\section{SSF format}
\label{app:ssfFormat}

\begin{figure*}[!ht]
    \centering
    \includegraphics[width=\linewidth]{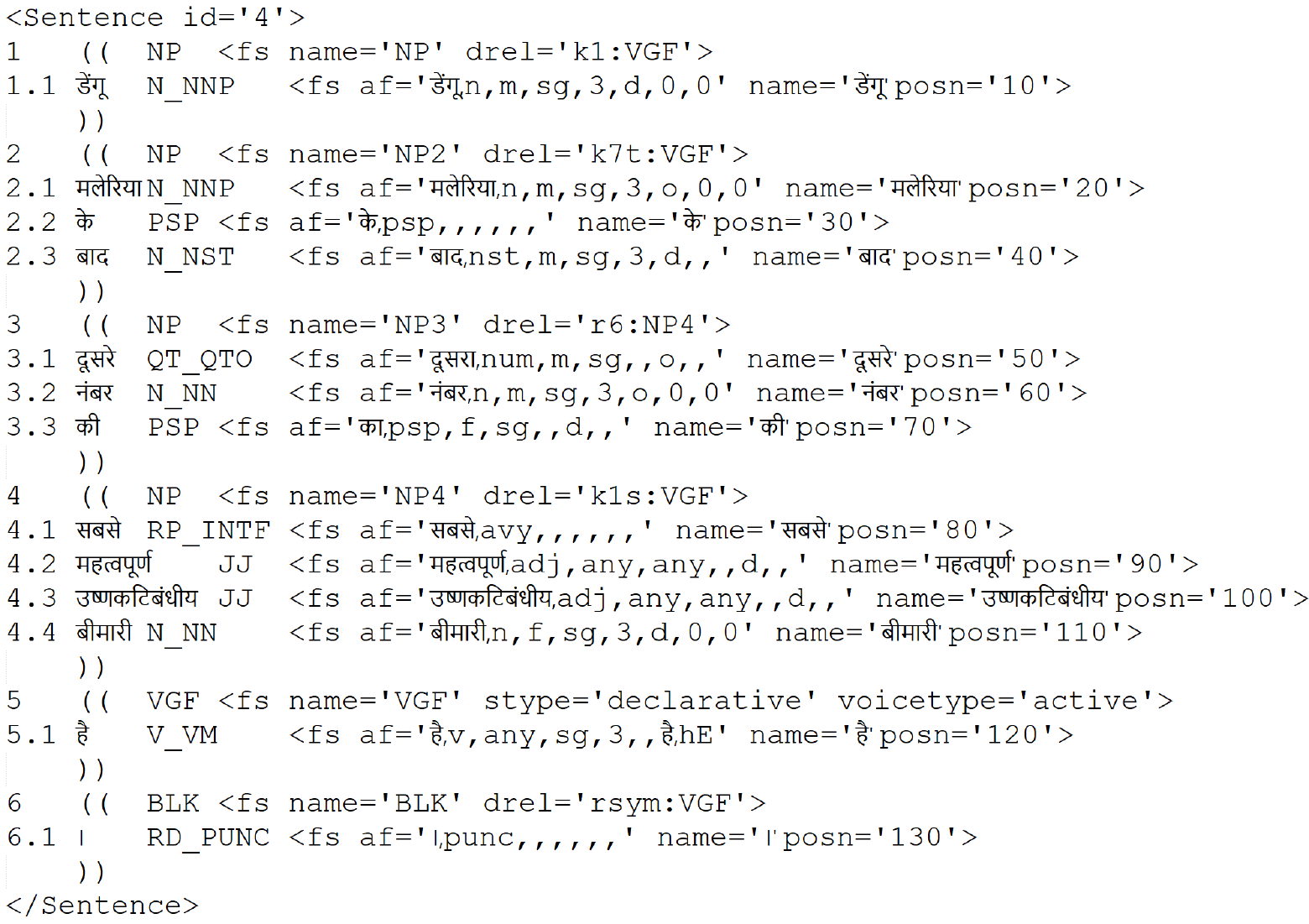}
    \captionof{figure}{A sample of an SSF formatted sentence in Hindi language.}
    \label{fig:SSFExample2}
\end{figure*}

We curate the \textsc{IndicSentEval} dataset from resources generated by the ILMT initiative, which serves as an Indic language counterpart to the SentEval dataset and offers labeled data for the eight probing tasks. We utilize the morph and chunk level Indic languages data~\cite{tandon2017unity,bhatt2009multi,xia2008towards}, which is available in Shakti Standard Format (SSF)~\cite{bharati2007ssf,bharati1995natural}. SSF is a highly readable representation for storing Indic language data with linguistic annotations. We refer the reader to read~\cite{bharati2007ssf} for more details about the SSF format. 

An example of a Hindi sentence in the SSF format is illustrated in
Fig.~\ref{fig:SSFExample2}. Each line in Fig.~\ref{fig:SSFExample2} delineates a word within a sentence. Every line of a sentence representation in SSF comprises four components: Address, token, Category, and Attribute-value pairs. The Address encompasses two numbers, denoting the chunk number the word resides in and its relative position within the chunk. Token signifies the POS tag corresponding to the word. The feature list, articulated as "<fs af = root, category, gender, number, person, case, tense, aspect>", encapsulates linguistic feature information of a word. The attribute fields are fixedly positioned and separated by commas. Attributes remain blank if a property is inapplicable to the word. Properties like root, category, gender, number, person, and case are delineated within the feature list.

\section{INDICSENTEVAL dataset statistics}
\label{app:dataStats}

\begin{table}[!ht]
\centering
\scriptsize
\resizebox{\columnwidth}{!}{\begin{tabular}{|l||c|c|c|c|c|c|c|c|}
% \begin{tabular}{|l|c|c|c|c|c|c|c|c|c|c|c|c|c|c|c|c|c|c|}
\hline
% Lang & \multicolumn{2}{c|}{SentLen} & \multicolumn{2}{c|}{WC} & \multicolumn{2}{c|}{BShift} & \multicolumn{2}{c|}{TreeDepth} & \multicolumn{2}{c|}{VerbGen} & \multicolumn{2}{c|}{VerbNum} & \multicolumn{2}{c|}{VerbPer} & \multicolumn{2}{c|}{SubjNum} & \multicolumn{2}{c|}{ObjNum}\\
\multirow{2}{*}{\textbf{Lang}} & \textbf{Sent} & \multirow{2}{*}{\textbf{BShift}}  & \textbf{Tree} & \textbf{Subj} & \textbf{Obj}& \textbf{Verb} & \textbf{Verb} & \textbf{Verb} \\
& \textbf{Len} &  & \textbf{Depth} & \textbf{Num} & \textbf{Num}& \textbf{Gen} & \textbf{Num} & \textbf{Per} \\
\hline
\texttt{hi}&12202&12202&8911&398&812&8911&8897&8897\\
\texttt{te}&3192&3192&3192&763&578&2243&2635&2501\\
\texttt{ur}&2363&2363&2363&374&356&1463&1465&1446\\
\texttt{ml}&7667&7667&7667&1556&642& -& -& -\\
\texttt{kn}&9806&9806&9806&4790&3690&1329&5371&5448\\
\texttt{mr}&12029&12029&12029&2488&3815&5934&7637&6791\\
\hline
\end{tabular}}
% %\vspace{-0.1cm}
\caption{Number of samples for different probing tasks per language in \textsc{IndicSentEval}. The symbol `-' signifies the absence of SSF data for the Malayalam language for the VerbGen, VerbPer, and VerbNum tasks.}
\label{tab:ProbingTrainTestSplit}
\end{table}

\begin{table}[!ht]
\centering
\scriptsize
\begin{tabular}{|l|l|}
\hline
\textbf{Language} & \textbf{Vocabulary Size}  \\ \hline
\textbf{Hindi} & 19589 \\
\hline
\textbf{Kannada} & 25310 \\
\hline
\textbf{Malayalam} & 28723 \\
\hline
\textbf{Marathi} & 35607 \\
\hline
\textbf{Telugu} & 5834 \\
\hline
\textbf{Urdu} & 5593 \\
\hline
\end{tabular}
\caption{Vocabulary size of each language from \textsc{IndicSentEval}.}
\label{tab:vocab}
\end{table}

\begin{figure*}[!ht]
    \centering
    \includegraphics[width=\linewidth]{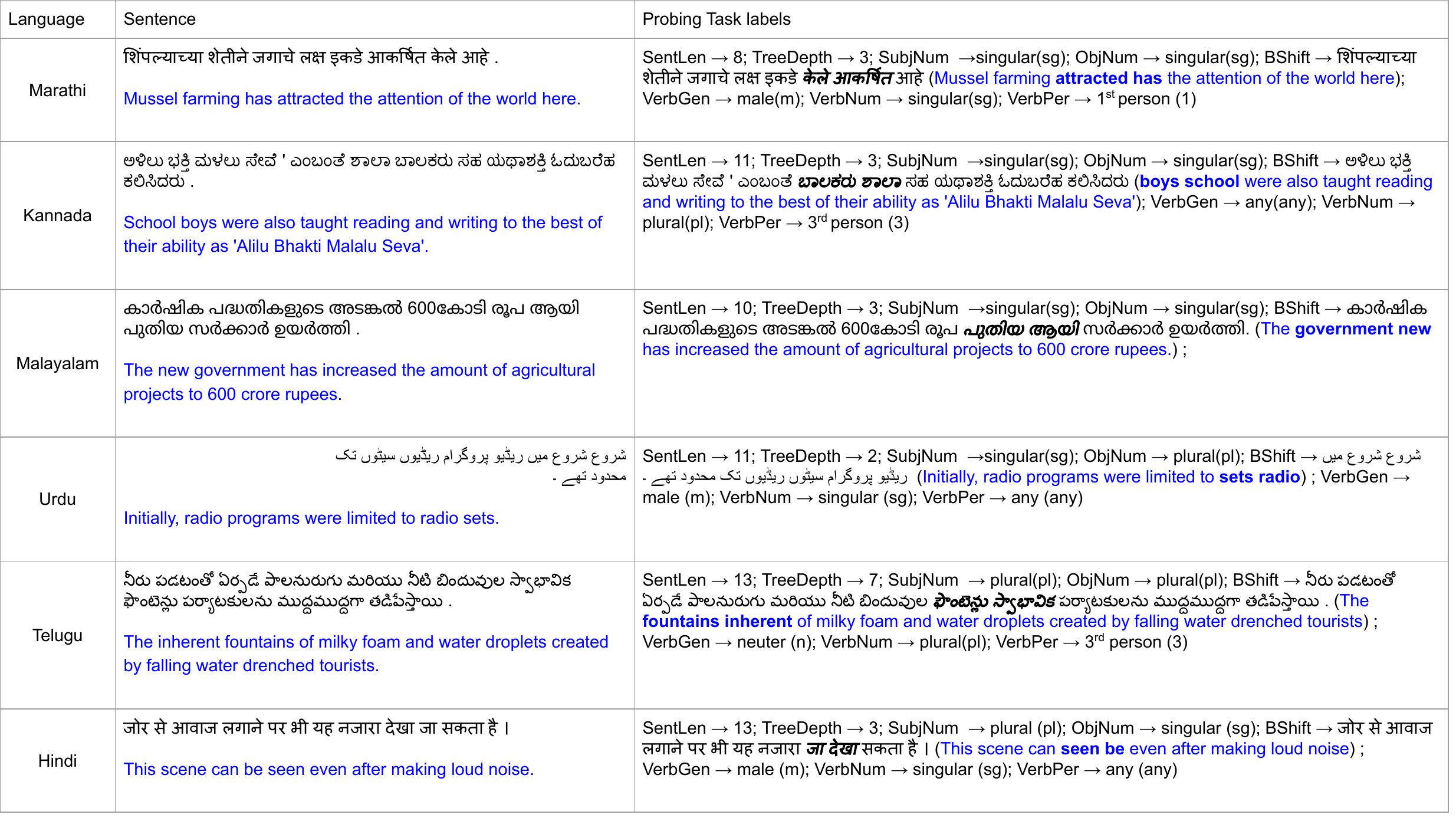}
    \captionof{table}{Examples for probing tasks w.r.t each language.}
    \label{tab:ProbingtaskExamples}
\end{figure*}

% Link for Examples of Probing Tasks: https://docs.google.com/drawings/d/1ntn8r-8_m45vEhTjbZjT4pdY9g3x0ryWNDjUhG-inbE/edit?usp=sharing

\section{Probing Tasks}
\label{probing_tasks}

\noindent\textbf{Surface level tasks}
\noindent(1) \textit{Sentence Length (SentLen):} Here, the objective is to predict the number of words in sentences, which has been grouped into 8 categories as shown in Table~\ref{tab:taskDetails} (see main paper).
%with lengths in these  intervals: 0: (0-5), 1: (6-8), 2: (9-12), 3: (13-16), 4: (17-20), 5: (21-25), 6: (26-28), 7: (29-32).

% \noindent(2) \textit{Word Content Analysis (WC):} The task is about predicting which of the target words appear on the given sentence. Each sentence contains a single target word, and the word occurs exactly once in the sentence. This is a classification task with $L_i$ words as targets for language $i$. This task allows us to analyze the content of sentences based on the occurrence of selected vocabulary words.

\noindent\textbf{Syntactic tasks}
\noindent(2) \textit{Tree Depth (TreeDepth):} The goal of this task is to predict the maximum depth of the sentence's syntactic tree, which is categorized into five options based on depth intervals as shown in Table~\ref{tab:taskDetails}. As constituency data in Indic languages is unavailable, we utilize dependency tree data to determine the tree depth. This task provides valuable insights into the structural complexity and organization of sentences.
%: 0: (0-2), 1: (3-5), 2: (6-8), 3: (9-11), 4: (12-20). 
\noindent(3) \textit{Bigram Shift (BShift):} This task involves binary classification aimed at predicting whether two consecutive tokens within a sentence are inverted or not.

%This task helps us analyze the model's understanding of the high-level semantics of the sentence. 

\noindent\textbf{Semantic tasks}
\noindent(4) \textit{Subject Number (SubjNum):} This task evaluates sentences to determine the number of the subject in the main clause. It categorizes the subjects as NN (singular) or NNS (plural or mass, such as ``colors,'' ``waves,'' etc.).
\noindent(5) \textit{Object Number (ObjNum):} This task involves identifying whether the object of the main clause in a sentence is singular or plural/mass. It uses the label NN for singular objects and NNS for plural or mass objects.
\noindent(6) \textit{Verb Gender (VerbGen):} This task involves categorizing the grammatical gender of the main verb in a sentence as masculine, feminine, neutral, or any.
\noindent(7) \textit{Verb Number (VerbNum):} This task assesses the number of the main verb in a sentence, determining whether it is singular, plural, or any.
\noindent(8) \textit{Verb Person (VerbPer):} This task involves categorizing the grammatical person of the sentence's main verb into one of 7 classes as shown in Table~\ref{tab:taskDetails}. Honorifics, especially common in Indic languages, are forms of address that convey respect and politeness based on social status, age, or relationship. A typical example from Hindi is the use of ``ji'' to respectfully address elders, as seen in ``dada-ji''. 
%: first, second, and third person, along with first, second, and third person honorific forms, or any. 
%This task classifies the grammatical person of the main verb of the sentence into one of the 7 classes: 1st person, 2nd person, 3rd person, 1st person honorific (respect), 2nd person honorific, 3rd person honorific, and any. Honorifics, prevalent in languages of Indic origin, are linguistic expressions used to show respect and politeness towards others, factoring in their social status, age, or relationship, with ``ji'' in Hindi serving as a common example to address elders respectfully, such as in ``dada-ji''.

\begin{figure*}[!ht]
    \centering
      \includegraphics[width=0.4\linewidth]{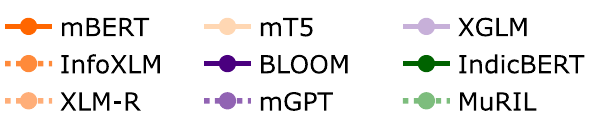} \\
      \includegraphics[width=0.325\linewidth]{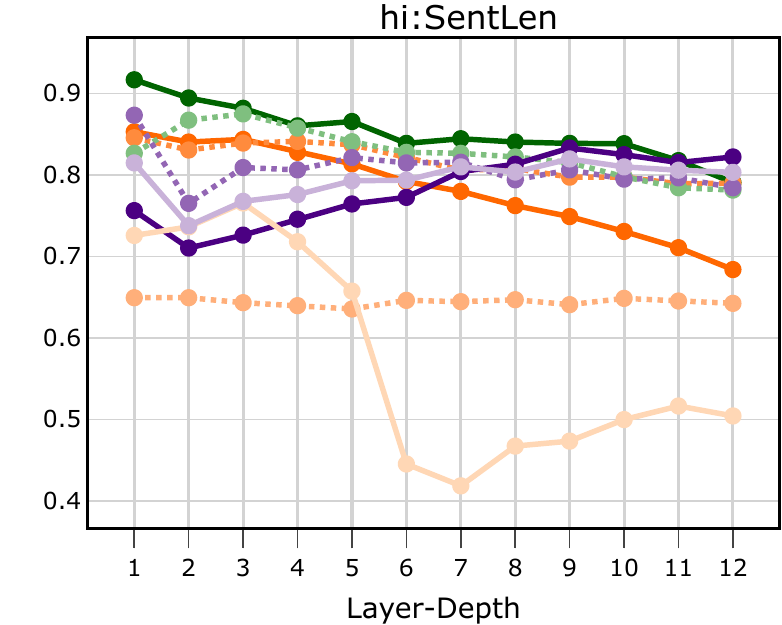}
      %\vspace{2pt}
      \includegraphics[width=0.325\linewidth]{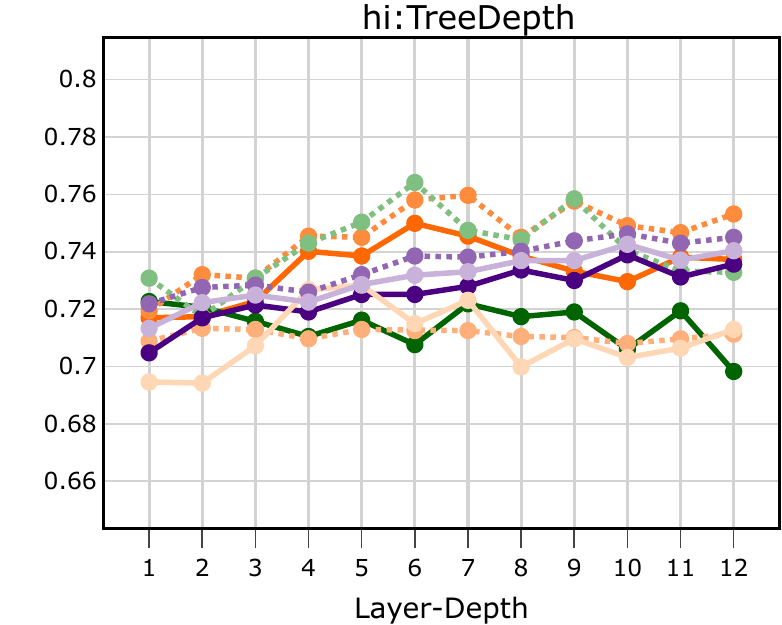}
      %\vspace{2pt}
      \includegraphics[width=0.325\linewidth]{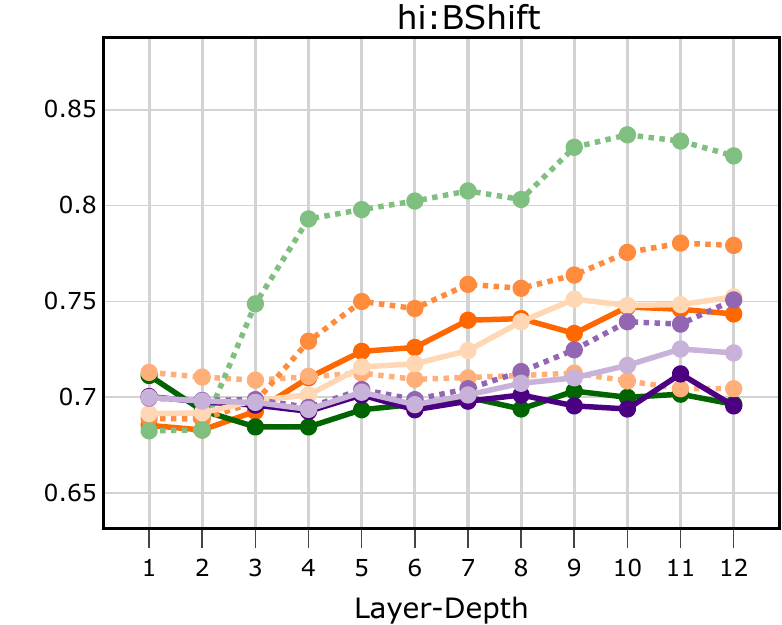}
      \includegraphics[width=0.325\linewidth]{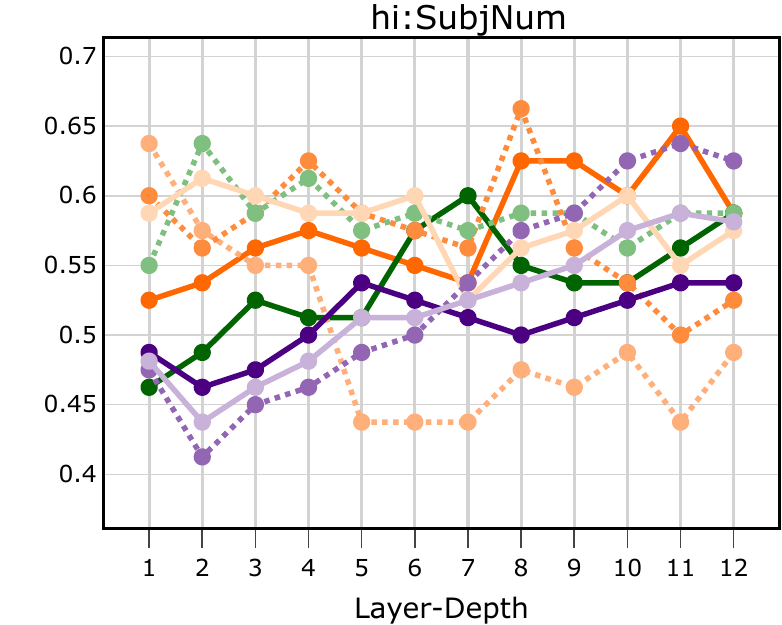}
      \includegraphics[width=0.325\linewidth]{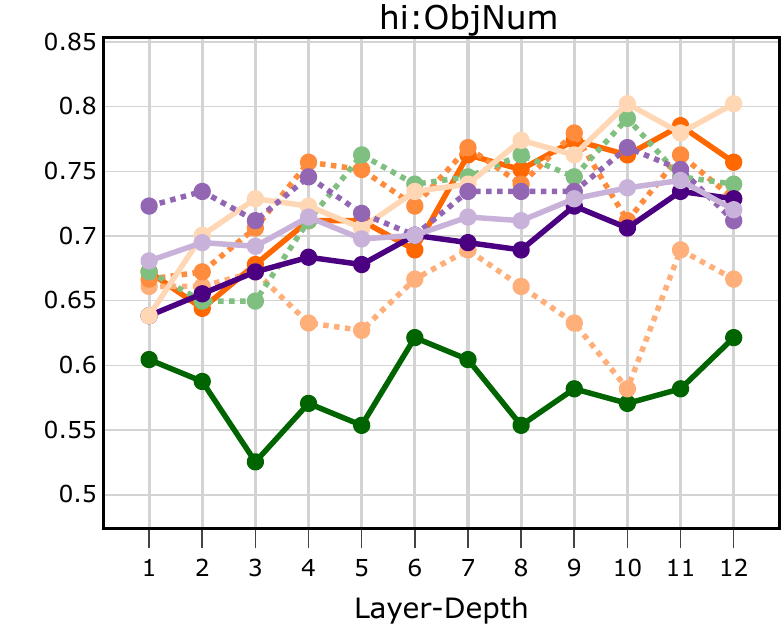}
      \includegraphics[width=0.325\linewidth]{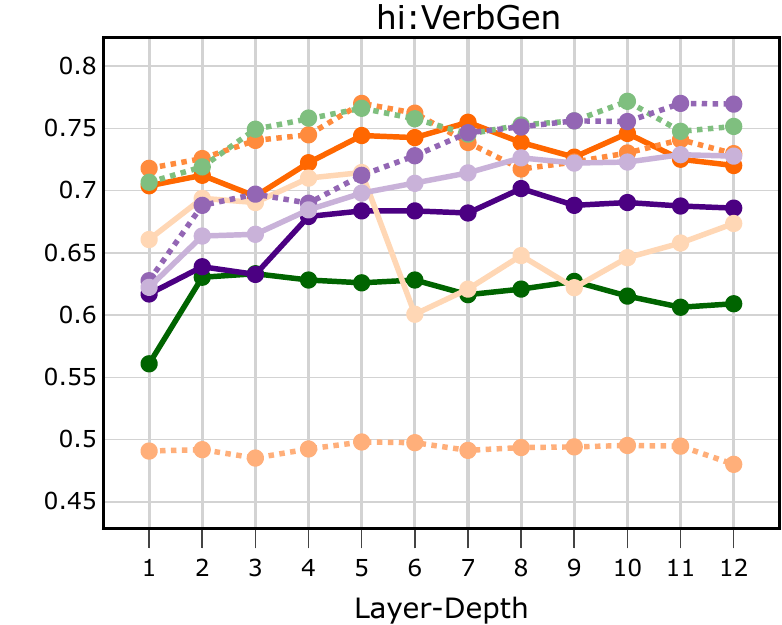}
     \includegraphics[width=0.325\linewidth]{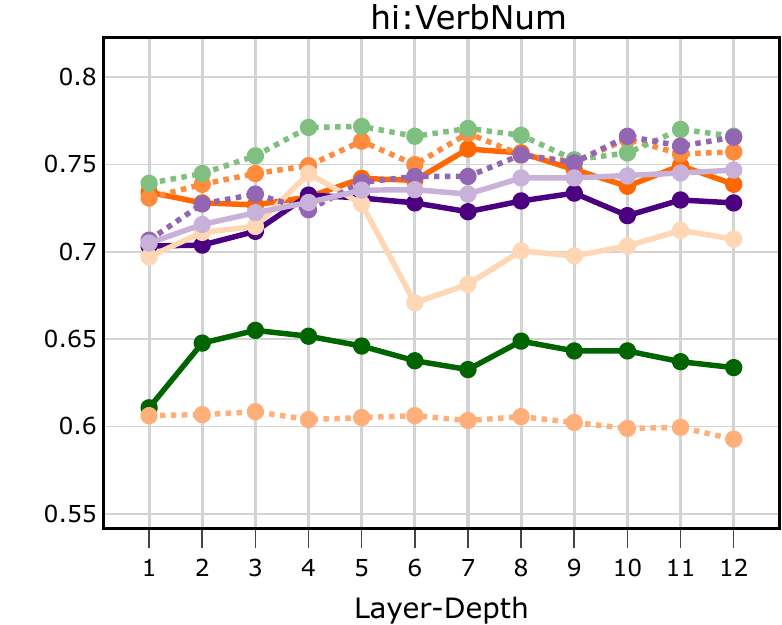}
     \includegraphics[width=0.325\linewidth]{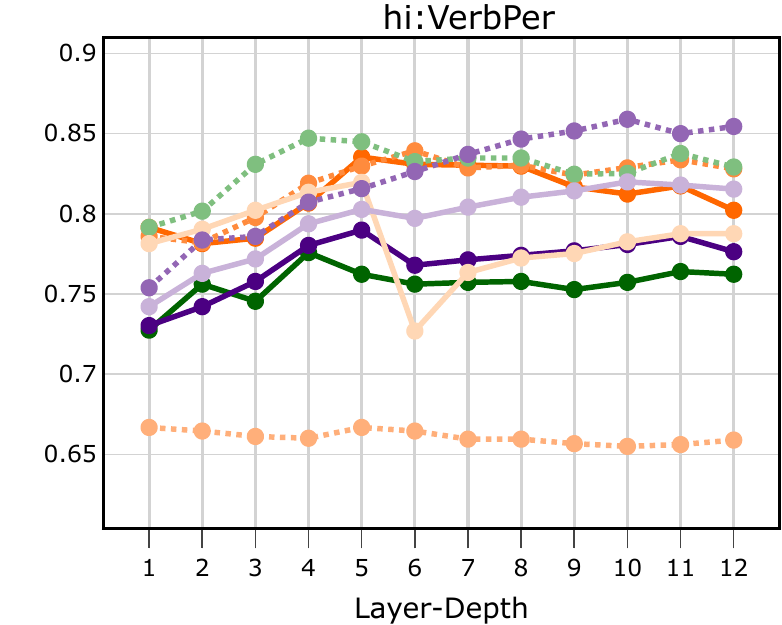}
\caption{Hindi language probing task results: Layerwise accuracy comparisons between various multilingual representations on 8 probing tasks.}
\label{fig:hi_probing_tasks}
\end{figure*}

\begin{figure*}[!ht]
    \centering
     \includegraphics[width=0.4\linewidth]{images/colorbar_final.pdf} \\
     \includegraphics[width=0.325\linewidth]{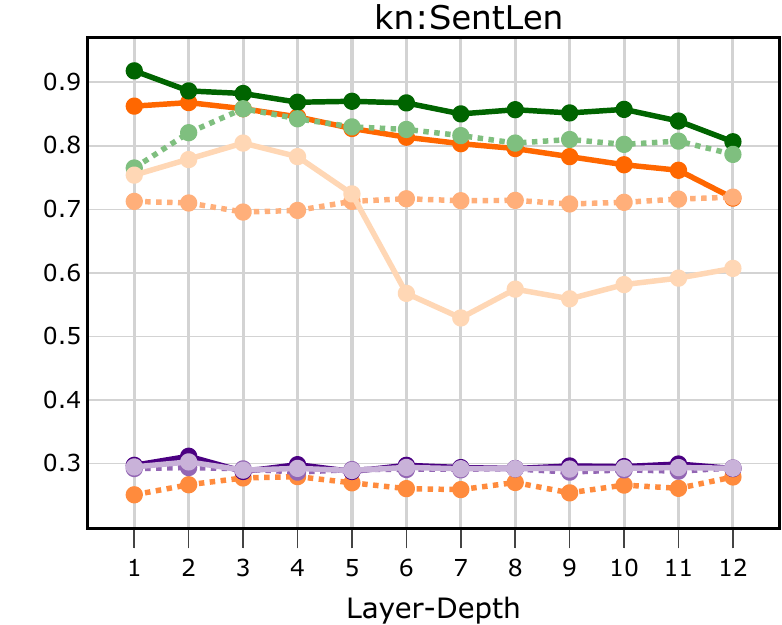}
    \includegraphics[width=0.325\linewidth]{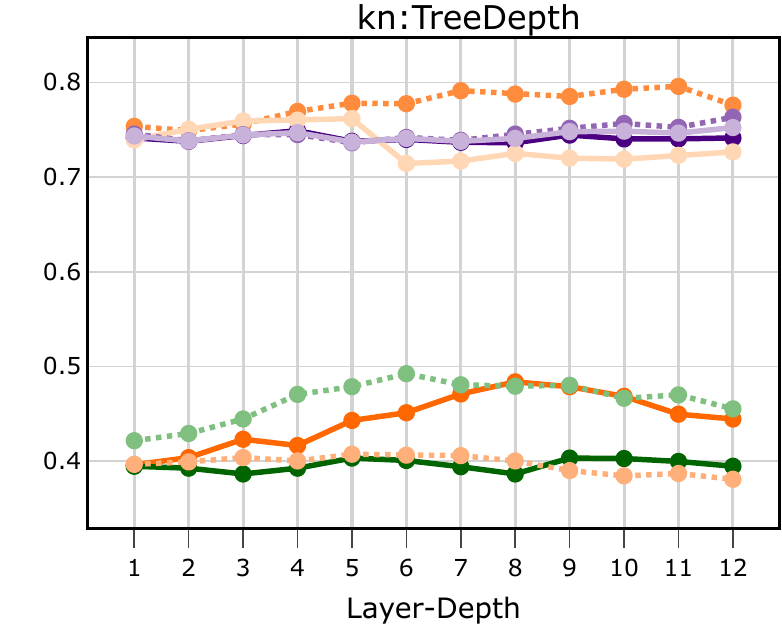}
    \includegraphics[width=0.325\linewidth]{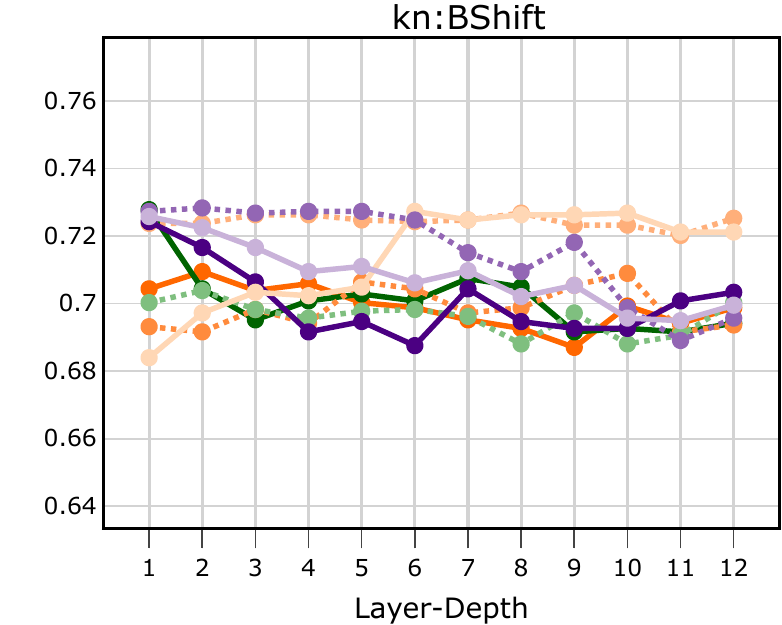}
    \includegraphics[width=0.325\linewidth]{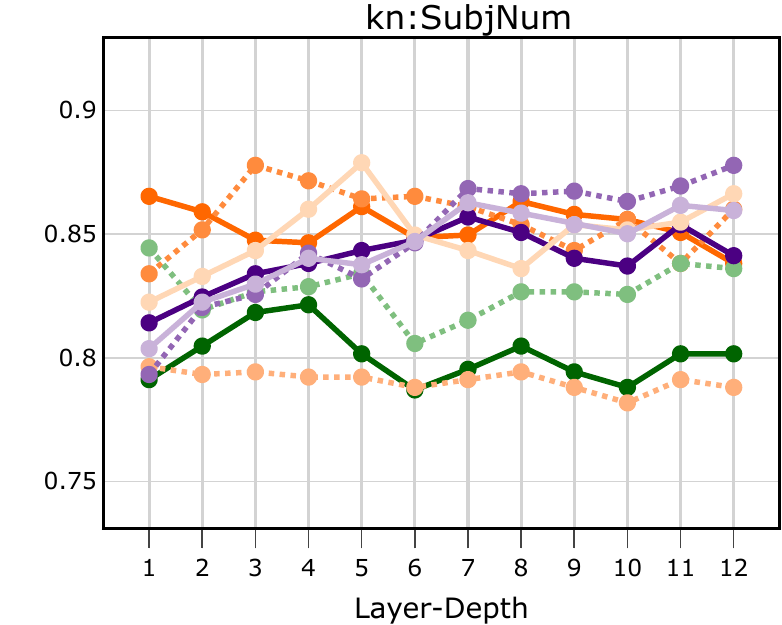}
    \includegraphics[width=0.325\linewidth]{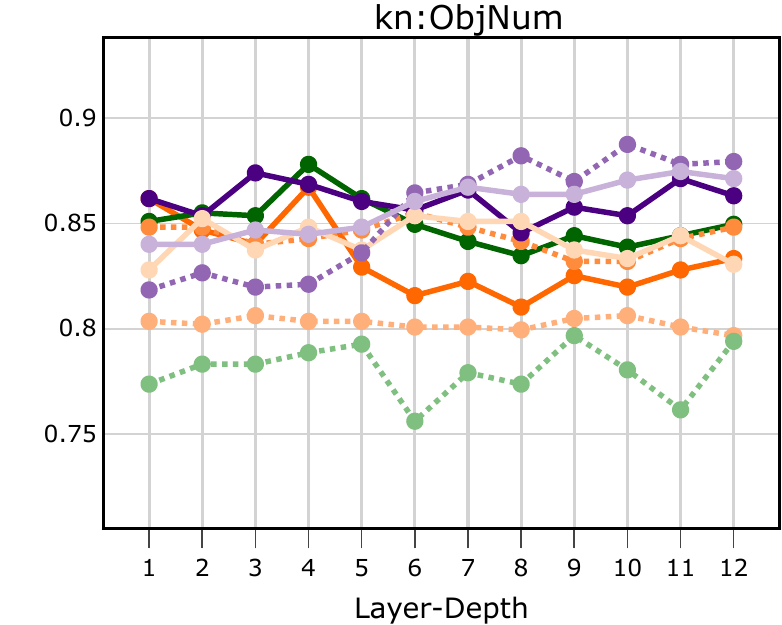}
    \includegraphics[width=0.325\linewidth]{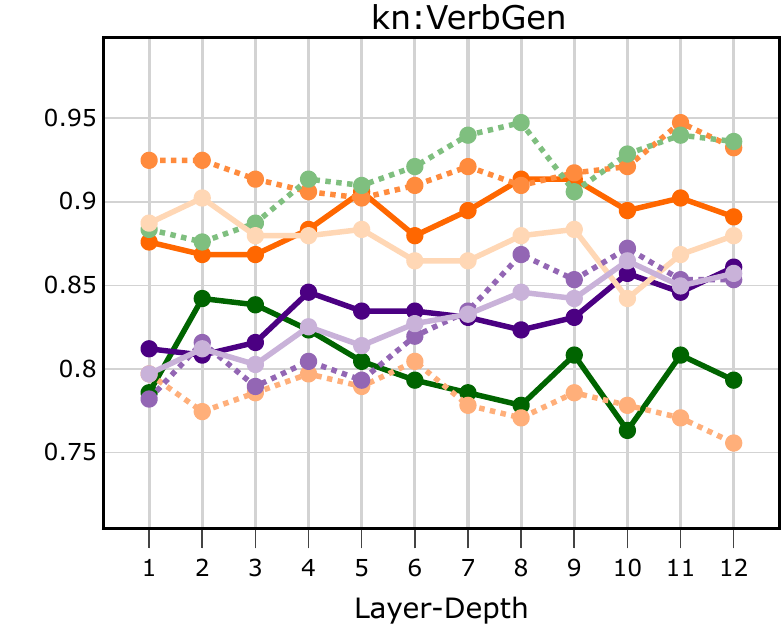}
    \includegraphics[width=0.325\linewidth]{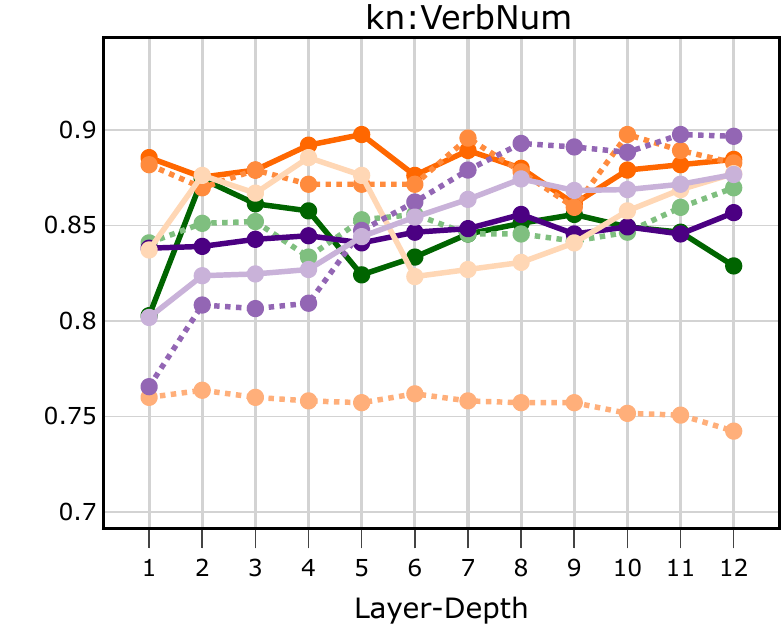}
    \includegraphics[width=0.325\linewidth]{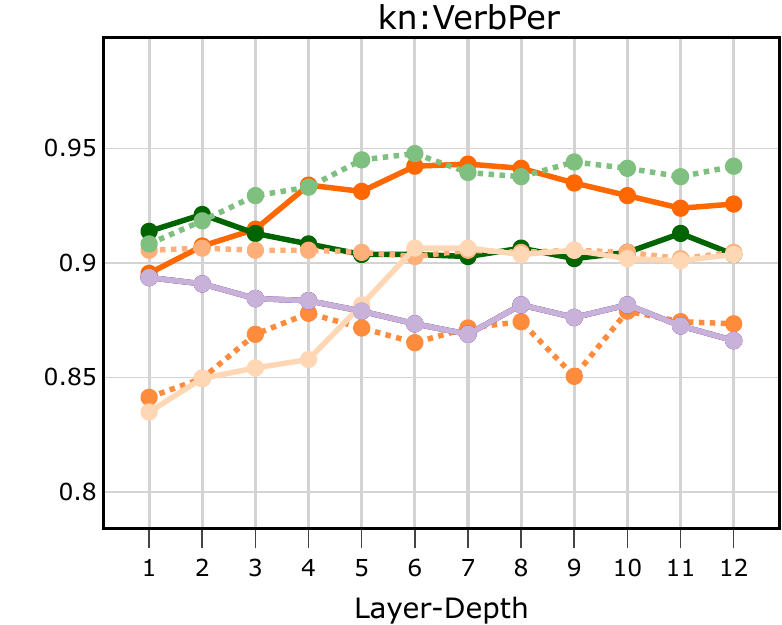}   
%\vspace{-0.2cm}
\caption{Kannada language probing task results: Layerwise accuracy comparisons between various multilingual representations on 8 probing tasks.}
\label{fig:kn_probing_tasks}
\end{figure*}

\begin{figure*}[!ht]
    \centering
     \includegraphics[width=0.325\linewidth]{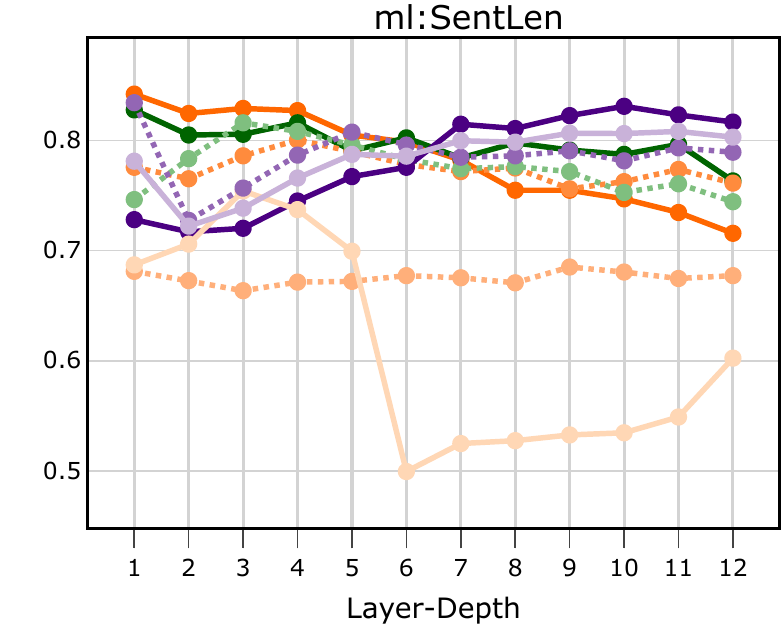}
    \includegraphics[width=0.325\linewidth]{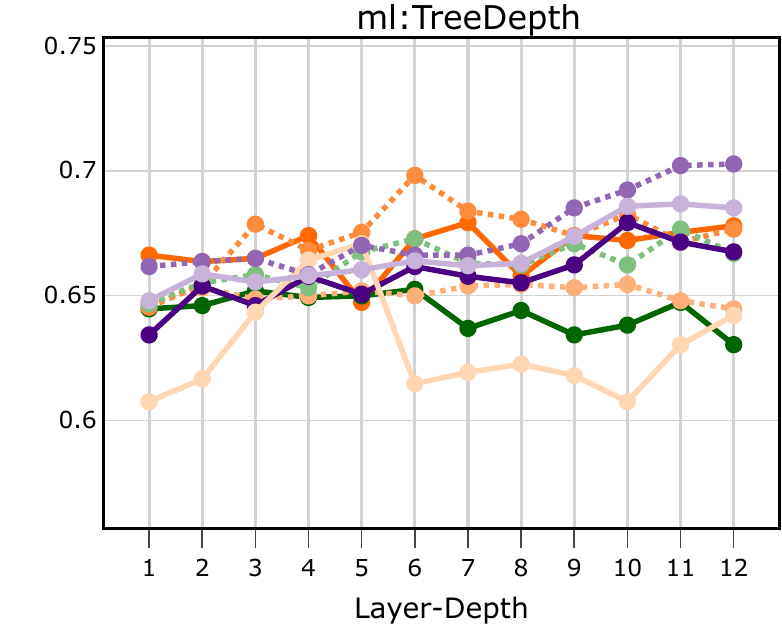}
    \includegraphics[width=0.325\linewidth]{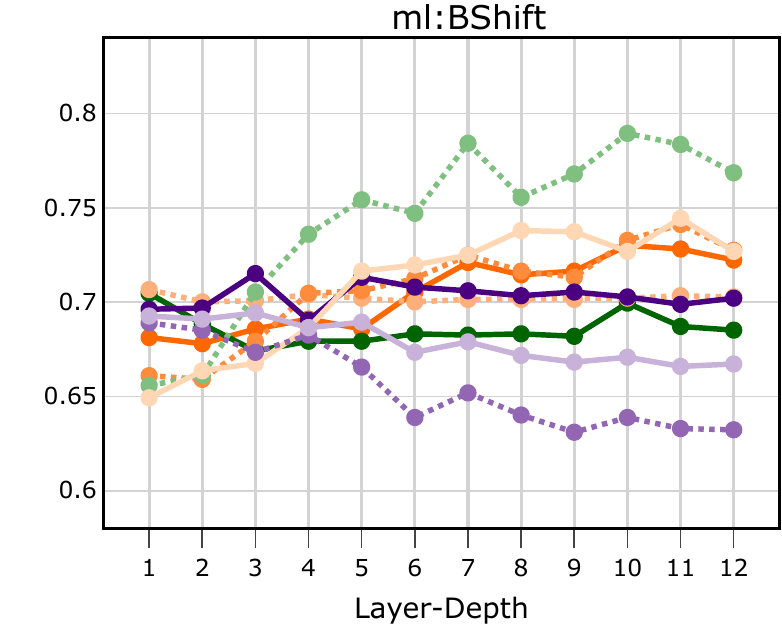}
    \includegraphics[width=0.325\linewidth]{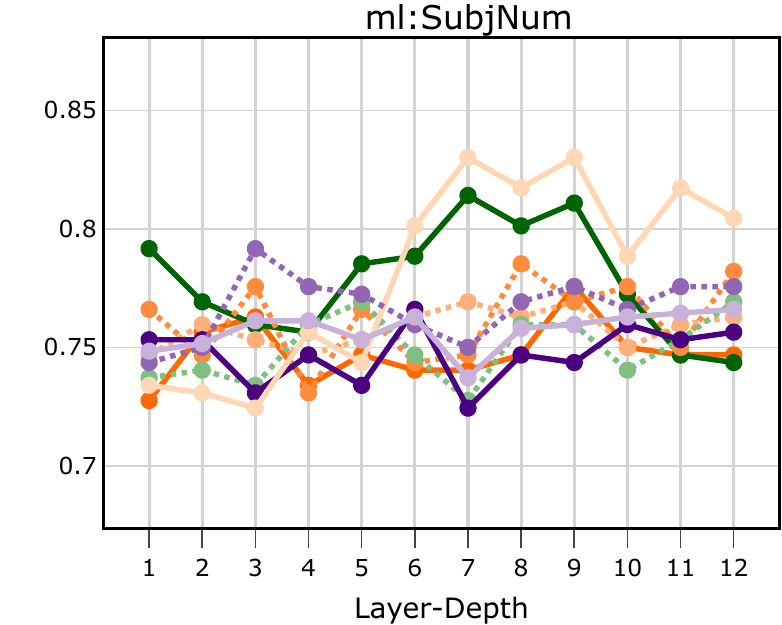}
    \includegraphics[width=0.325\linewidth]{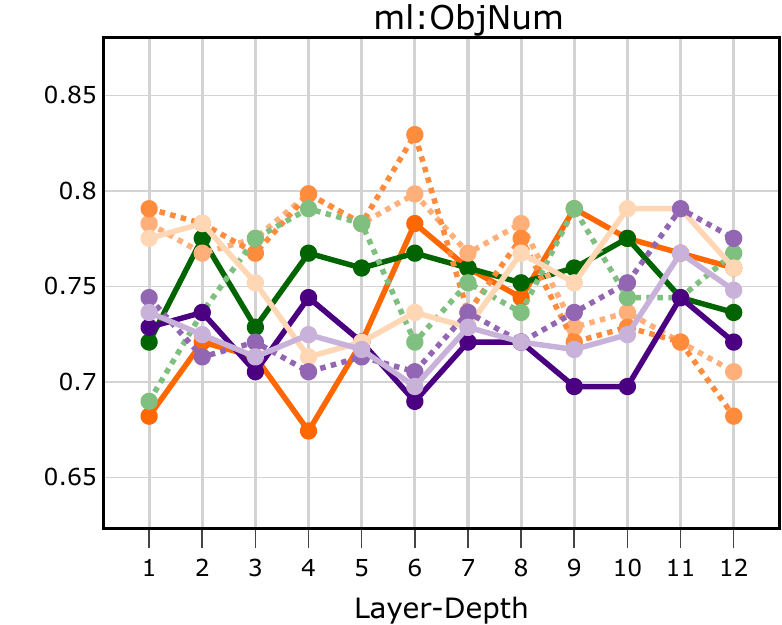} 
%\vspace{-0.2cm}
\caption{Malayalam language probing task results: Layerwise accuracy comparisons between various multilingual representations on 5 probing tasks.}
\label{fig:ml_probing_tasks}
\end{figure*}

\section{Text Perturbation Examples}
\label{app:perturbationExamples}

\begin{figure*}[!ht]
    \centering
    \includegraphics[width=0.7\linewidth]{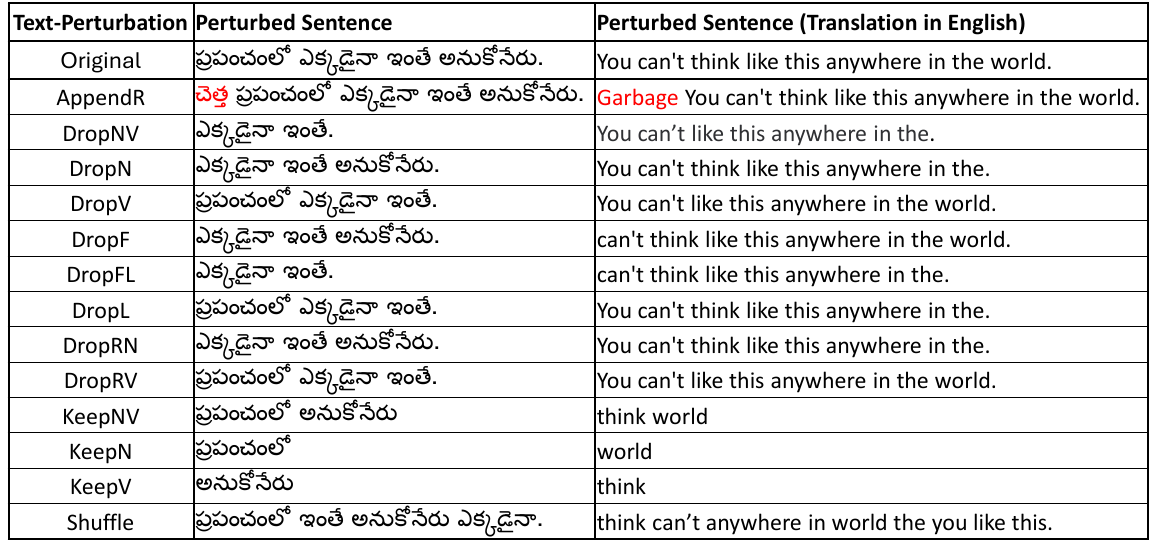}
    \captionof{table}{Text perturbation examples for Telugu language.}
    \label{fig:Sample of Telugu Perturbations}
\end{figure*}

\begin{figure*}[!ht]
    \centering
    \includegraphics[width=1\linewidth]{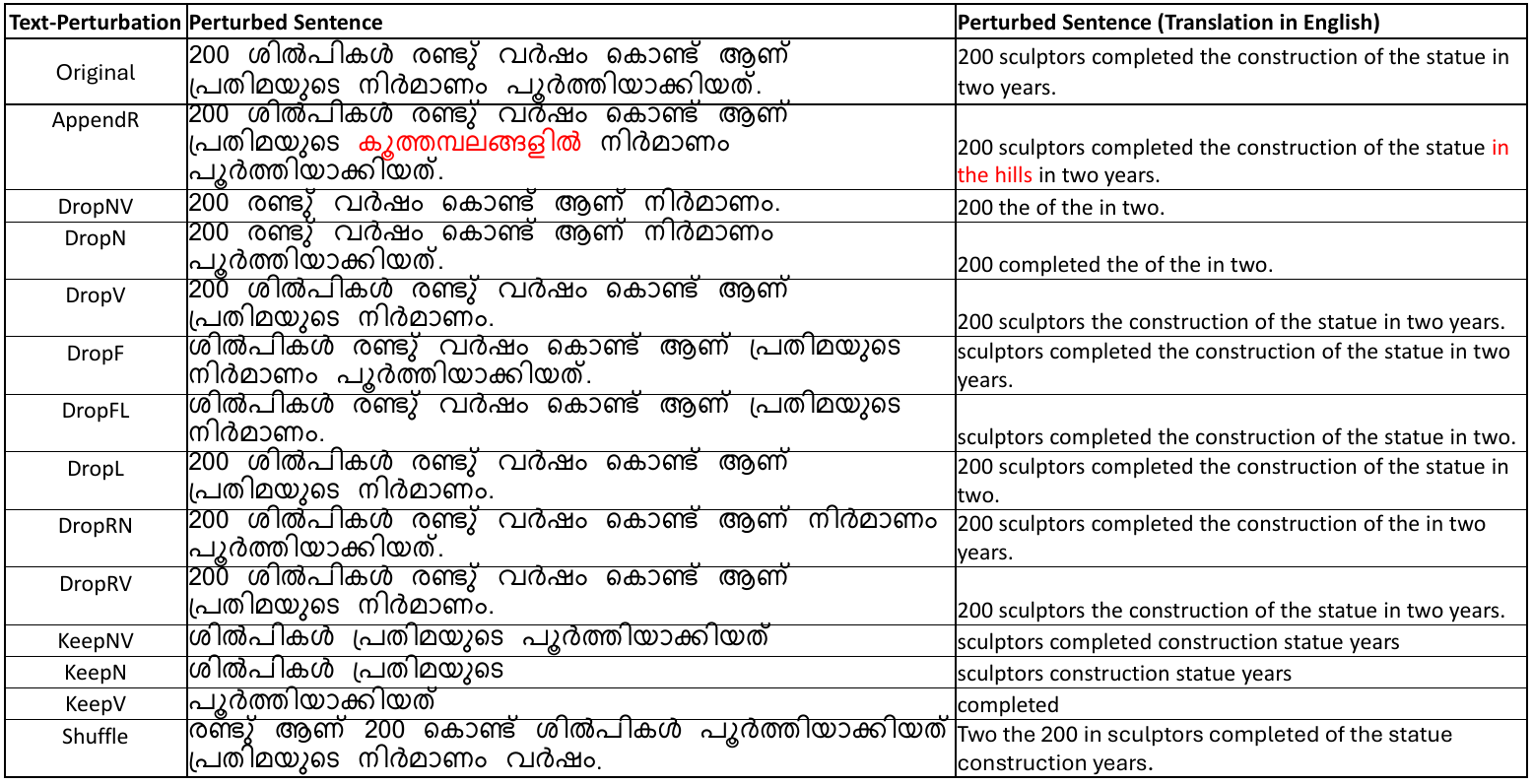}
    \captionof{table}{Text perturbation examples for Malayalam language.}
    \label{fig:Sample of Malayalam Perturbations}
\end{figure*}

\begin{figure*}[!ht]
    \centering
    \includegraphics[width=\linewidth]{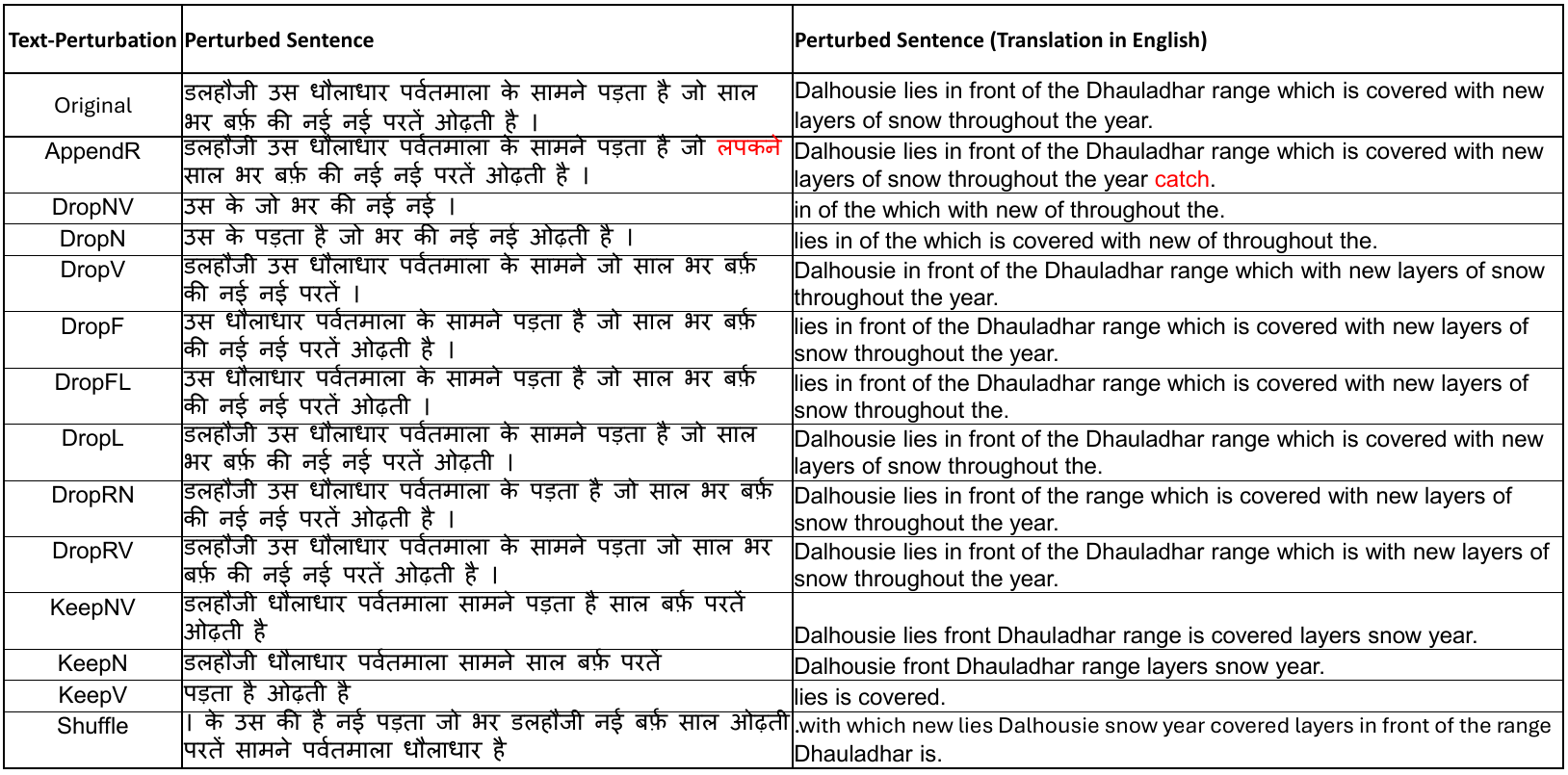}
    \captionof{table}{Text perturbation examples for Hindi language.}
    \label{fig:Sample of Hindi Perturbations}
\end{figure*}

\begin{figure*}[!ht]
    \centering
    \includegraphics[width=\linewidth]{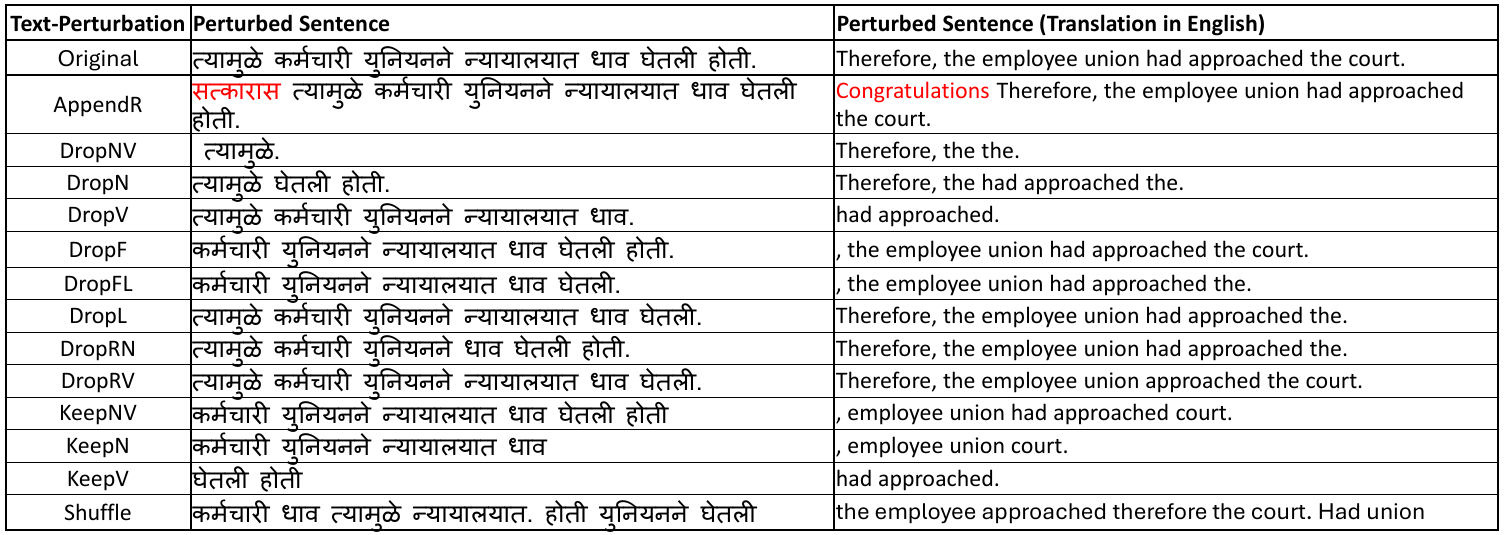}
    \captionof{table}{Text perturbation examples for Marathi language.}
    \label{fig:Sample of Marathi Perturbations}
\end{figure*}

\begin{figure*}[!ht]
    \centering
    \includegraphics[width=\linewidth]{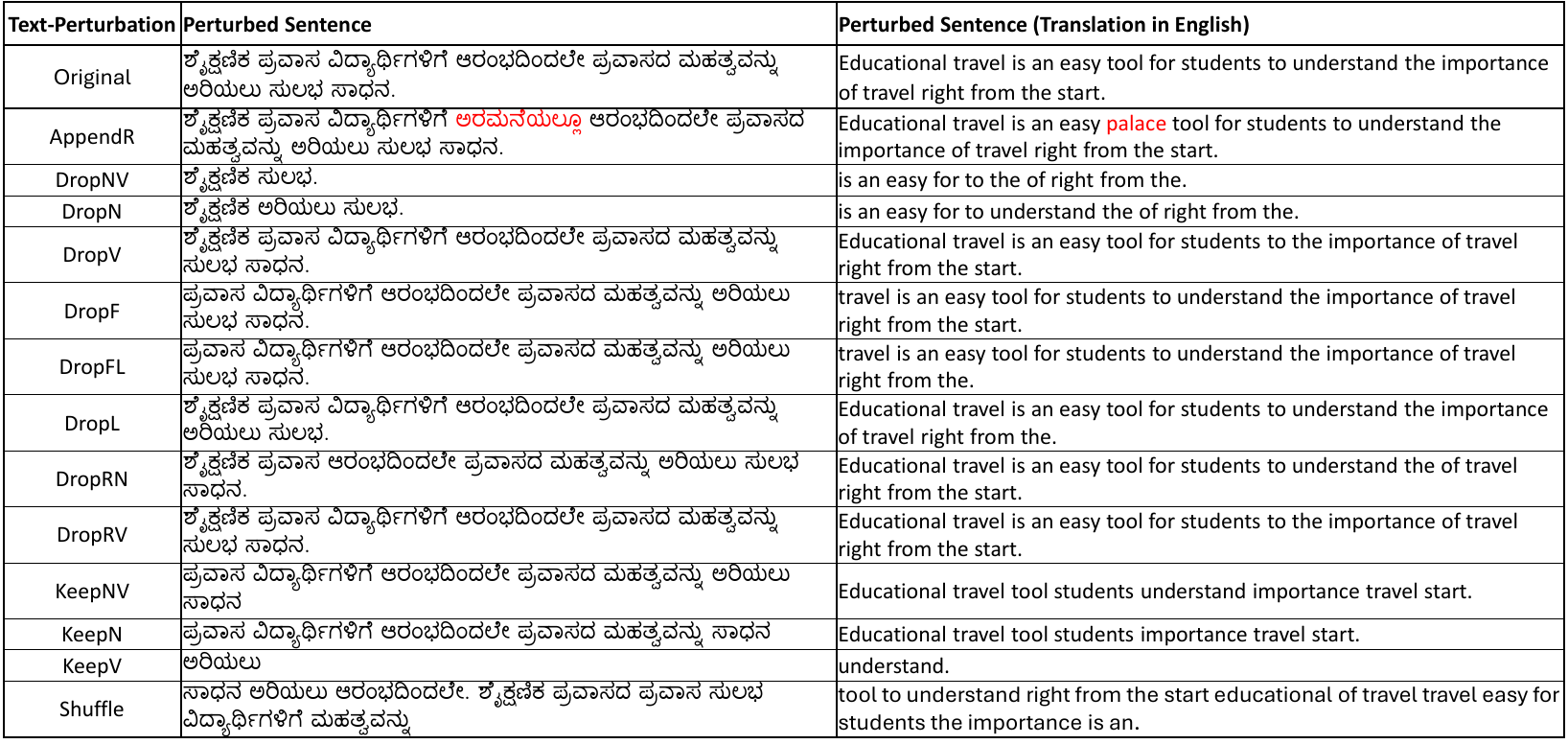}
    \captionof{table}{Text perturbation examples for Kannada language.}
    \label{fig:Sample of Kannada Perturbations}
\end{figure*}

\begin{figure*}[!ht]
    \centering
    \includegraphics[width=\linewidth]{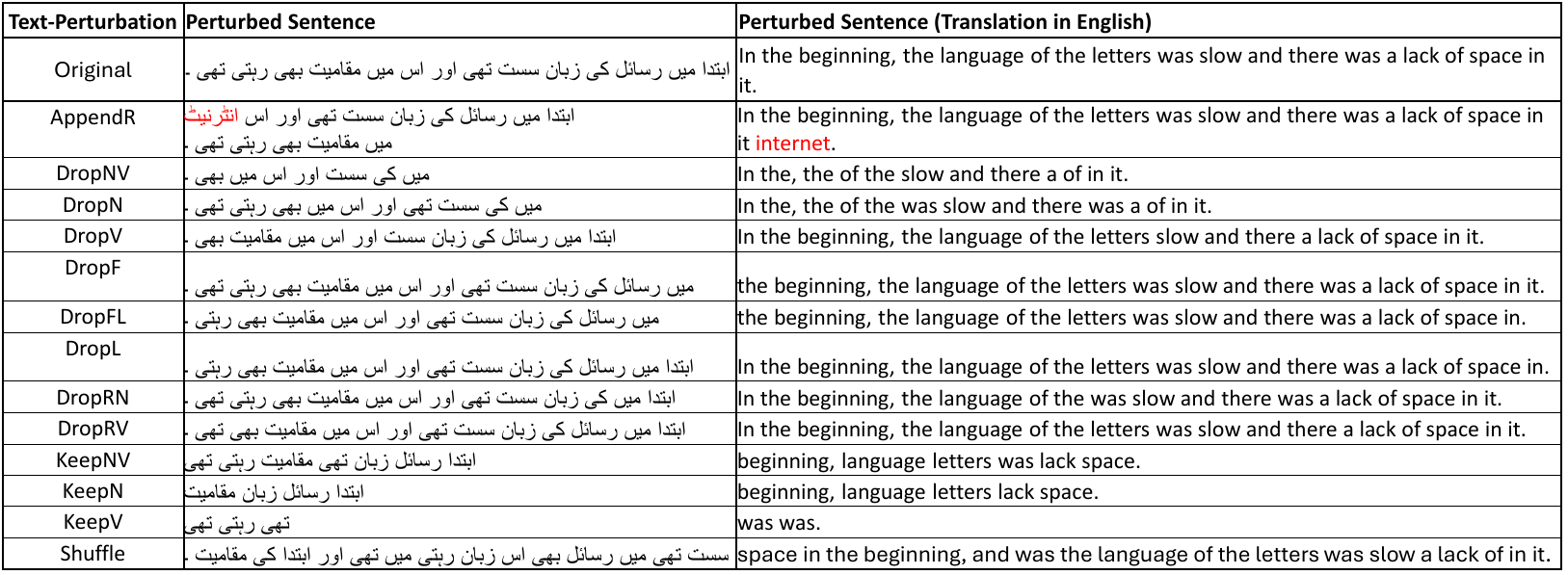}
    \captionof{table}{Text perturbation examples for Urdu language.}
    \label{fig:Sample of Urdu Perturbations}
\end{figure*}

Tables~\ref{fig:Sample of Telugu Perturbations}-\ref{fig:Sample of Urdu Perturbations} display an illustrative example of text perturbations for each perturbation type per language. For each language, the tables present the original sentence followed by a perturbed sentence resulting from each of the thirteen perturbations. Additionally, English translations are provided for enhanced comprehension.

% \begin{figure*}[!t]
%     \centering
%     \includegraphics[width=0.8\linewidth]{images/SSFExample2.png}
%     \captionof{figure}{A sample of an SSF formatted sentence in Hindi language}
%     \label{fig:SSFExample2}
% \end{figure*}

%https://docs.google.com/drawings/d/1ZGVwKxf3PtQ0lfBOMXr5hyUowb-nprLMJutcxKRU17M/edit

%https://docs.google.com/drawings/d/1JQjTd2PzULyed_6YKTtikv9XTKJdxVwV3xhDLuK15a0/edit?usp=sharing

%https://docs.google.com/drawings/d/1GH3JDGXV3ltcQVbPHoMFensh26ktXpkMOg1u_0h5S-o/edit?usp=sharing

%https://docs.google.com/drawings/d/1bBiAt0lC_PNADIhesvNxY9i06T6UM6OoCq8dwdo_87w/edit?usp=sharing

%https://docs.google.com/drawings/d/1OPAtKj21MWoUiWoD07smBevY4CTi8tVqn_Oqc76kiFk/edit?usp=sharing

%https://docs.google.com/drawings/d/12_nfkhFc0LwbFNTqCEMlsacSb1On6wJCtRVwEFDnzTA/edit?usp=sharing

% \begin{figure*}[!ht]
%     \centering
%     \includegraphics[width=\linewidth]{images/senlen_unimulti.pdf}
%     \vspace{-0.2cm}
%     \includegraphics[width=\linewidth]{images/treedepth_unimulti.pdf}
%     \includegraphics[width=\linewidth]{images/bshift_unimulti.pdf}
%     \caption{Probing task results: Average layerwise accuracy was computed across Universal multilingual (mBERT, XLM-R, InfoXLM, BLOOM, mT5)  and Indic multilingual (IndicBERT, MuRIL) representations on surface-level (top row) and syntactic probing tasks (bottom two rows).}
%     \label{fig:appendixsynatxandsemantics}
% \end{figure*}

% \begin{figure*}[!ht]
%     \centering
%     \includegraphics[width=\linewidth]{images/subnum_unimulti.pdf}
%     \vspace{-0.2cm}
%     \includegraphics[width=\linewidth]{images/objnum_unimulti.pdf}
%     \includegraphics[width=0.9\linewidth]{images/gender_unimulti.pdf}
%     \includegraphics[width=0.9\linewidth]{images/number_unimulti.pdf}
%     \includegraphics[width=0.9\linewidth]{images/person_unimulti.pdf}
%     \caption{Probing task results: Average layerwise accuracy was computed across Universal multilingual (mBERT, XLM-R, InfoXLM, BLOOM, mT5) and Indic multilingual (IndicBERT, MuRIL) representations on semantic probing tasks.}
%     \label{fig:appendixsemanticprobing}
% \end{figure*}

\section{Details of Multilingual Models}
\label{app:modelDetails}

\begin{table}[!ht]
    \centering
    \scriptsize
   \resizebox{\columnwidth}{!}{ \begin{tabular}{|l|l|c|}
    \hline
\textbf{Model}&\textbf{Pretraining objectives}&\textbf{MSL}\\\hline
mBERT-base~\cite{devlin2018bert}&MLM, NSP	&	512\\\hline
XLM-R-base~\cite{conneau2019unsupervised}&MLM, CLM, TLM&512\\\hline
InfoXLM-base~\cite{chi2020InfoXLM}&MLM, TLM, CLM, XLCO	&512\\\hline
BLOOM-base~\cite{workshop2022bloom}&CLM	&512\\\hline
mT5-base~\cite{xue2020mt5}&	MLM, CLM	&512\\\hline
mGPT~\cite{shliazhko2024mgpt}&	CLM	&512\\\hline
XGLM~\cite{lin-etal-2022-shot}&	CLM	&512\\\hline
IndicBERT-base~\cite{kakwani2020indicnlpsuite}	&MLM	&128\\\hline
MuRIL-base~\cite{khanuja2021muril}&	MLM, TLM	&512\\\hline
    \end{tabular}}
    %\vspace{-0.1cm}
    \caption{Details of multilingual Transformer-based models used in this study: MLM (masked language modeling), CLM (causal LM), TLM (translation LM), XLCO (cross-lingual contrastive learning), MSL (maximum sequence length).}
    \label{tab:modelDetails}
\end{table}

\begin{table}[!ht]
\scriptsize
\centering
\resizebox{\columnwidth}{!}{\begin{tabular}{|l|l|l|l|l|l|l|p{0.75in}|}
\hline
\textbf{} & \textbf{\texttt{hi}} & \textbf{\texttt{kn}} & \textbf{\texttt{ml}}& \textbf{\texttt{mr}} &\textbf{\texttt{te}} &\textbf{\texttt{ur}} &\textbf{Total Tokens (for all Indic languages)}\\ \hline
\textbf{mBERT (11)} &\multicolumn{6}{c|}{Detailed information not known publicly} & 184M\\ \hline
\textbf{IndicBERT (23)}  & 1.84B & 712M & 767M & 560M & 671M & - & 7.59B \\ \hline
\textbf{XLM-R (15)}  & 1.71B & 169M & 313M & 175M & 249M & 730M & 3.99B\\ \hline
\textbf{InfoXLM (13)} & 1.17B & 95.6M & 327M & 130M &225M  & 289M & 3.467B\\ \hline
\textbf{MuRIL(16)}  & 4.8B & 2.4B & 2.7B & 2.6B & 2.66B & 3.3B &37.76B\\ \hline
\textbf{BLOOM (13)} & \multicolumn{6}{c|}{Detailed information not known publicly} &2.7B\\ \hline
\textbf{mT5 (11)} & 24B & 1.1B & 1.8B  & 14B & 1.3B &2.4B  &58.3B\\ \hline
\textbf{mGPT} & 1.1B & 0.11B & 0.11B  & 0.11B & 0.11B &  0.2B&2.18B\\ \hline
\textbf{XGLM} & 3.45B & 446M &458M & 935M  & 689M &1.35B  &11.0B\\ \hline
\end{tabular}}
\caption{Tokens for pretraining multilingual models.}
\label{tab:Tokens_multilingual}
\end{table}

\subsection{Training data proportion of the tested languages}

Table~\ref{tab:data_proprtion_models} shows the proportion of training data for each Indic language within each model's total Indic language tokens. From Table~\ref{tab:data_proprtion_models}, we make the following observations: 

\noindent\textbf{Balanced vs. Imbalanced Distribution:}
\begin{itemize}
    \item MuRIL model stands out with the most balanced distribution among the six languages shown (12.7\% for Hindi, with other languages between 6-9\%).
    \item In contrast, mGPT shows high Hindi focus (50.5\%) with much smaller proportions for other languages.
\end{itemize}

\noindent\textbf{Hindi Dominance:}
\begin{itemize}
    \item Most models allocate the largest share of their Indic language tokens to Hindi.
    \item This varies dramatically from 12.7\% in MuRIL to 50.5\% in mGPT.
\end{itemize}

\noindent\textbf{Coverage Beyond Major Languages:}

\begin{itemize}
    \item MuRIL allocates 51.1\% of its tokens to "Other Indic" languages beyond the six shown, suggesting much broader coverage.
    \item This aligns with research showing MuRIL supports 17 Indian languages in total.
\end{itemize}

\begin{table}[!ht]
    \centering
    \scriptsize
 \resizebox{\columnwidth}{!}{   \begin{tabular}{|l|c|c|c|c|c|c|c|}
        \hline
        \textbf{Model} & \textbf{hi (Hindi)} & \textbf{kn (Kannada)} & \textbf{ml (Malayalam)} & \textbf{mr (Marathi)} & \textbf{te (Telugu)} & \textbf{ur (Urdu)} & \textbf{Other Indic} \\
        \hline
        IndicBERT & 24.2\% & 9.4\% & 10.1\% & 7.4\% & 8.8\% & -- & 40.1\%* \\\hline
        XLM-R     & 42.9\% & 4.2\% & 7.8\%  & 4.4\% & 6.2\% & 18.3\% & 16.2\%* \\\hline
        InfoXLM   & 33.7\% & 2.8\% & 9.4\%  & 3.7\% & 6.5\% & 8.3\% & 35.6\%* \\\hline
        MuRIL     & 12.7\% & 6.4\% & 7.2\%  & 6.9\% & 7.0\% & 8.7\% & 51.1\%* \\\hline
        mT5       & 41.2\% & 1.9\% & 3.1\%  & 24.0\% & 2.2\% & 4.1\% & 23.5\%* \\\hline
        mGPT      & 50.5\% & 5.0\% & 5.0\%  & 5.0\% & 5.0\% & 9.2\% & 20.3\%* \\\hline
        XGLM      & 31.4\% & 4.1\% & 4.2\%  & 8.5\% & 6.3\% & 12.3\% & 33.2\%* \\
        \hline
    \end{tabular}}
    \caption{Training data proportion of the six Indic languages for both universal and Indic multilingual language models.}
    \label{tab:data_proprtion_models}
\end{table}

\noindent\textbf{Hyper-parameters.}
We train logistic regression with a regularization parameter $C$=20 and use a L2 penalty term. For multi-class tasks, we use the ``multinomial'' setting while training. All experiments were done on a machine with a T4 GPU. 

\section{Probing Results}
\label{probing_results}
\begin{figure*}[!t]
    \centering
    \includegraphics[width=0.4\linewidth]{images/colorbar_final.pdf} \\
     \includegraphics[width=0.325\linewidth]{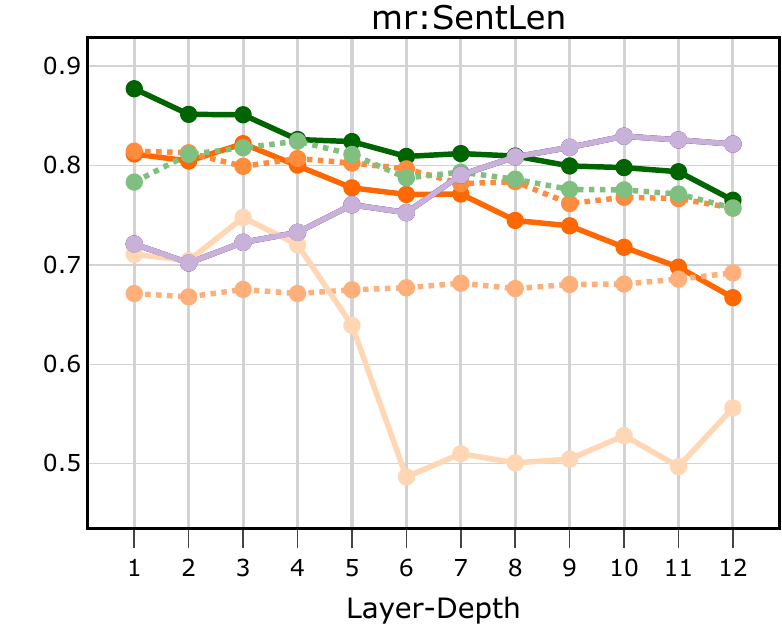}
    \includegraphics[width=0.325\linewidth]{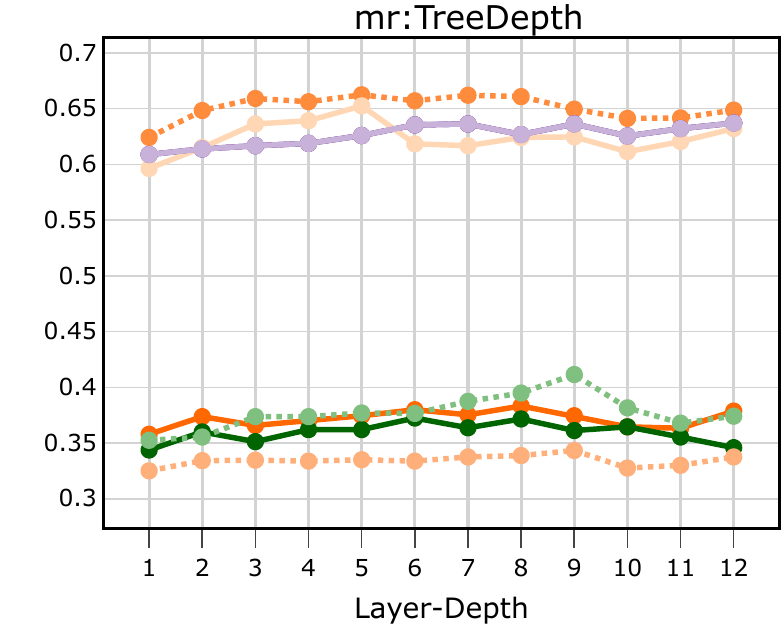}
    \includegraphics[width=0.325\linewidth]{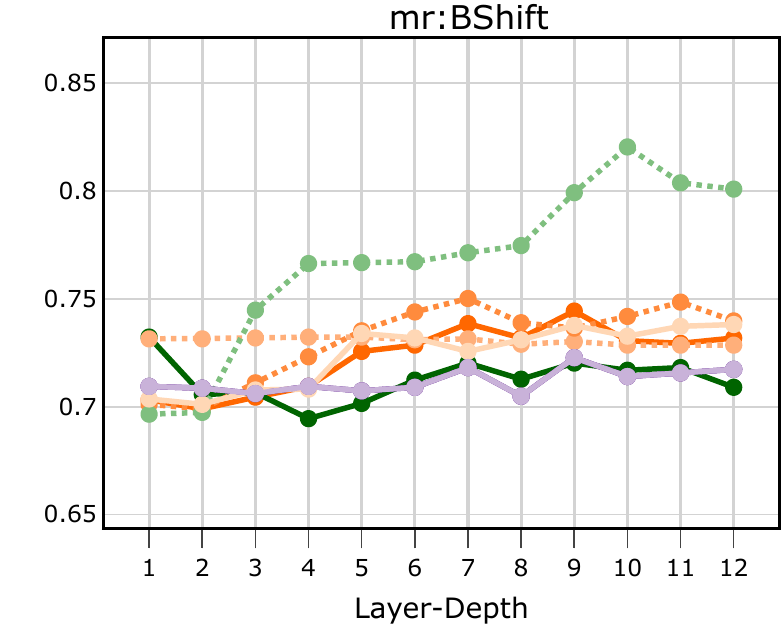}
    \includegraphics[width=0.325\linewidth]{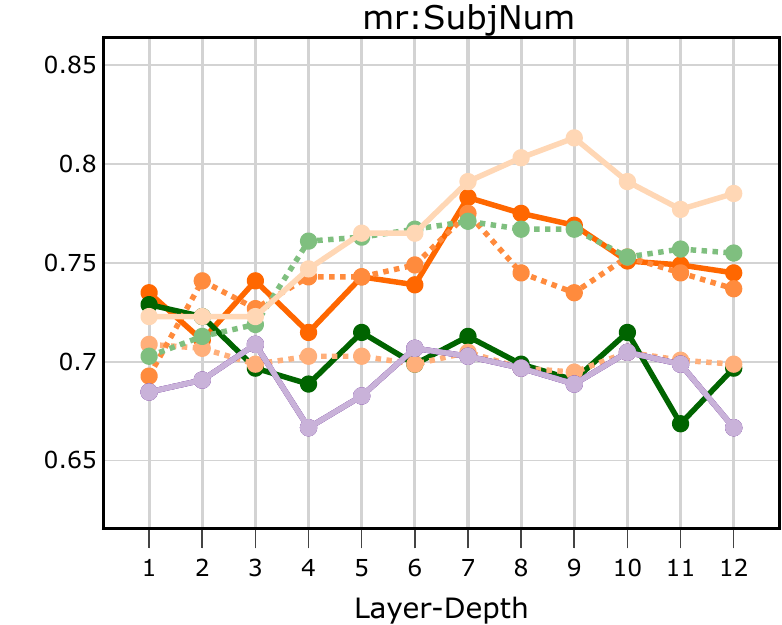}
     \includegraphics[width=0.325\linewidth]{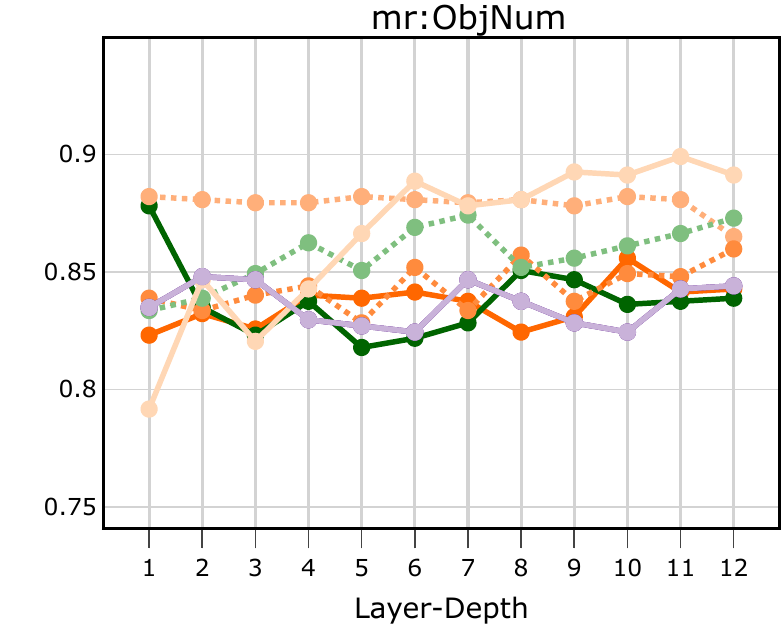}
    \includegraphics[width=0.325\linewidth]{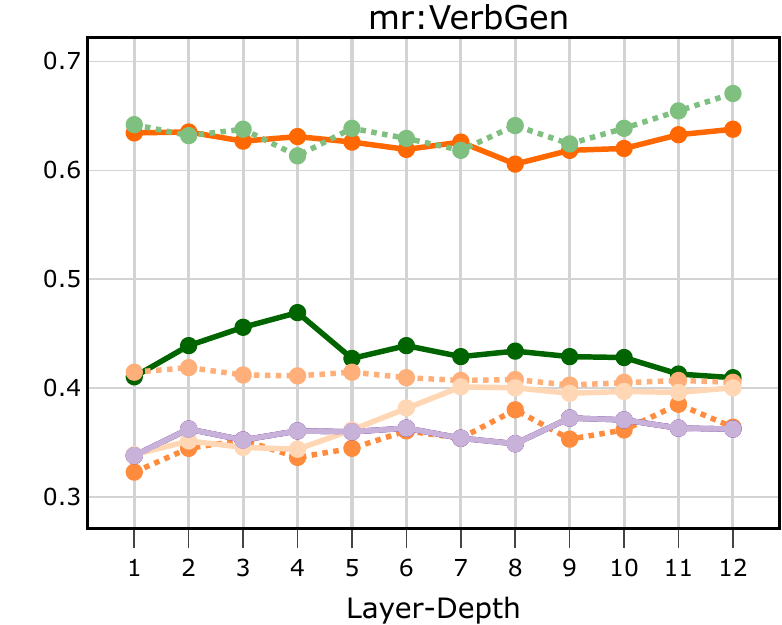}
    \includegraphics[width=0.325\linewidth]{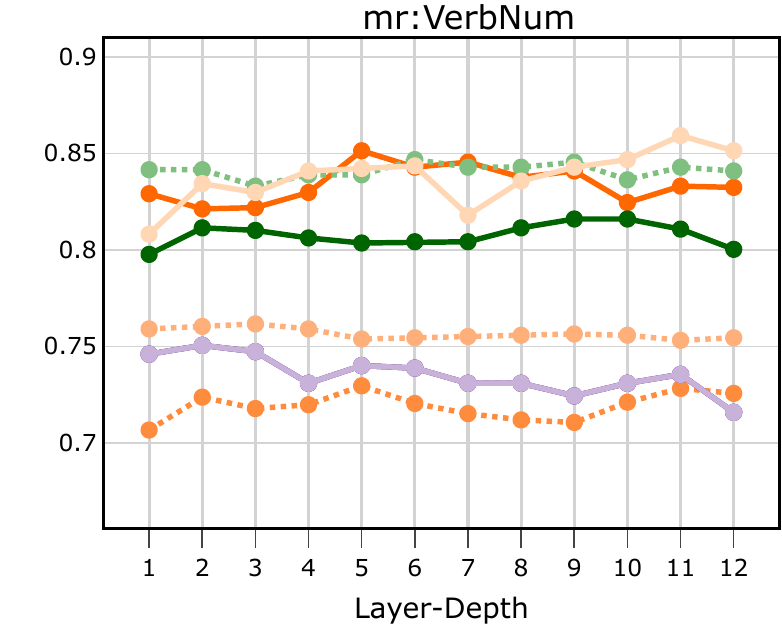}
    \includegraphics[width=0.325\linewidth]{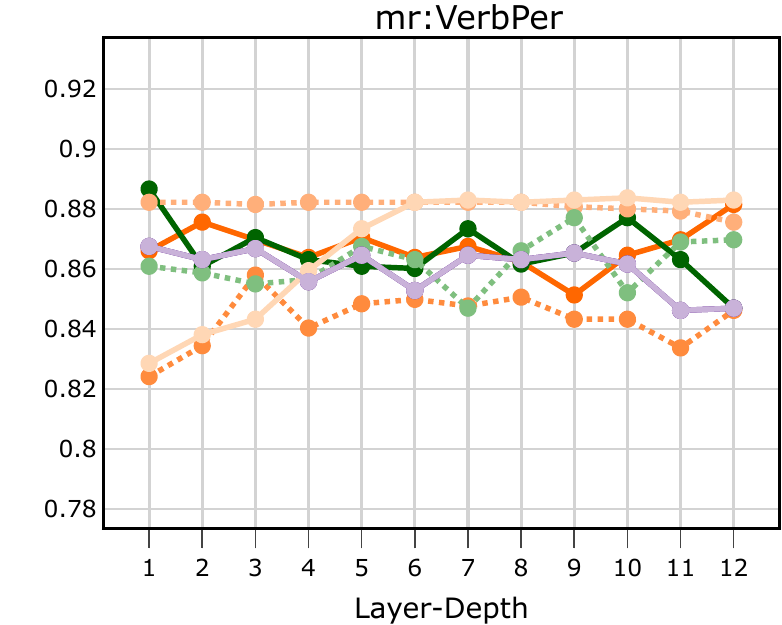}
     
\caption{Marathi language probing task results: Layerwise accuracy comparisons between various multilingual representations on 8 probing tasks.}
\label{fig:mr_probing_tasks}
\end{figure*}

\begin{figure*}[!t]
    \centering
    \includegraphics[width=0.4\linewidth]{images/colorbar_final.pdf} \\
    \includegraphics[width=0.325\linewidth]{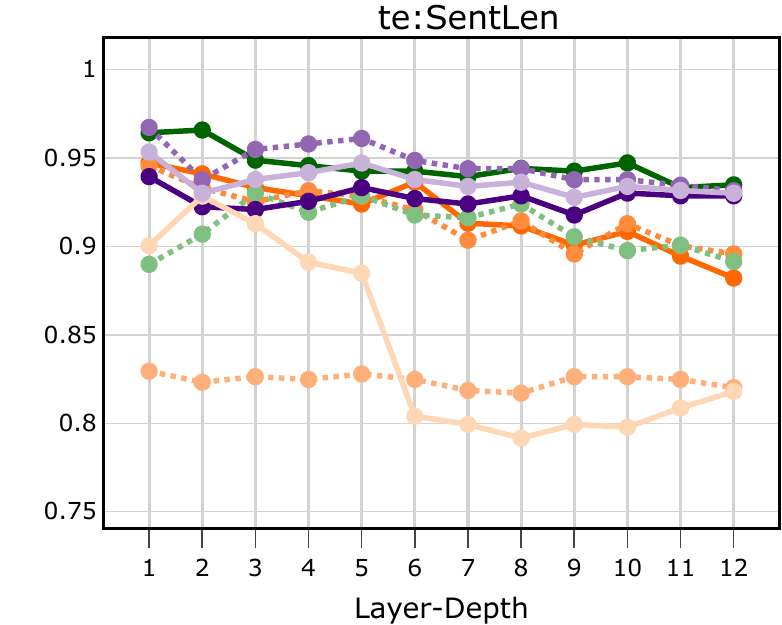}
    \includegraphics[width=0.325\linewidth]{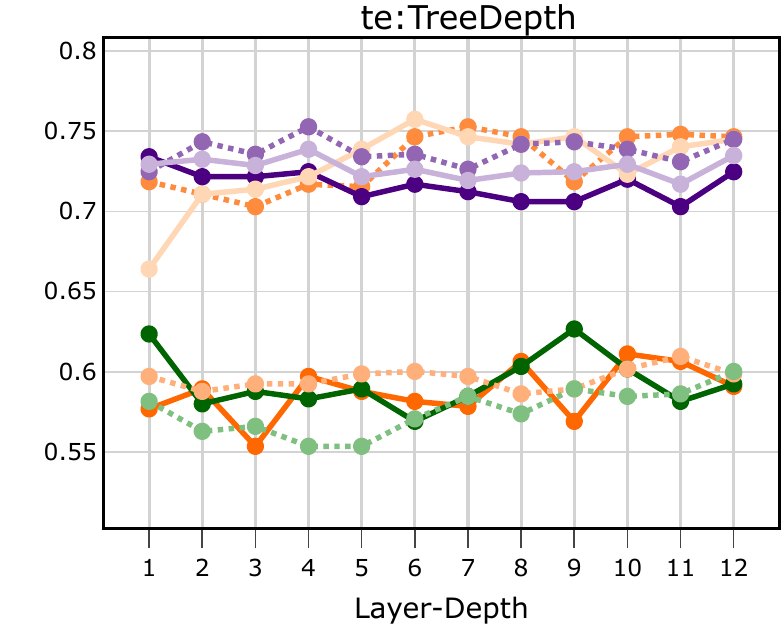}
    \includegraphics[width=0.325\linewidth]{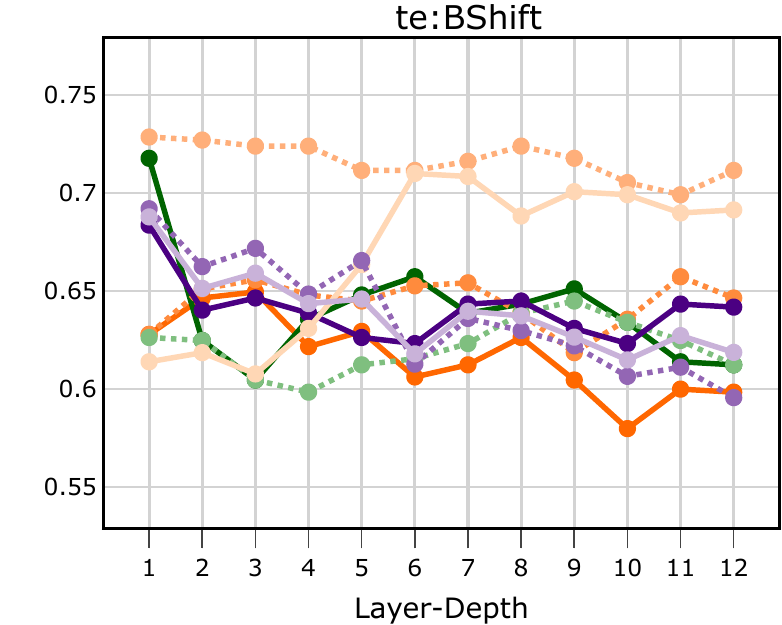}
    \includegraphics[width=0.325\linewidth]{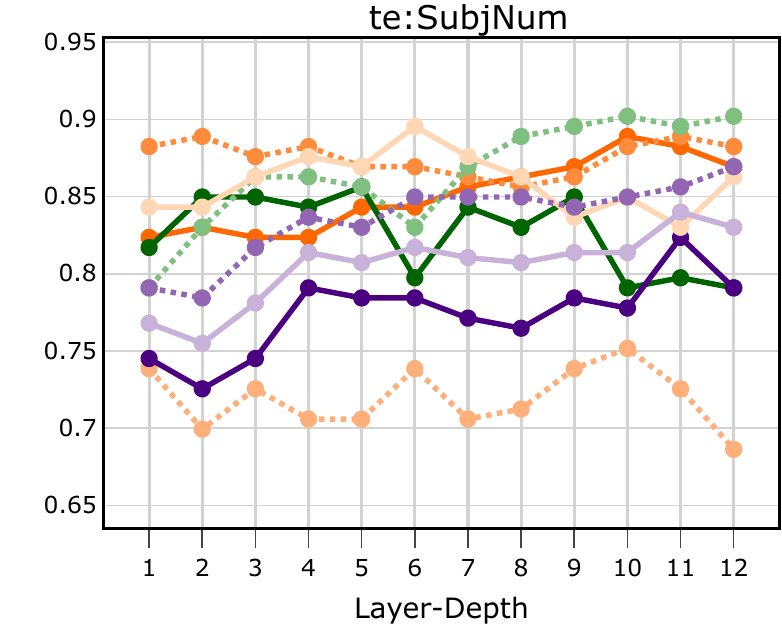}
    \includegraphics[width=0.325\linewidth]{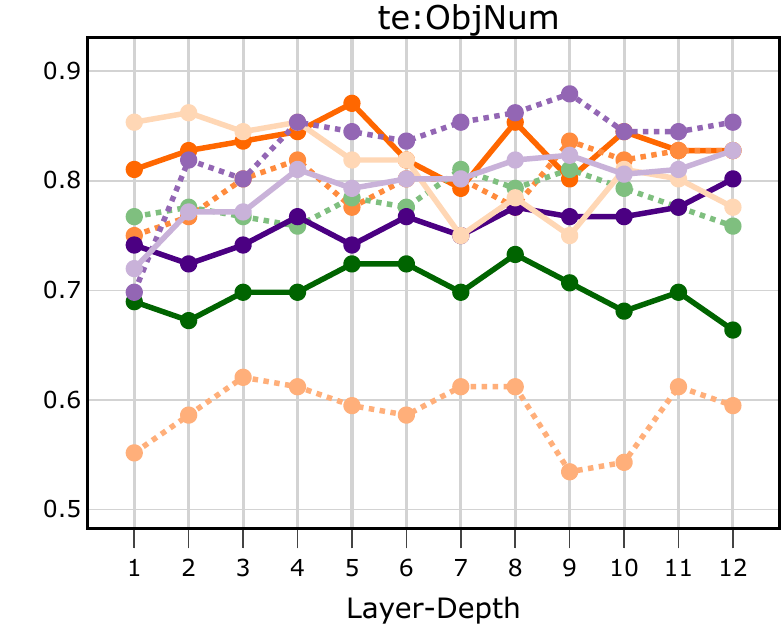}
    \includegraphics[width=0.325\linewidth]{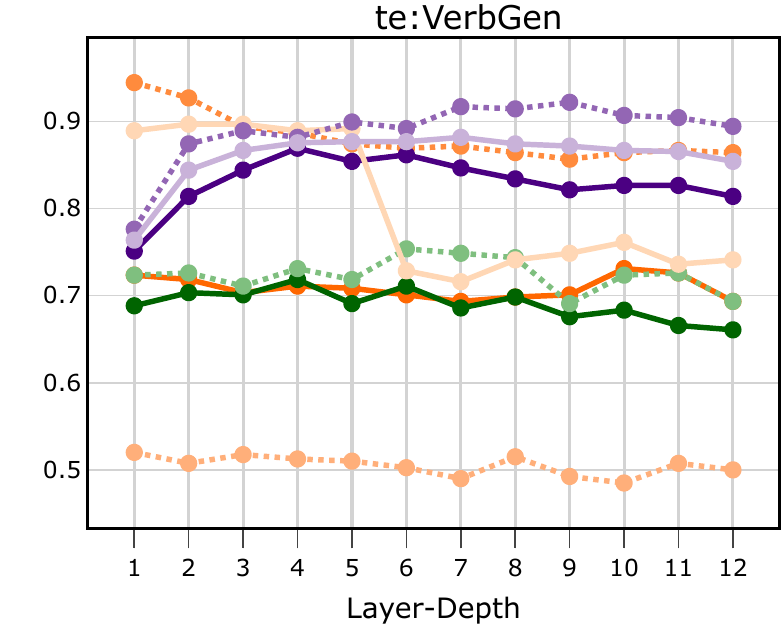}
    \includegraphics[width=0.325\linewidth]{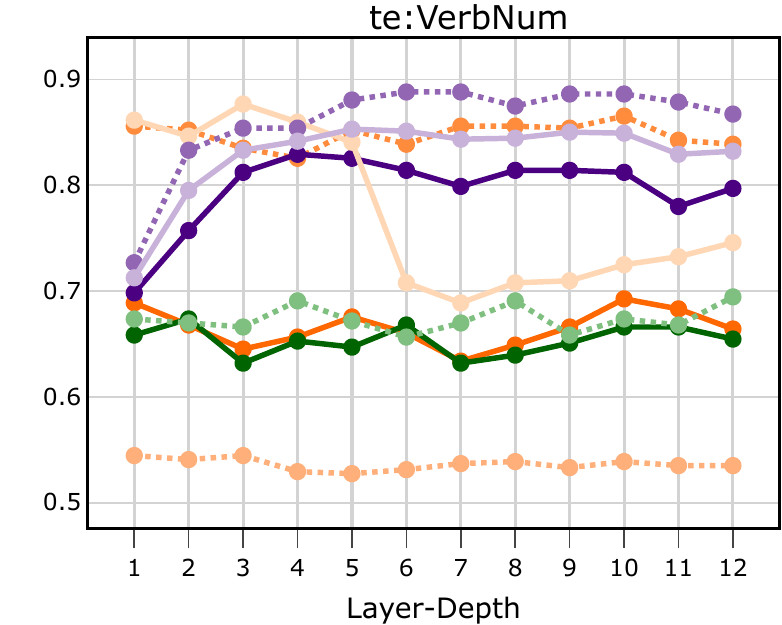}
    \includegraphics[width=0.325\linewidth]{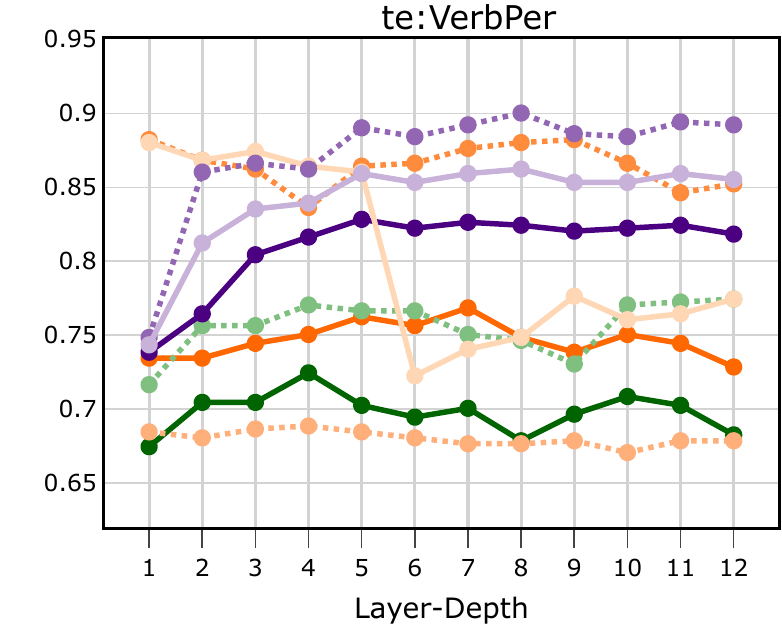}
     
\caption{Telugu language probing task results: Layerwise accuracy comparisons between various multilingual representations on 8 probing tasks.}
\label{fig:te_probing_tasks}
\end{figure*}

\begin{figure*}[!t]
    \centering
    \includegraphics[width=0.4\linewidth]{images/colorbar_final.pdf} \\
    \includegraphics[width=0.325\linewidth]{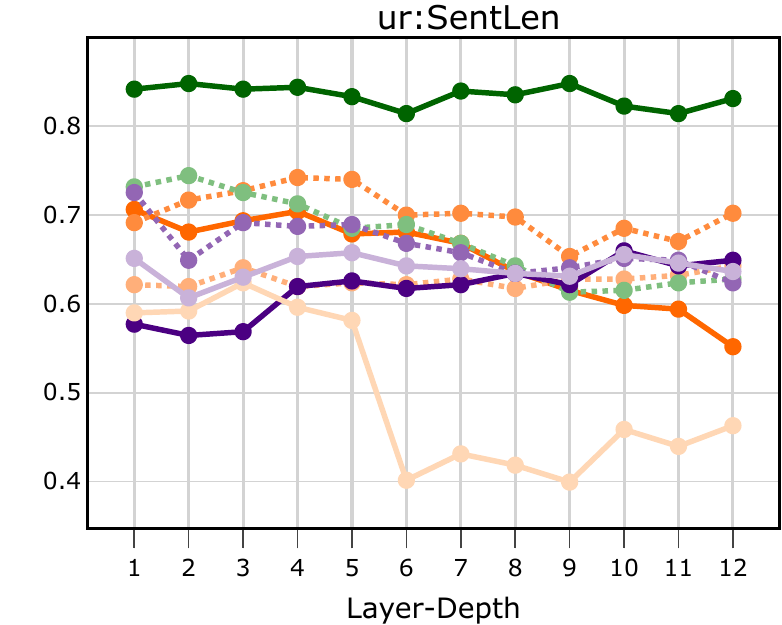}
    \includegraphics[width=0.325\linewidth]{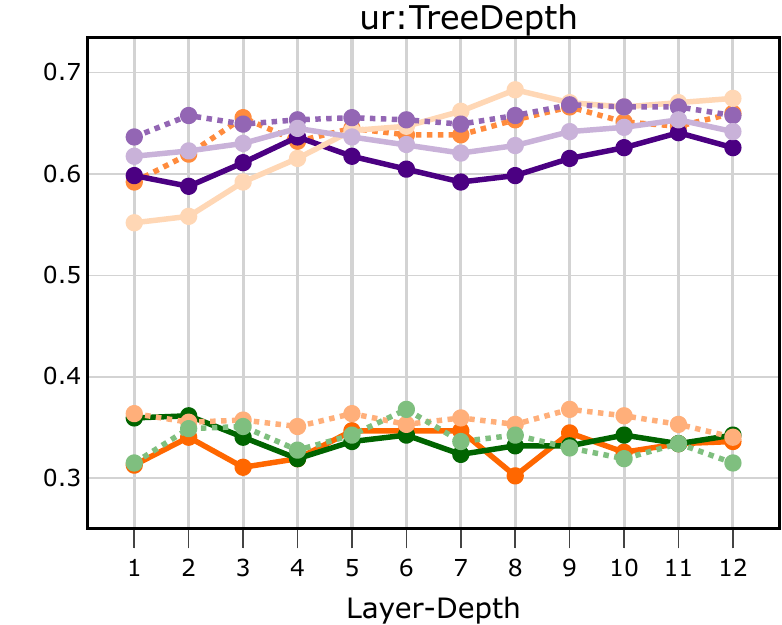}
    \includegraphics[width=0.325\linewidth]{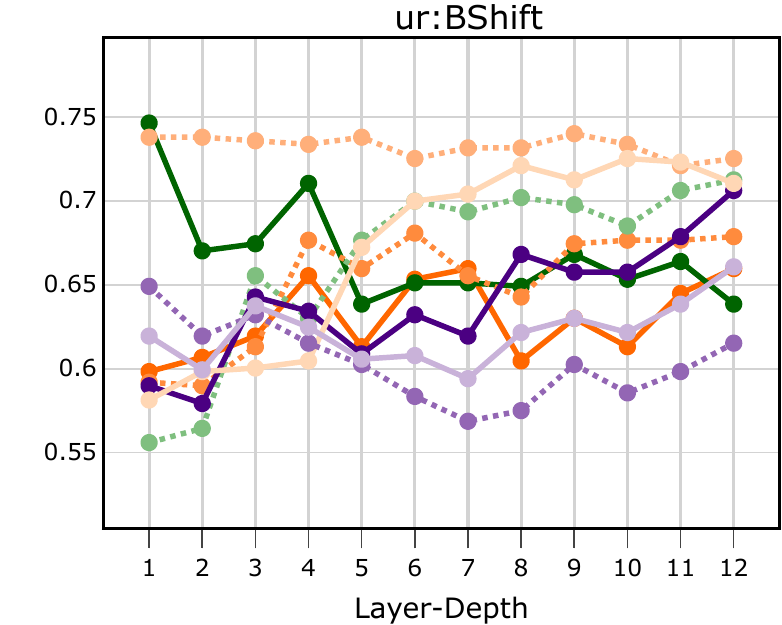}
    \includegraphics[width=0.325\linewidth]{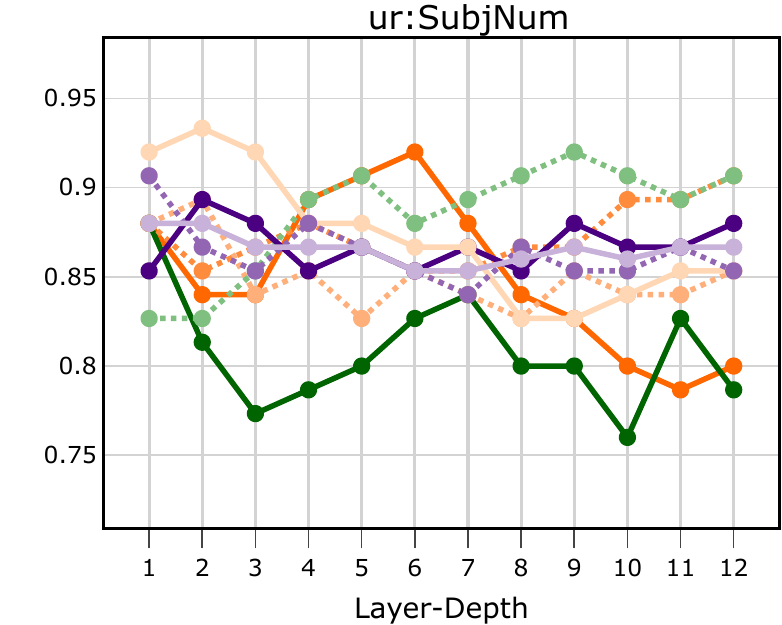}
    \includegraphics[width=0.325\linewidth]{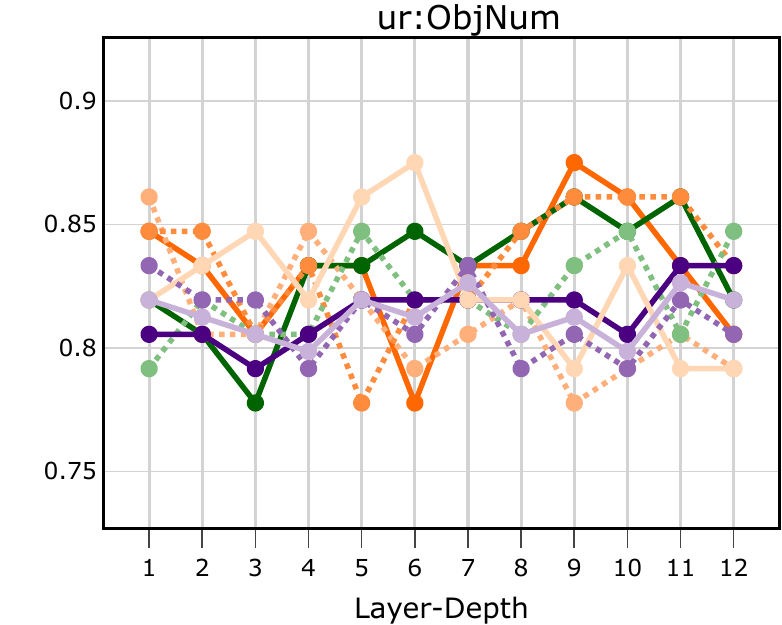}
    \includegraphics[width=0.325\linewidth]{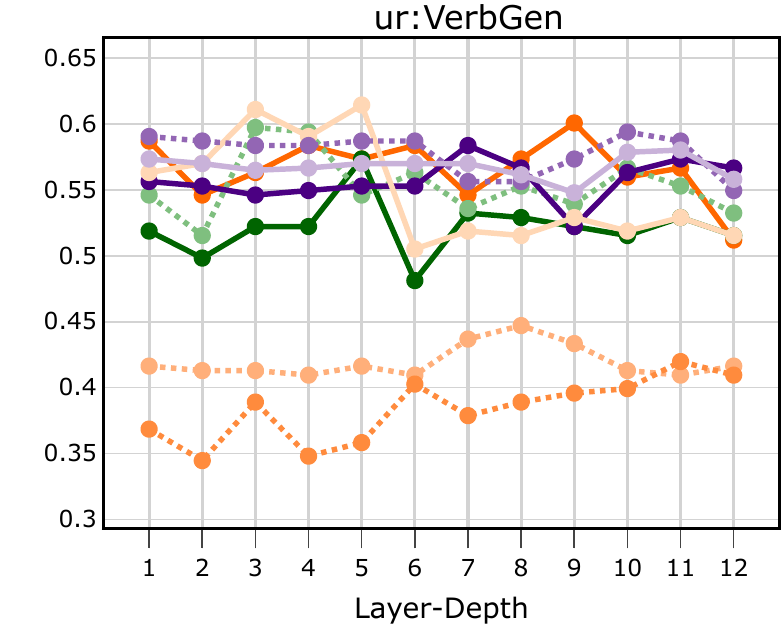}
    \includegraphics[width=0.325\linewidth]{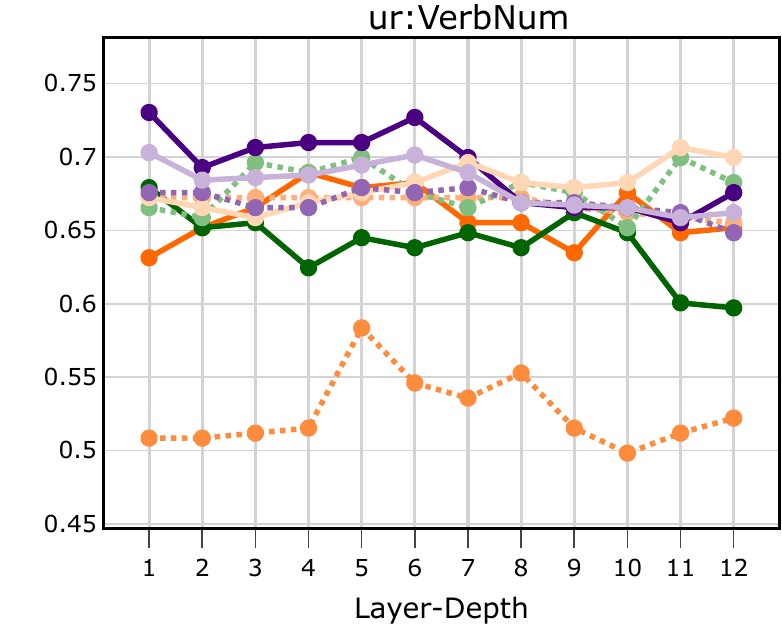}
    \includegraphics[width=0.325\linewidth]{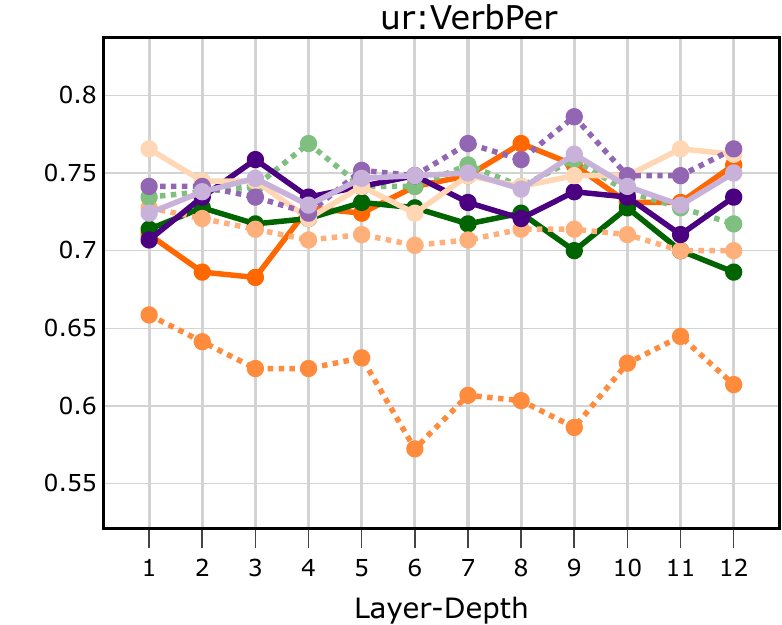}
\caption{Urdu language probing task results: Layerwise accuracy comparisons between various multilingual representations on 8 probing tasks.}
\label{fig:ur_probing_tasks}
\end{figure*}

We report the layerwise probing accuracies for individual multilingual models in 
Figs.~\ref{fig:hi_probing_tasks} to~\ref{fig:ur_probing_tasks}.

\noindent\textbf{Surface-level tasks}
Analyzing the performance across languages, we observe the following patterns: (i) For universal multilingual models (mBERT, mT5 and InfoXLM) as well as for Indic models (IndicBERT and MuRIL), there is a trend of higher accuracy in the early (or lower) layers, which decreases in the later (or higher) layers. This pattern is expected, as surface-level tasks generally require minimal processing. (ii) Notably, the universal model BLOOM deviates from this trend, showing lower accuracy in the early layers and higher accuracy in the later layers. This unusual pattern in  BLOOM could be attributed to its unique autoregressive architectural design, and its approach to representing languages, especially those less common in the dataset. This suggests that BLOOM may require its deeper layers to effectively grasp the nuances of these languages, even for tasks at the surface level. (iii) Overall, among all the models, IndicBERT reports best accuracy, while XLM-R displays poorer performance. This is likely because IndicBERT is specifically trained on Indic languages, making it more attuned to the nuances and idiosyncrasies of these languages.
On the other hand, models like XLM-R, which are designed for universal applicability, might struggle with specific linguistic features unique to Indic languages. These features include script differences, morphological complexity, and visual factors such as orthography and word length.

\noindent\textbf{Syntactic Tasks}
% In the last two rows of Fig.~\ref{fig:surface_syntactic_tasks}, we display the accuracy scores for syntactic tasks. 
For TreeDepth, we observe that probing accuracy tends to be higher in the middle layers for various (model, language) combinations. This trend is particularly notable in three universal models, mBERT, mT5 and InfoXLM, and an Indic model, MuRIL. Moreover, this pattern is consistent for three languages: \texttt{hi}, \texttt{kn} and \texttt{ml}. However, other multilingual models do not exhibit any clear layer-wise trend, and the same applies to the other three Indic languages. This suggests a model-specific and language-specific affinity in handling syntactic complexity, where certain models are more adept at processing syntactic information in specific layers, and this proficiency varies across different languages.
In contrast to TreeDepth, for BShift task, we generally observe higher probing accuracy in the later layers for four of the models across various languages. Notably, BLOOM exhibits a decreasing trend in accuracy for \texttt{kn} and \texttt{te}, while showing an increasing trend for \texttt{ur}. This suggests different models' layers may specialize in different types of syntactic processing, with some models better handling tasks like BShift in their later layers.
Overall, when comparing performance across models and tasks, InfoXLM and mT5 stand out for their superior accuracy in capturing tree depth information across different languages. Conversely, MuRIL shows notable proficiency in BShift. These distinctions highlight how different models may be better suited for different types of syntactic analyses.
These observations contribute to a deeper understanding of how various multilingual models process syntactic information, demonstrating both model-specific and language-specific trends and capabilities in linguistic tasks.
%for some languages, XLM-R works better while for other languages, MuRIL gives best results.
%(iii) 
%Overall, we conclude that multilingual Transformer models process syntactic information from middle to later layers. Also, there is no one best model that works across all six languages. 

\noindent\textbf{Semantic Tasks}
These include SubjNum and ObjNum, and VerbGen, VerbNum and VerbPer. We do not have results for \texttt{ml} for VerbGen, VerbNum and VerbPer, since we do not have labeled data for \texttt{ml} for those tasks. 

From SubjNum and ObjNum results, we make the following observations. For mBERT, InfoXLM, mT5, BLOOM as well as MuRIL, we observe an increasing trend from lower to higher layers for \texttt{hi}, \texttt{mr} and \texttt{te}. This suggests that for these languages, the models become more proficient in handling semantic tasks related to SubjNum and ObjNum as we move to the higher layers. %Within the Indic models, specifically MuRIL, this increasing trend is also evident for the same three languages (Hindi, Marathi, and Telugu). This consistency across both universal and Indic models for these languages indicates a shared pattern in how these models process semantic information. 
When considering other languages and models, we note that IndicBERT exhibits an increasing trend for the ObjNum task, specifically for \texttt{ur}. This highlights the variability in model performance based on the language. Interestingly, the middle layers of the models show higher probing accuracy for languages such as \texttt{hi} and \texttt{ml} in the SubjNum task. %This indicates that, for these languages, the mid-level layers are more effective in semantic processing related to SubjNums. 
Similar to its performance in surface and syntactic tasks, XLM-R exhibits lower accuracy and lacks a discernible trend when it comes to capturing semantics.
% We make the following observations for the SubjNum and ObjNum tasks: (1) An increasing trend from lower to higher layers observed in universal multilingual (mBERT, InfoXLM, and BLOOM) for languages Hindi, Marathi, and Telugu. In Indic models, we observe this pattern in MuRIL for same three languages. Considering other languages and models, Indic BERT has increasing trend for ObjNum for Urdu. We also observe that middle layers display higher probing accuracy for languages such as Hind and Malayalam for SubjNum.
MuRIL has the highest probing accuracy although it performs the worst for \texttt{kn} ObjNum task.

%(2) Finding SubjNum in Hindi and ObjNum in Malayalam is difficult. (3) In general, later layers seem to show better results compared to earlier or middle layers.
For the verb-related tasks, Indic model MuRIL, performs the best for most languages and tasks. Also, in most cases, the last layer is the most predictive, except in \texttt{te} for gender and person detection, where initial layers provide better results. This indicates that gender and person detection in \texttt{te} is straightforward and does not need deep processing. For the universal models, mBERT and BLOOM report an increasing trend across languages and tasks. On the other hand, mT5 showcases an increasing trend for \texttt{kn} and \texttt{mr} languages and decreasing trend for \texttt{hi}, \texttt{te} and \texttt{ur} languages.
Models, including, IndicBERT, XLM-R and InfoXLM do not show any trend and have constant performance across layers.

\begin{figure*}[!ht]
    \centering
    \includegraphics[width=0.4\linewidth]{images/colorbar_final.pdf} \\
    \includegraphics[width=0.49\linewidth]{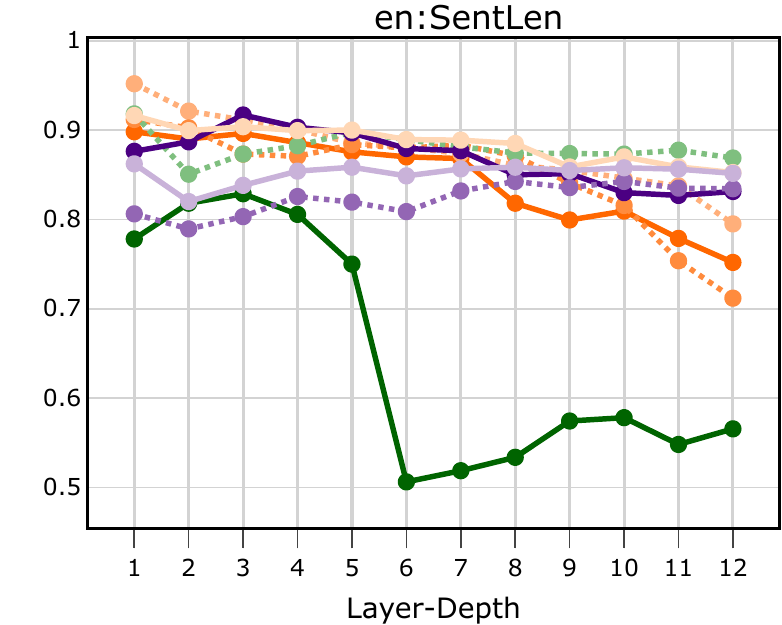}
    %%\vspace{-0.2cm}
    \includegraphics[width=0.49\linewidth]{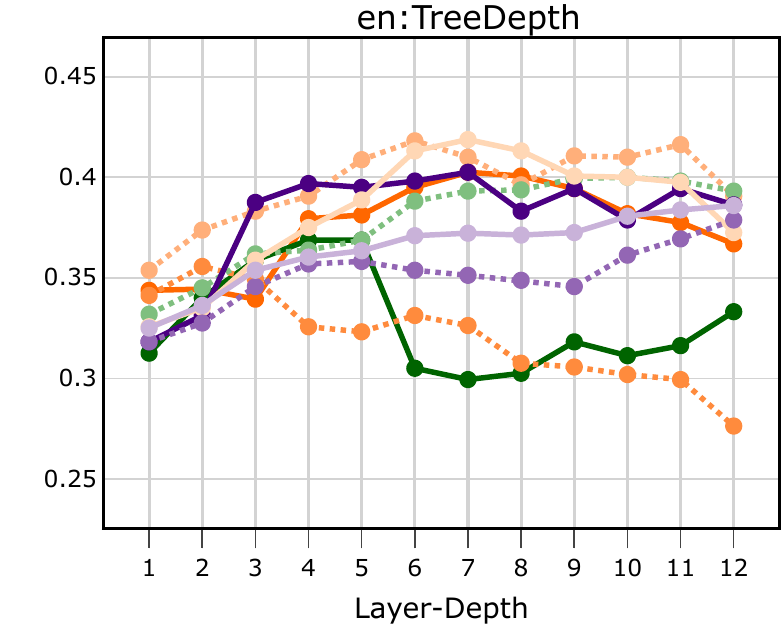}
    \includegraphics[width=0.49\linewidth]{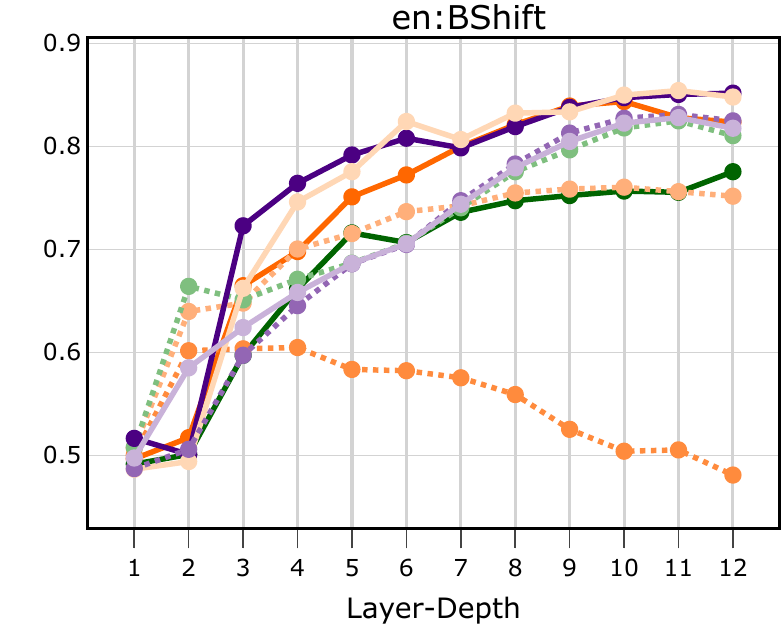}
    \includegraphics[width=0.49\linewidth]{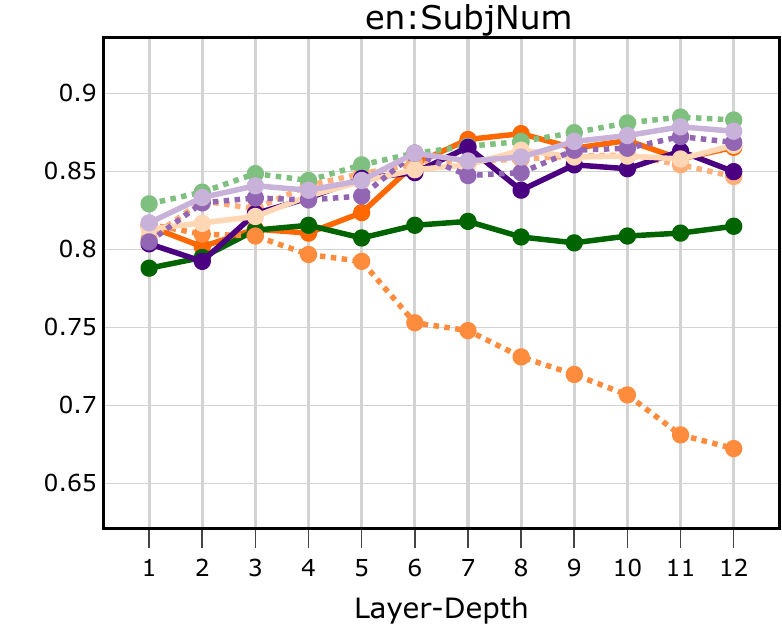}
    \includegraphics[width=0.49\linewidth]{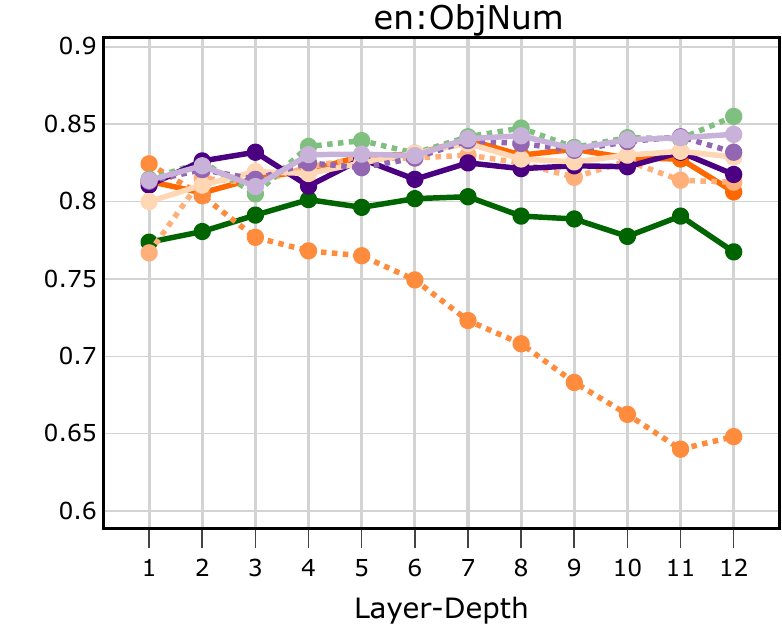}
    \caption{English language probing results: Layerwise accuracy was computed across Universal multilingual (mBERT, XLM-R, InfoXLM, BLOOM, mT5, mGPT and XGLM) and Indic multilingual (IndicBERT, MuRIL) representations on surface-level,  syntactic probing and semantic probing tasks.}
    \label{fig:appendixenglishresults}
\end{figure*}

\section{Extended Discussion}
\label{app:discussion}

We evaluated 9 multilingual Transformer-based models on 8 probing tasks to understand linguistic structures in 6 Indic languages, using our contributed \textsc{IndicSentEval} dataset. Indic-specific models like MuRIL and IndicBERT excel due to targeted training, while universal models like mBERT, InfoXLM, BLOOM, mGPT, and XGLM show mixed results. Perturbation analysis reveals decoder-based models are the most robust, as also noted by \citet{neerudu2023robustness}. Verbs and word order are key signals for encoding linguistic structures. TreeDepth is the most sensitive to perturbations, while SubjNum and ObjNum are the most resilient.

Overall, our study represents the first analysis of the interpretability of both Universal and Indic multilingual language models across six Indic languages where several languages have large training corpora while some have less. Our scientific findings from this model interpretability analysis via both probing and perturbations shed light on how language models capture language hierarchy and how training data influences the language understanding across layers in these models. 
%In the probing analysis, Indic-specific models like MuRIL and IndicBERT are likely the best at capturing language hierarchy while Universal models like mBERT, InfoXLM, BLOOM, mGPT and XGLM show mixed results. 
Surprisingly, universal models show greater resilience to perturbations in at least four Indic languages. In contrast, the universal model (mBERT) and the Indic-specific models (IndicBERT and MuRIL) display a more significant accuracy drop across all the Indic languages. This suggests the necessity for multilingual models that are robust to perturbations and can capture language hierarchy regardless of their training data. Overall, our findings demonstrate significant variability in the ability of current multilingual models to capture surface, syntactic, and semantic structures across different Indic languages. A hierarchy of language proficiency is discernible primarily for languages with more extensive training datasets. This highlights the necessity for innovative approaches in multilingual modeling that can accurately capture language structures, even in low-resource languages.

\end{document}